\documentclass[review]{elsarticle}

\usepackage{lineno,hyperref}
\usepackage{amsfonts}
\usepackage{algorithm}
\usepackage{algpseudocode}
\usepackage{bm}
\usepackage{hyperref}
\usepackage[table,xcdraw]{xcolor}
 \usepackage{lscape}

\newtheorem{lemma}{Lemma}

\newenvironment{proof}{\paragraph{Proof:}}{\hfill$\square$}

\begin{document}
	
	\begin{frontmatter}
		
		\title{GANs for learning from very high class conditional noisy labels
			%			\tnoteref{mytitlenote}
		}
		%		\tnotetext[mytitlenote]{Fully documented templates are available in the elsarticle package on \href{http://www.ctan.org/tex-archive/macros/latex/contrib/elsarticle}{CTAN}.}
		
		%% Group authors per affiliation:
		\author{Sandhya Tripathi\corref{mycorrespondingauthor}}		\cortext[mycorrespondingauthor]{Corresponding author}
		\ead{sandhya.tripathi@iitb.ac.in} %\fnref{myfootnote}}
		\author{N. Hemachandra}
		\address{IEOR, IIT Bombay}
		%\fntext[myfootnote]{Since 1880.}
		
		%% or include affiliations in footnotes:
		%		\author[mymainaddress,mysecondaryaddress]{Elsevier Inc}
		%		\ead[url]{www.elsevier.com}
		%		
		%		\author[mysecondaryaddress]{Global Customer Service\corref{mycorrespondingauthor}}

		%		\address[mymainaddress]{1600 John F Kennedy Boulevard, Philadelphia}
		%		\address[mysecondaryaddress]{360 Park Avenue South, New York}
		
		\begin{abstract}

			We use Generative Adversarial Networks (GANs) to design a class conditional label noise (CCN) robust scheme for binary classification. It first generates a set of correctly labelled data points from noisy labelled data and {\em 0.1\% or 1\%} clean labels such that the generated and true (clean) labelled data distributions are {\em close}; generated labelled data is used to learn a {\em good} classifier. The mode collapse problem while generating correct feature-label pairs and the problem of skewed feature-label dimension ratio ($\sim$ 784:1) are avoided by using Wasserstein GAN and a simple data representation change. Another WGAN with information theoretic flavour on top of the new representation is also proposed. The major advantage of both schemes is their significant improvement over the existing ones in presence of very high CCN rates, \emph{without} either estimating or cross validating over the noise rates. We proved that KL divergence between clean and noisy distribution increases w.r.t. noise rates in symmetric label noise model; can be extended to high CCN rates. This implies that our schemes perform well due to the adversarial nature of GANs. Further, use of generative approach (learning clean joint distribution) while handling noise enables our schemes to perform better than  discriminative approaches like GLC, LDMI and GCE; even when the classes are highly imbalanced. Using Friedman F test and Nemenyi posthoc test, we showed that on high dimensional binary class synthetic, MNIST and Fashion MNIST datasets, our schemes outperform the existing methods and demonstrate consistent performance across noise rates.
		\end{abstract}
		
		
	\end{frontmatter}
	
	%\linenumbers
	
	\section{Introduction} \label{sec: int}
	Label noise in supervised data for classification task has become ubiquitous due to the deluge of data availability, but from unreliable sources, in many cases. It renders the learning algorithms with very bad performance. In particular, for currently popular Deep Neural Networks (DNNs), \cite{zhang2016understanding} have shown that even though DNNs have high capacity to fit random labels, they lead to large generalization error.
	% \textcolor{red}{see again about noise type and approaches}
	Noise type, defined in terms of label flipping probability or noise rates, can be broadly categorized as follows: (1) Uniform/Symmetric/Random label noise (SLN): label flipping probability is same across the dataset, (2) Class conditional noise (CCN): label flipping probability depends on the class of data point, (3) Instance dependent noise (IDN): label flipping probability depends on the data point and hence different for all data points. Four major approaches for learning in the presence of label noise \cite{sastry2017robustbookchapter} are as follows: (1) Noise cleaning: correct labels are restored (2) Eliminating noisy points: after identifying the noisy points they are eliminated (3) Designing schemes for dealing with label noise: goal is to minimize the effect of label noise (4) Noise tolerant algorithms: designing algorithms that are unaffected by the label noise. 
	
	Recently, use of  a small set of clean labels by algorithms for learning from label noise is gaining popularity \cite{hendrycks2018using,ren2018learning}. Also, ideas like importance re-weighting \cite{liu2016classification}, human assisted approaches for noise matrix estimation \cite{han2018masking}, dimensionality of deep representations \cite{ma2018dimensionality}, have been used for dealing with label noise. However, most of the solutions estimate the noise rates or demonstrate the results on noise rates which reduce to SLN in binary class setup. Generative Adversarial Networks (GANs) \cite{goodfellow2014generative} have taken everything by storm by being able to provide a solution in some way to almost every learning problem. GANs exploit the adversarial relation between two networks: generator and discriminator, to  synthesize new samples by mimicking a target distribution.
	
	In this paper, using a small set of clean labels along with noisy labels, we propose a GAN based class conditional label noise robust binary classification framework. We adapt GANs in our setup by feeding small set of clean labelled data points to the discriminator and noisy labelled data points to the generator and asking the generator to output \emph{correctly} labelled data points.  Unlike the noise cleaning approach where only labels are corrected, our GANs based correctly labelled data generation scheme generates correct new feature-label pairs. Also, unlike existing schemes, our scheme completely avoids any estimation of or cross-validation over the noise rates. It decreases one source of estimation error. Also, due to the adversarial nature of GANs, we observe that it performs better than the existing schemes in the presence of very high noise. 
	
\begin{figure*}[!h]
	\centering
	\includegraphics[width=0.95\textwidth]{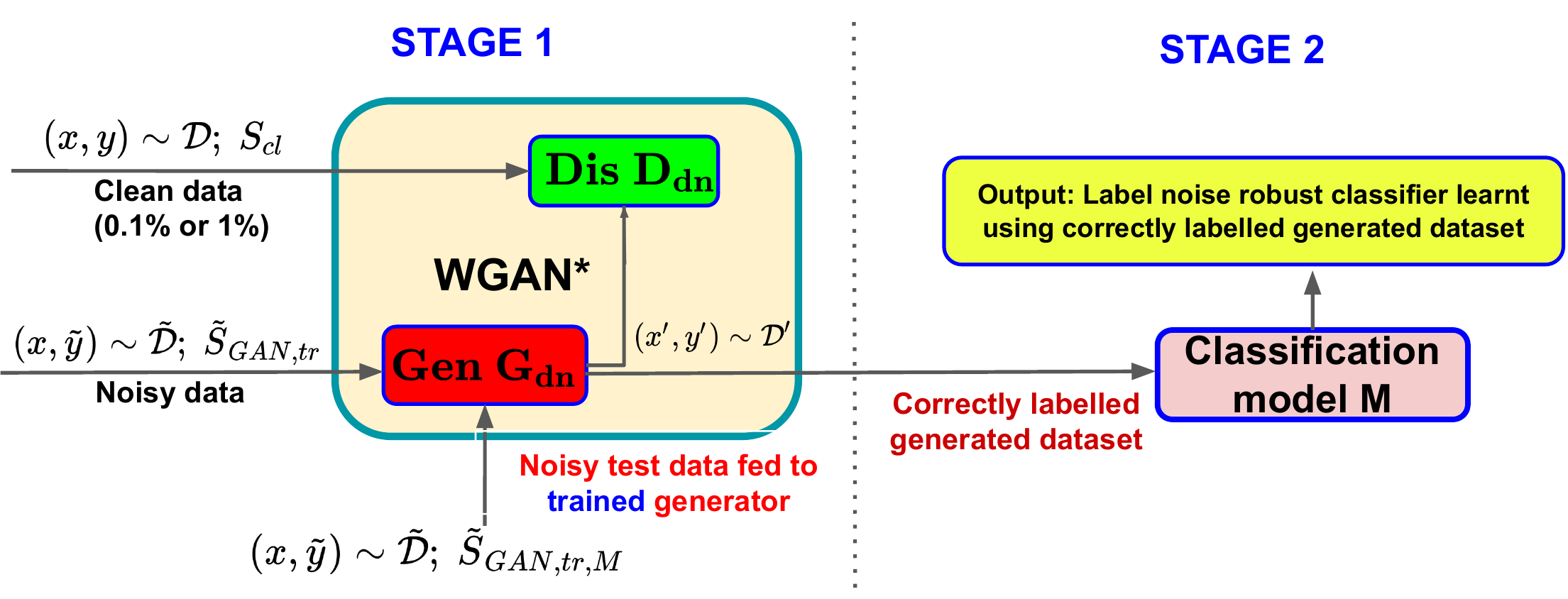}
	\caption{A schematic of our WGAN based schemes WGANXtraY(k) and WGANXtraYEntr(k). Stage 1 uses clean (from $\mathcal{D}$ distribution) and noisy (from $\tilde{\mathcal{D}}$ distribution) labelled data to generate correctly labelled (from $\mathcal{D}^{\prime}$) data. Stage 2 uses correctly labelled generated data to train classification model $M$.}
	\label{fig : schematic_WGAN*}
	\end{figure*}

	For imbalanced datasets, class conditional label noise can swap majority and minority class. However, due to the generative nature of our schemes, they are able to maintain original imbalance and hence doesn't require any special modification for learning. Our \textbf{main contributions} are as follows: 
	\begin{itemize}
		\item We propose novel Wasserstein GAN based schemes for learning from CCN corrupted binary classification data when a small set of clean labels can be procured.
		\begin{itemize}
			\item They do not require the knowledge (by estimation or cross-validation)  of noise rates and hence are robust to the datasets where an opponent/adversary is strategically corrupting the labels.
			\item They use representation and information theoretic ideas.
			\item In presence of very high noise (close to noise rate value of $0.5$), they show significant advantage over the existing label noise learning methods that use clean labels and those which do not require any knowledge about the noise rates.
		\end{itemize}
		\item Generative nature of the proposed schemes enable them to learn from class imbalanced and CCN corrupted data, without any modifications in the schemes.
		\item In addition to WGAN, we attempted to adapt PacGAN \cite{lin2018pacgan}, WGAN-GP \cite{gulrajani2017improved} and a variant inspired by VEEGAN \cite{srivastava2017veegan} for generating correctly labelled data but they did not work (see \ref{subsec: GAN_whichdint_work}); we adapt WGAN for our problem (which is known to address possible mode collapse).
		\item By using Friedman F test\footnote{The samples used by the Friedman F test are the accuracy or AM values for different noise rates for a fixed dataset; For a fixed dataset, null hypothesis is that the average performance (accuracy or AM values) is same for all schemes.} and Nemenyi posthoc tests\footnote{If, for a given dataset, the null hypothesis for Friedman test is rejected, then pairwise tests are performed to check if there is a statistically significant difference between the accuracy or AM values of the schemes in the considered pair.}, we demonstrated that WGAN based schemes lead to statistically significant improvement of accuracy or AM values over the loss function based discriminative approaches, viz., GLC, GCE, LDMI.
	\end{itemize}
% 	To statistically verify our empirical claims, we performed Friedman F test and Nemenyi posthoc tests and observed that WGAN based schemes either lead to significant improvement over the loss function based discriminative approaches or are on par with them.
	A schematic of our schemes can be seen in Figure \ref{fig : schematic_WGAN*}.

	\subsection{Related work} \label{subsec: related_work}
	% In this section, we provide existing literature related to the various threads we have considered. 
	In the last decade, label noise problem has gained a lot of attention from researchers due to its prevalence in various real life situations. \cite{natarajan2013learning,ghosh2015making,van2015learning,patrini2016loss,liu2016classification,tripathi2019cost} focus on effect of label noise problems on non-deep classification schemes and provide solutions which either consider a label noise robust loss function or modifies the loss function to make it robust. 
	% A probabilistic approach is proposed by \cite{zhao2018classification} where the noisy data is grouped into various states of a Markov chain without estimating the noise rate but shown only for not very high dimensional datasets. 
	% Approximate stationary distribution of this Markov chain is then used to get the final classification model \textcolor{red}{to write the last one more clearly to avoid incorrect comparison}.
	For deep learning schemes, \cite{patrini2017making} propose two algorithms called forward and backward loss correction to learn from label noise corrupted data, \cite{ghosh2017robust_deep} identify label noise robust loss functions to be used by neural networks and \cite{DBLP:journals/ml/MenonRN18} provide consistency results when the noise is instance dependent. \cite{han2018co-teaching} propose a co-teaching approach using two communicating neural networks. \cite{chen2019understanding} propose a relation between noisy data test accuracy and noise matrix. The above schemes which modify the loss function either require the knowledge of noise rates or estimate them.
	% After understanding the basic difference in working of deep neural networks with clean labels and noisy labels, \cite{ma2018dimensionality} provides a dimensionality driven learning algorithm. Even though the later scheme doesn't require noise estimates, it was demonstrated only on SLN induced datasets.
	\cite{amid2019robustBitempered} propose a theoretically sound loss function called bi-tempered loss function that is a non-convex generalization of logistic loss and requires tuning two temperature parameters. However, it is shown to be robust only for SLN case. \cite{zhang2018generalized} propose a loss function, viz., generalized cross entropy loss (GCE) that is shown to be asymptotically robust to SLN uniformly and CCN under some conditions. Also, \cite{xu2019LDMI} propose a loss function, viz., $L_{DMI}$ that is  invariant to the type and level of noise. We compared our schemes with the above two methods and observed that at high CCN rates our schemes have higher accuracies when the final classifier is learnt using a multi-layer perceptron. We recently came across the work of \cite{kim2019nlnl} and \cite{wang2019symmetric} and they are observed to be performing better than GCE. However, the negative learning based approach \cite{kim2019nlnl} is observed to be non-converging at high noise rates and the robustness of symmetric cross entropy \cite{wang2019symmetric} is approximate and holds uniformly only for SLN but not CCN.
	
	Of late, a reasonable assumption of access to a small set of clean labels during training the model from noisy labelled data is being used
	\cite{li2017learningDistillation,vahdat2017toward,veit2017learning,ren2018learning,hendrycks2018using}. The clean label set is used to estimate the noise rates in all the above schemes. \cite{hendrycks2018using} show that their scheme is better than others mentioned before. Hence, we compare our scheme to Gold Loss Correction (GLC) proposed by \cite{hendrycks2018using} and demonstrate that without even estimating the noise rates, our schemes 
	% is comparable to GLC at low noise rates and 
	lead to significant improvement over GLC at very high noise rates.
	
	\cite{thekumparampil2018robustness} and \cite{kaneko2019label} study label noise problem for conditional GAN (cGAN) and Auxiliary classifier GAN (ACGAN). Former work proposes two schemes, viz., RCGAN, which assumes knowing true noise matrix and RCGAN-U which estimates the noise matrix to make cGAN robust to label noise. \cite{kaneko2019label} also propose two schemes, viz., rACGAN and rcGAN where the discriminator is modified by including a noise transition model. We would like to emphasize that the above mentioned problem is different as label noise is affecting the sample generation quality and related aspects. However, in our problem label noise is affecting the classification performance. To summarize, our scheme differs from the aforementioned works in the context of (1) the task at hand, i.e., sample generation task vs classification task (2) noise rate, i.e.,  estimation vs non-estimation (3) the type of noise, i.e., uniform flipping and asymmetric multi-class noise (reduces to SLN in case of binary class) vs. CCN (4) our schemes' implicit capability of handling class imbalance in addition to CCN.
	
	\textbf{Organization} We introduce the label noise problem and GAN framework in Section \ref{subsec: label_noise_into} and \ref{subsec: GAN_int}. The main result on WGAN based schemes is provided in Section \ref{subsec: ourscheme} along with the theoretical results and some insights about which GAN architecture didn't work and why WGAN based schemes work in imbalanced data setup too. We demonstrate the performance of our schemes and compare them to existing schemes in Section \ref{sec: exper}. We conclude with discussion in Section \ref{sec: disc}.

	\section{GANs for generating clean labelled data} \label{sec: main_WGAN_gen}
	In this section, we first briefly describe label noise problem and basics of GANs. Then, we show which variants of GANs can be adapted for generating correctly (clean) labelled data as per true data distribution. We propose two schemes based on representational changes and information theoretic changes in objective function which leads to better results in final classification model. Finally, we explain why our schemes work well without any modification, in the presence of label noise and class imbalance.
	
	\subsection{Label noise problem} \label{subsec: label_noise_into}
	Let $\mathcal{D}$ be the joint distribution over $\mathbf{X}\times Y$ with $\mathbf{X} \in \mathcal{X} \subseteq \mathbb{R}^{n}$ and $Y\in \mathcal{Y} = \{-1,1\}.$ Let the in-class probability 
	% and class marginal 
	on $\mathcal{D}$ be denoted by $\eta(\mathbf{x}):=P(Y=1|\mathbf{x})$. 
	% and $\pi := P(Y=1)$.
	% Let the decision function be $f:\mathbf{X}\mapsto \mathbb{R}$, hypothesis class of all measurable functions be $\mathcal{H}$.
	% and class of linear hypothesis be $\mathcal{H}_{lin}= \{ (\mathbf{w},b), \mathbf{w}\in \mathbb{R}^n, b \in \mathbf{R}: \Vert \mathbf{w} \Vert_2 \leq W\}.$
	Let $\tilde{\mathcal{D}}$ denote the distribution on $\mathbf{X}\times \tilde{Y}$ obtained by inducing noise to $\mathcal{D}$ with $\tilde{Y} \in \tilde{\mathcal{Y}} = \{-1,1\}$. The corrupted sample is $\tilde{S} = \{(\mathbf{x}_1,\tilde{y}_1),\ldots,(\mathbf{x}_m,\tilde{y}_m)\} \sim \tilde{\mathcal{D}}^m.$ The label noise is class dependent and the noise rates are  defined as $\rho_+ := P(\tilde{Y}=-1|Y=1, \mathbf{X}=\mathbf{x})$, $\rho_- := P(\tilde{Y}=1|Y=-1, \mathbf{X}=\mathbf{x})$. This model is referred to Class Conditional Noise (CCN) model. In such cases, the corrupted in-class probability is $\tilde{\eta}(\mathbf{x}) := P(\tilde{Y}=1|\mathbf{x}) = (1-\rho_+ - \rho_-)\eta(\mathbf{x}) + \rho_-$.  If $\rho_+ = \rho_- = \rho$, then the model is known as Symmetric Label Noise (SLN) model.
	% and the corrupted class marginal is $\tilde{\pi}:= P(\tilde{Y}=1) = (1-\rho_+ - \rho_-)\pi+\rho_-.$
	Most existing solutions in literature for label noise problems either cross-validate over the noise rates or estimate them. Since, estimating noise rate is a density estimation problem (a hard problem), an alternative and clean solution is the one which doesn't require any knowledge (estimation or cross-validation) about the noise rates. %\textcolor{blue}{this in intro??}
	
	\subsection{Generative Adversarial Networks (GANs)} \label{subsec: GAN_int}
	GANs \cite{goodfellow2014generative} are defined by a two player  min-max game between a discriminator and a generator with the objective that after training the generator starts generating samples from the true data distribution. The generator is a neural network that uses samples from a simple prior distribution like Gaussian or uniform to generate samples from true data distribution. The discriminator is also a neural network which is fed with samples from true data distribution (real) and samples generated by the generator (fake) and its job is to  distinguish real samples from fake ones and provide feedback to the generator. Mathematically, the objective function for the two players can be described as a min-max optimization problem given below:
	\begin{equation} \label{eq: GAN_gen_obj}
	\min\limits_{G}\max\limits_{D} \mathbb{E}_{p_{\mathcal{D}}(\mathbf{x})}[f(D(\mathbf{x}))] + \mathbb{E}_{p(\mathbf{z})}[f(-D(G(\mathbf{z})))]
	\end{equation}
	where $G : \mathcal{Z} \mapsto \mathcal{X}$ is the generator network, which maps from the latent space (prior) $\mathcal{Z}$ to the input space $\mathcal{X}$; $D: \mathcal{X} \mapsto \mathbb{R}$ is the discriminator network, which maps from the input space to the decision space whether the sample is real or fake; and $f$ is concave function. $p(\mathbf{z})$ and $p_{\mathcal{D}}(\mathbf{x})$ are the density functions of the random variables in $\mathcal{Z}$ and $\mathcal{X}$ respectively. The traditional GAN formulation can be recovered with $f(\mathbf{x}) = -\log(1+\exp(-\mathbf{x}))$ and WGAN \cite{arjovsky2017wasserstein} with $f(\mathbf{x}) = \mathbf{x}$. The two major issues with GANs are of mode collapse and instability of the iterates from the optimization method while training.  In our context, mode collapse is a serious issue because samples containing only one mode (class) will lead to a biased and extreme classifier. Instability would have adverse  effects as in label noise models the performance of a scheme is evaluated by averaging over various random trials. Instability in GAN training would induce a lot of variations in final performance across various trials of the classification model trained on generated samples. In this paper, we use WGAN as it is known to be more stable while training and less prone to the problem of mode collapse. We would like to emphasize that conventionally GANs are generative models and are generally not used for classification. Hence, for them the role of labels is minimal.

	\subsection{WGAN for generating correctly labelled data} \label{subsec: ourscheme}
	Due to label noise problem, we have to deal with two data distributions: clean $\mathcal{D}$ and label noise corrupted $\tilde{\mathcal{D}}$ with one of marginals $\mathbf{X}$ common and the other marginal $Y$ and $\tilde{Y}$ different. Usually, we only have access to corrupted sample $\tilde{S}$. However, these days a small set of clean labels is easily procurable \cite{hendrycks2018using,ren2018learning}. We also leverage this fact and use a small number ($m_{cl}$) of data points with clean labels, say set $S_{cl} = \{(\mathbf{x}_1,{y}_1),\ldots,(\mathbf{x}_{m_{cl}},{y}_{m_{cl}})\} \sim {\mathcal{D}}^{m_{cl}}$. Now, consider the framework where the samples (with clean labels) from $\mathcal{D}$ are fed to the discriminator in place of the data points from true data distribution and the samples (with noisy labels) from $\tilde{\mathcal{D}}$ are fed to the generator in place of the sample points $\mathbf{z}$. The goal of the generator is to generate samples (with clean labels) from a distribution $\mathcal{D}'$ which is as close as possible to $\mathcal{D}$. Using this idea, we propose 2 schemes, WGANXtraY(k) and WGANXtraYEntr(k) described in following subsections.
	
	A natural thought would be that the job of generator has been simplified as $\tilde{\mathcal{D}}$ and $\mathcal{D}$ are already very close as marginals of $\mathbf{X}$ are same and transforming one to another shouldn't be difficult. However, the reality is quite counter intuitive due to the adversarial nature of WGANs. We observed that farther the two distributions are ($\mathcal{D}$ and $ \mathcal{Z}$ in WGANs or $\mathcal{D}$ and $\tilde{\mathcal{D}}$ in our case), the better is the performance of WGANs for generating samples from $\mathcal{D}$. In our case, we can quantify whether the distributions are far or not by computing the KL divergence between $\mathcal{D}$ and $\tilde{\mathcal{D}}$. With this context, we first show in Lemma \ref{lem: KL_SLN } that when $\tilde{\mathcal{D}}$ is SLN corrupted, the KL divergence between $\mathcal{D}$ and $\tilde{\mathcal{D}}$ is an increasing function of noise rate $\rho$. A proof is available in \ref{subsec: KL_SLN_proof}.
	
	\begin{lemma} \label{lem: KL_SLN }
		Let $\mathcal{D}$ and $\tilde{\mathcal{D}}$ be clean distribution and symmetric label noise corrupted distribution with noise rate $\rho$  respectively. Then, the KL divergence between $\mathcal{D}$ and $\tilde{\mathcal{D}}$ given as follows:
		\begin{small}
			\begin{eqnarray} \nonumber
			KL[\mathcal{D}\Vert \tilde{\mathcal{D}}]_{SLN} = \mathbb{E}_{\mathbf{X}}\left[ -\eta(\mathbf{x}) \log \left(1+ \rho \left(\frac{1-2\eta(\mathbf{x})}{\eta(\mathbf{x})}\right)\right) \right. \\ \label{eq: KL_SLN}
			- \left. (1-\eta(\mathbf{x}))\log \left(1- \rho \left(\frac{1-2\eta(\mathbf{x})}{\eta(\mathbf{x})}\right)\right)\right],
			\end{eqnarray}
		\end{small}
		is an increasing function of $\rho \in (0,0.5)$.
	\end{lemma}
	If one uses the observation that higher the adversarial relation (in terms of divergence between $\mathcal{D}$ and $\tilde{\mathcal{D}}$), better is the performance of WGAN, then, Lemma \ref{lem: KL_SLN } implies that performance of our WGAN based scheme for generating correctly labelled data will improve as the noise rates increase. Since, we do not have a measure to see the absolute improvement, we consider the improvement relative to the existing method of GLC from \cite{hendrycks2018using}. And, indeed, our claim is true as verified empirically for SLN corrupted 3 binary versions of MNIST dataset and 1 binary version of Fashion-MNIST dataset where for noise rate $\rho \in (0.45,0.5)$ our scheme outperforms GLC. Details are available in Table \ref{tab: SLN_incr_rho_inc_per1} and \ref{tab: SLN_incr_rho_inc_per2} of  \ref{subsec: exp_SLN_rho_inc}.
	
	Next, similarly, for the case of CCN, we would like to argue that our WGAN based schemes for generating correctly labelled data, work well at higher noise rate due to increase in adversarial nature via monotonicity of $KL[\mathcal{D}\Vert \tilde{\mathcal{D}}]_{CCN}$. However, the monotonicity in case of CCN is not straight forward because we would have to verify the increasing nature of KL divergence between $\mathcal{D}$ and $\tilde{\mathcal{D}}$, $KL[\mathcal{D}\Vert \tilde{\mathcal{D}}]_{CCN}$ w.r.t. two variables, viz., $\rho_+, \rho_-$. Unfortunately, for the two ways, viz., projection based and total derivative approach, in which we considered monotonicity lead to a negative result. We formalize this in Lemma \ref{lem: KL_CCN} whose proof is available in  \ref{subsec: KL_CCN_proof}.
	
	% ************************* \\
	% However, this uniform monotonicity of SLN case doesn't hold for CCN corrupted distribution \textcolor{blue}{--why??} \textcolor{red}{the first derivative of the conditional clean and noisy distribution is not positive for all values of the noise rates}. Since, the monotonicity doesn't hold for a given $x$, it wont be holding for the expectation. Next, we obtain lower bounds on noise rates $\rho_+$ and $\rho_-$ for which the KL divergence between clean and CCN corrupted distribution is an increasing function w.r.t. noise rates. \textcolor{blue}{we will see the role of the total and directional (Geatuex and Frechet, i think) derivatives here??}
	
	% \textcolor{blue}{is total derivative notion useful?}
	% \textcolor{red}{total derivative approach also leads to the same conditions on the values of noise rates along with the in-class probability. \\
	% Checked by substituting the high noise rates, in the derivative values but they are of opposite sign and hence do not get satisfied simultaneously.} \\
	% % http://www.math.ttu.edu/~klong/5311-spr09/diff.pdf
	% % \textcolor{red}{bounds on noise rates for CCN case}
	% % \textcolor{red}{show using the examples of Synthetic case in SLN noise and showing that KL between clean and corrupted increases with  increase in noise rate}. 
	% %\textcolor{blue}{$\rho_+ = \max(*, *, *)$ gives a lower bound?}
	% ***************
	
	\begin{lemma} \label{lem: KL_CCN}
		Let $\mathcal{D}$ and $\tilde{\mathcal{D}}$ be clean distribution and class conditional label noise corrupted distribution with noise rates $\rho_+$ and $\rho_-$. Then, the KL divergence between $\mathcal{D}$ and $\tilde{\mathcal{D}}$ given as follows:
		\begin{small}
			\begin{eqnarray*} \nonumber
				KL[\mathcal{D}\Vert \tilde{\mathcal{D}}]_{CCN} = \mathbb{E}_{\mathbf{X}}\left[ -\eta(\mathbf{x}) \log \left(1-  \rho_+ -\rho_- \left( 1 - \frac{1}{\eta(\mathbf{x})}\right)\right) \right. \\ 
				- \left. (1-\eta(\mathbf{x}))\log \left(1- \rho_- + \rho_+ \left(\frac{\eta(\mathbf{x})}{1 - \eta(\mathbf{x})}\right)\right)\right],
			\end{eqnarray*}
		\end{small}
		is not an increasing function of $\rho_+$ and $\rho_-$. 
		% when:
		% \begin{small}
		% \begin{eqnarray} \label{eq: rho_cond_CCN}
		%     \rho_+ \geq \max\limits_{\mathbf{x}} \bigg\{ \frac{\rho_-(1- \eta(\mathbf{x}))}{\eta(\mathbf{x})} \bigg\}, ~
		%     \rho_- \geq \max\limits_{\mathbf{x}} \bigg\{ \frac{\rho_+ \eta(\mathbf{x})}{(1-\eta(\mathbf{x}))} \bigg\}.
		% \end{eqnarray}
		% \end{small}
		% \textcolor{red}{the above conditions simultaneously hold only when $\eta(\mathbf{x})=0.5$ and the noise rates are equal. This makes the condition very restrictive.} 
		% \textcolor{blue}{may be some other approach?}
	\end{lemma}

	Since, our claim is for high noise rates only, we can use the continuity arguments from SLN case to show that in the considered regime of high noise rates, the adversarial relation plays a role in good performance of our scheme even when the noise is of CCN type. Based on our empirical experience (Table \ref{tab: SLN_incr_rho_inc_per1} and \ref{tab: SLN_incr_rho_inc_per2} of \ref{subsec: exp_SLN_rho_inc}), we observe that the following inequality holds in most of the cases:
	$$ h(S, \tilde{S})_{\rho, met} > h(S, \tilde{S})_{\rho, GLC},$$
	where $\rho$ is the symmetric label noise rate, $h(\cdot)$ is a performance metric that can be taken as the test accuracy and $met \in \{ WGANXtraY(k), WGANXtraYEntr(k))\}$. Because of the strict inequality, if we make an $\epsilon$ perturbation to $\rho \in (0.45,0.49)$ to obtain $\rho_+ \in (0.45,0.49)$ and $\rho_- \in (0.45, 0.49)$, we get the following relation,
	$$ h(S, \tilde{S})_{\rho_+, \rho_-, met} > h(S, \tilde{S})_{\rho_+, \rho_-, GLC}.$$
	This implies that at high noise rates, a large value of $KL[\mathcal{D} \Vert \tilde{\mathcal{D}}]_{CCN}$ leads to better performance of our schemes relative to GLC as verified empirically on 2 synthetic datasets and 10 real datasets in Figure \ref{fig: MNIST} and \ref{fig: Syn&FashionMNIST}.
	
	We would like to emphasize that our scheme is different from conventional WGAN in two ways: 1) data representation approach, i.e., the labels are appended in addition to the features and 2) the latent variable is not standard normal or uniform, instead it follows corrupted and hence \emph{unknown} data distribution $\tilde{\mathcal{D}}$. Also, unlike the existing work on learning with noisy labels, use of a generative model in label noise problem allows us to circumvent the problem of either estimating or tuning the noise rates. Even though \cite{ma2018dimensionality,zhao2018classification,tripathi2019cost} do not need the noise rates, the context is restricted either to symmetric label noise or non-high dimensional datasets. Next, we formally describe our scheme.
	\subsubsection{WGANY} \label{sssec: WGANY}
	Consider a WGANY (Y is due to use of labelled data) with generator $G_{dn}: \mathcal{X}\times \tilde{\mathcal{Y}} \mapsto \mathcal{X}\times \mathcal{Y} $ and discriminator (critic) $D_{dn}: \mathcal{X}\times \mathcal{Y} \mapsto \mathbb{R}$. If the generator models a distribution $\mathcal{D}'$, then the goal is to minimize the divergence between \emph{clean} distribution $\mathcal{D}$ and \emph{correctly labelled generated} distribution $\mathcal{D}'$. Hence, the WGANY objective function (adapted from \cite{gulrajani2017improved}) is given as follows:
	\begin{equation} \label{eq: WGAN_ourobjec}
	\min\limits_{G_{dn}}\max\limits_{D_{dn}} \mathop{\mathbb{E}}_{(\mathbf{x},y)\sim \mathcal{D}}[D_{dn}(\mathbf{x},y)] - \mathop{\mathbb{E}}_{(\mathbf{x}',y')\sim \mathcal{D}'}[D_{dn}(\mathbf{x}',y')],
	\end{equation}
	where $D_{dn}$ belongs to the set of $1$-Lipschitz functions and $\mathcal{D}'$ is model distribution implicitly defined by $(\mathbf{x}',y') := (G^{f}_{dn}(\mathbf{x},\tilde{y}), G^{l}_{dn}(\mathbf{x},\tilde{y})) =  G_{dn}(\mathbf{x},\tilde{y})$, with $(\mathbf{x},\tilde{y}) \sim \tilde{\mathcal{D}}$. Lipschitz constraint is enforced by clipping the weights of the discriminator to lie within a compact set $[-c,c]$ for some constant $c>0$. As the last dimension of the output, $(\mathbf{x}',y')$, of the generator $G_{dn}$ is supposed to be a label, we first convert the last dimension of the generator output to a probability value using sigmoid function. Now, the later can be interpreted as in-class probability $\eta(\mathbf{x}')$ and thresholded at 0.5 to get the label, i.e., $\eta(\mathbf{x}') >0.5 \Rightarrow y' = 1$ else $y' = -1$. 
	
	% We call this WGANY.
	
	As the objective function of WGAN is a strict adversarial divergence and we have just appended the label vector as an extra dimension, the convergence result of \cite{liu2017approximation} continues to hold for our scheme. However, there are other aspects that are peculiar to the use of WGANY for generating correctly labelled data. For a given corrupted data point $(\mathbf{x},\tilde{y})$ we expect our generator to either correct the label and output $(\mathbf{x},y)$ or synthesize a new feature vector along with its correct label $(\mathbf{x}',y')$. Since, the feature space is generally continuous, we believe that second event happens with probability 1. Now, the feature dimension $n$ being far greater than the label dimension of $1$ obstructs in learning the correct feature-label pair by the generator $G_{dn}$. We demonstrate this in Figure \ref{fig: MNIST_7-9_WGNAY_bad} and \ref{fig: Fashion_MNIST_7-9_WGNAY_bad} of  \ref{subsec: WGANY_bad_perf}. In addition, binary nature of $Y$ complicates this problem. To resolve this problem, we propose following two solutions. 
	% \textcolor{blue}{two solutions/alternatives??}
	
	\subsubsection{WGANXtraY(k)} \label{sssec: WGANXtraY}
	WGANXtraY($k$) stands for WGAN with extra $k$ dimensions consisting of scaled $Y$ values. Here, $k>2$ is taken to be an odd natural number to avoid ties. In other words, a $(n+1)$ dimensional data point $(\mathbf{x},y)$ would become a $(n+k)$ dimensional data point $(\mathbf{x},y,ly,\cdots,l(k-1)y)$ where $l>1$ can be chosen arbitrarily. We exploit the fact that for binary classification, sign of label $y$ is sufficient and scale the labels so that the label information is same, i.e., for a given $y \in \{-1,1\}$, $sign(y) = sign(l(k-1)y), ~k>2$. For example, if for a negatively labelled data point $(\mathbf{x},-1)$, we take $k = 3$ and $l = 5$, then the $(n+3)$ dimensional appended data point is $(\mathbf{x},-1,-5,-10)$. For new generated samples, unlike WGANY where sigmoid function is applied to only last dimension,  in WGANXtraY(k), the sigmoid transformation is applied to last $k$ dimensions individually. If $(k+1)/2$ dimensions out of last $k$ dimensions of generator output have probability values greater than $0.5$, then the last dimension is assigned the correct label as $y' =1$ else $y'=-1$.
	% \textcolor{blue}{ties need to be broken by some rule or ties do not occur, i.e., ties occur with prob zero?}

	\subsubsection{WGANXtraYEntr(k)} \label{sssec: WGANXtraYEnt}
	We require that the label $y'$ assigned by the generator to the generated sample $(\mathbf{x}',y')$ should be correct and not noisy. Inspired by the use of entropy while making class assignments as in \cite{springenberg2016iclr}, we attempt to satisfy this requirement by adding entropy terms to the generator and discriminator of WGANXtraY(k). We want the class distribution $p(y'|\mathbf{x}')$ conditioned on generated data point $\mathbf{x}'$ to be highly peaked, i.e., $G_{dn}(\mathbf{x},\tilde{y})$ should be certain of the label assigned. This can be achieved by minimizing the Shannon entropy $H[p(y'|\mathbf{x}')]$ for the sample $(\mathbf{x}',y')$, as we would want that for any draw from $\mathbf{x}'$, the conditional distribution should lead to same label $y'$ almost every time. The conditional entropy over the samples from the generator $G_{dn}$ can be written as the expectation of $H[p(y'|\mathbf{x}')]$ over the prior distribution $\tilde{\mathcal{D}}$, i.e., $\mathop{\mathbb{E}}_{(\mathbf{x},\tilde{y})\sim \tilde{\mathcal{D}}}[H[p(G^l_{dn}(\mathbf{x},\tilde{y})|G^f_{dn}(\mathbf{x},\tilde{y}))]]$ and its estimate is given below:
	% \begin{small}
	\begin{eqnarray} \nonumber
	\mathcal{L}_{G_{dn}, entr} =  \frac{1}{m_{g}}\sum\limits_{i=1}^{m_g}H[p(G^l_{dn}(\mathbf{x}_i,\tilde{y}_i)|G^f_{dn}(\mathbf{x}_i,\tilde{y}_i))] \\
	= \frac{1}{m_{g}}\sum\limits_{i=1}^{m_g}\left(- \sum\limits_{u \in \{-1,1\}} p(y'_i = u|\mathbf{x}'_i)\log p(y'_i = u|\mathbf{x}'_i)\right),
	\end{eqnarray}
	% \end{small}
	where $m_g$ is the number of independently drawn noisy samples $(\mathbf{x},\tilde{y})$ from corrupted training dataset $\tilde{S}$.
	%\textcolor{blue}{Eq 9 is an empirical estimate of LHS?} \textcolor{red}{Yes, the empirical estimate of the Shannon's entropy $H(\cdot)$}
	The conditional probability $p(y'|\mathbf{x}')$ can be obtained from the last $k$th  dimension of $G_{dn}(\mathbf{x},\tilde{y})$ before the final label is determined by using majority voting from last $k$ dimensions. Similarly, for the discriminator, we would want that $D_{dn}$ should decide that the samples from $\mathcal{D}$ are \emph{clean} and the generated samples $G_{dn}(\mathbf{x},\tilde{y})$  from model distribution $\mathcal{D}'$ are \emph{ correctly labelled generated (gen\_cor)} data points. To achieve this, it should have low entropy on the conditional distribution of \emph{clean} samples from $\mathcal{D}$ and high entropy on the conditional distribution of \emph{gen\_cor} samples from $\mathcal{D}'$. The former requirement can be satisfied by minimizing estimate of $E_{(\mathbf{x},y)\sim \mathcal{D}}[H[p(A|(\mathbf{x},y))]]$ given in Eq. (\ref{eq: Dis_entr_clean}) and the later can be satisfied by maximizing the estimate of $E_{(\mathbf{x},\tilde{y}) \sim \tilde{D}}[H[p(A|G_{dn}(\mathbf{x},\tilde{y}))]]$ given in Eq. (\ref{eq: Dis_entr_cleaned}).
	Here, $A$ denotes the type of data point fed in discriminator and takes value $a$ belonging to the set  $\{clean,~ gen\_cor\}$.
	% is either clean or  correctly labelled generated (gen\_cor).
	% \begin{scriptsize}
	\begin{eqnarray} \nonumber
	\mathcal{L}_{D_{dn}, entr}^{(1)} = \frac{1}{m_d}\sum\limits_{i=1}^{m_d}H[p(A|(\mathbf{x}_i,y_i))] \\ \label{eq: Dis_entr_clean}
	= \frac{1}{m_d}\sum\limits_{i=1}^{m_d} \big( -\sum_{a} p(a|(\mathbf{x}_i,y_i))\log p(a|(\mathbf{x}_i,y_i))  \big) \\ \nonumber
	% \end{eqnarray}
	% \begin{eqnarray} \nonumber
	\mathcal{L}_{D_{dn}, entr}^{(2)} = \frac{1}{m_d}\sum\limits_{i=1}^{m_d}H[p(A|(G_{dn}(\mathbf{x}_i,\tilde{y}_i)))] \\ \label{eq: Dis_entr_cleaned}
	= \frac{1}{m_d}\sum\limits_{i=1}^{m_d} \big( -\sum_{a} p(a|(G_{dn}(\mathbf{x}_i,\tilde{y}_i)))\log p(a|(G_{dn}(\mathbf{x}_i,\tilde{y}_i)))  \big),
	\end{eqnarray}
	% \end{scriptsize}
	where $m_d$ is the number of independently drawn samples from clean dataset $S_{cl}$ (for Eq. (\ref{eq: Dis_entr_clean})) and number of labelled generated data points obtained from generator (for Eq. (\ref{eq: Dis_entr_cleaned})). Thus, we want to minimize the following term:
	% \begin{small}
	\begin{eqnarray}
	\mathcal{L}_{D_{dn}, entr} = \mathcal{L}_{D_{dn}, entr}^{(1)} - \mathcal{L}_{D_{dn}, entr}^{(2)}.
	\end{eqnarray}
	% \end{small}
	Note that adding entropy maximization terms have been observed to be performing a regularization \cite{grandvalet2005semi}. From this perspective, WGANXtraYEntr(5) can be considered as an entropy regularized version of the WGANXtraY(5).
	
	\ref{alg: DenoisigGANs} presents the overall classification scheme to make predictions on clean test data  using a classifier learnt on CCN label corrupted training data  \emph{without using any knowledge about noise rates $\rho_+, \rho_-$}. We consider 4 parts of the datasets, $S_{cl}$ and $S_{te}$ are clean and $\tilde{S}_{GAN,tr}$ and $\tilde{S}_{G,tr,M}$ are noisy. $S_{cl}$ is a small set (usually $0.1\%$ or $1\%$ of training dataset) whose correct labels are procured for training $D_{dn}$ and $G_{dn}$ and we refer to it as gold fraction as in \cite{hendrycks2018using}. $\tilde{S}_{GAN,tr}$ is noisy dataset which is used to train $D_{dn}$ and $G_{dn}$. $\tilde{S}_{G,tr, M}$ is fed to the trained generator $G_{dn}$ to obtain correctly labelled data which in turn is used to train the final classification model $M$. $S_{te}$ is the test dataset used to evaluate the classification model $M$. The two schemes, WGANXtraY(k) and WGANXtraYEntr(k) differ in objective functions for $D_{dn}$ and $G_{dn}$ which is reflected in line number 5  and 10 of \ref{alg: DenoisigGANs}. We would like to point out that due to use of feature-label pair as one data point (entity) to be fed in discriminator, using the later as a classifier network is not straightforward. 
	
	\begin{algorithm}[!htbp]
		\renewcommand{\thealgorithm}{Algorithm WGAN*(k)}
		\floatname{algorithm}{}
		\caption{Schemes WGANXtraY(k) and WGANXtraYEntr(k) for learning from CCN corrupted data} %-- \textcolor{blue}{2 schemes?}} %ALG1}
		\label{alg: DenoisigGANs}
		\begin{algorithmic}[1]
			\Statex \hspace{-0.44cm}\textbf{Input:} Training data: ${S}_{cl}$, $\tilde{S}_{GAN,tr}$, $\tilde{S}_{G,tr, M}$; test data: $S_{te}$
			\Statex \hspace{-0.44cm}\textbf{Require:} Learning rate $\alpha$, Clipping parameter $c$, Batch size $m_b$, No. of critic iterations $n_c$, No. of iterations $n_{it}$, No. of appends $k$, Scaling number $l$, Initial critic parameters $w_0$, Initial generator parameters $\theta_0$.
			\Statex \hspace{-0.44cm}\textbf{Output:} Predicted labels for test data $S_{te}$.
			\For{$t = 0,\cdots, n_{it}$}
			\For{$t' = 0, \cdots, n_{c}$}
			\State Sample clean data $\{(\mathbf{x}_i,y_i)\}_{i=1}^{m_b}$ from $S_{cl}$; construct $\mathbf{d}_{i} := (\mathbf{x}_i,y_i,ly_i,\cdots,l(k-1)y_i), \forall i$.
			\State Sample noisy data $\{(\mathbf{x}_i,\tilde{y}_i)\}_{i=1}^{m_b}$ from $\tilde{S}_{GAN,tr}$; construct $\tilde{\mathbf{d}}_{i} := (\mathbf{x}_i,\tilde{y}_i,l\tilde{y}_i,\cdots,l(k-1)\tilde{y}_i), \forall i$.
			\State WGANXtraY(k): {\small $g_w \leftarrow \nabla_{w}[\frac{1}{m_b}\sum\limits_{i=1}^{m_b}D_{dn}(\mathbf{d}_{i}) - \frac{1}{m_b}\sum\limits_{i=1}^{m_b}D_{dn}(G_{dn}(\tilde{\mathbf{d}}_{i}))]$} 
			\Statex \textbf{OR} WGANXtraYEntr(k): $g_{w} \leftarrow g_{w} - \nabla_{w}\mathcal{L}_{D_{dn},entr} $
			\State $w \leftarrow w + \alpha \cdot$ RMSProp$(w,g_w)$
			\State $w \leftarrow $ clip$(w,-c,c)$
			\EndFor
			\State Sample noisy data $\{(\mathbf{x}_i,\tilde{y}_i)\}_{i=1}^{m_b}$ from $\tilde{S}_{GAN,tr}$; construct $\tilde{\mathbf{d}}_{i} := (\mathbf{x}_i,\tilde{y}_i,l\tilde{y}_i,\cdots,l(k-1)\tilde{y}_i), \forall i$.
			\State WGANXtraY(k): \begin{small} $g_{\theta} \leftarrow - \nabla_{\theta} \frac{1}{m_b}\sum\limits_{i=1}^{m_b}D_{dn}(G_{dn}(\tilde{\mathbf{d}}_i))$ \end{small}
			\Statex \textbf{OR} WGANXtraYEntr(k): $g_{\theta} \leftarrow g_{\theta}  + \nabla_{\theta}\mathcal{L}_{G_{dn},entr}$
			\State $\theta \leftarrow \theta - \alpha\cdot$RMSProp$(\theta, g_{\theta})$
			\EndFor 
			\State Sample noisy data $\{(\mathbf{x}_i,\tilde{y}_i)\}_{i=1}^{|\tilde{S}_{G, tr, M}|}$ from $\tilde{S}_{G, tr, M}$; construct $\tilde{\mathbf{d}}_{i} := (\mathbf{x}_i,\tilde{y}_i,l\tilde{y}_i,\cdots,l(k-1)\tilde{y}_i), \forall i$. 
			\State Get correctly labelled generated samples $(\mathbf{x}_i',y'_i) = G_{dn}(\tilde{\mathbf{d}_i}), \forall i = 1,\ldots,|\tilde{S}_{G, tr, M}|$, where $y'_i$ is obtained by majority voting in the last $k$ dimensions of the generator output.
			\State Train classification model $M$ on $\{(\mathbf{x}'_i,y'_i)\}_{i=1}^{|\tilde{S}_{G, tr, M}|}$.
			\State Obtain predictions for test data $S_{te}$ using $M$.
		\end{algorithmic}
	\end{algorithm}
	
	\subsubsection{Good GANs for generating samples need not be good for generating correctly labelled data} \label{sssec: goodGAN_not} The GAN literature is growing continuously with new architectures, new objective functions, new priors, new regularization techniques and a lot of theoretical analysis too. We explored the recent PacGAN \cite{lin2018pacgan} which uses the fact that there is a fundamental  connection between packing samples and mode collapse. Unfortunately, it did not help for generating correctly labelled data as the labels $y$ got packed along with the feature vectors. This lead to the loss of label information content in between the features and hence no diversity in samples. As VEEGAN \cite{srivastava2017veegan} is known for reducing mode collapse, we used a similar idea by adding a reconstruction loss to the generator. This reconstruction loss is the mean squared error between the feature vector of corrupted sample $(\mathbf{x},\tilde{y})$ and feature vector of the generated sample $(\mathbf{x}',y')$. The issue of trade-off between generating new sample such that it is close to the sample fed to the generator and making sure the label is correct, doesn't let the generator learn the original clean distribution $\mathcal{D}$. In addition, the gradient penalty proposed by \cite{gulrajani2017improved} for WGAN also did not led to much improvement for the task of generating correctly labelled data. These observations are based on experiments on binary version of MNIST \cite{Mnistlecun1998gradient} and Fashion MNIST \cite{xiao2017fashion} datasets. Details in Table \ref{tab: Mnist7-9_GAN_not_work}, \ref{tab: Mnist1-7_GAN_not_work}, \ref{tab: Fashion_Mnist7-9_GAN_not_work} and \ref{tab: Fashion_Mnist2-3_GAN_not_work} of  \ref{subsec: GAN_whichdint_work}.
	% . \textcolor{blue}{need to illustrate these in supplementary material ?? }
	
	\subsubsection{WGANs based schemes for noisy imbalanced data} \label{sssec: WGAN_imb} 
	For imbalanced and symmetric label noise ($\rho < 0.5$) corrupted data, majority (minority) class continues to be in majority (minority) (Lemma 3 \cite{tripathi2019cost}). However, this result doesn't hold for CCN corrupted data, which makes learning from such data a difficult problem. We observe that our WGAN based schemes for generating correctly labelled data need no modification even when the data is imbalanced (Figure \ref{fig: MNIST_imb}). This is because our schemes attempt to learn the distribution with original imbalance ratio even though the available corrupted data has different imbalance ratio. This advantage makes our scheme suitable for learning from imbalanced noisy data without any knowledge of imbalance ratio or $\rho_+, \rho_-$. Two algorithms to deal with the noisy imbalanced dataset problem proposed in \cite{natarajan2017cost} require tuning over the asymmetric classification cost $\alpha$ and cross validation over noise rates. Also, radial basis oversampling proposed in \cite{koziarski2019radial} only demonstrates their results when the majority class doesn't flip to minority (or vice versa) and hence doesn't completely deal with the complexity of imbalanced label noise learning. \cite{tripathi2019cost} provide solution for the label noise (but only SLN) and imbalance problem without requiring the noise rate.
	% but the noise type is only SLN.
	% \textcolor{blue}{to mention carefully, that PAKDD also doesn't need noise rates, but, class imbalance problem doesn't occur as problem considered is SLN one??}
	
	% \textcolor{blue}{write code in  -- as part of RR?}
	
	% \textcolor{blue}{as overview of things to do -- cost, cases where the rates are quite separated, e,g. .2 and .46, some others that are already listed?}
	
	% \subsubsection{Learning from label noise as a semi-supervised problem} \textcolor{red}{to include if the results from semi-supervised results are good} The corrupted data available for training has only labels as noisy; the feature vectors are clean. If one ignores the labels, then problem of learning from label noise corrupted data can be posed as an unsupervised learning problem. Since, we know the number of classes, a major task of deciding the number of clusters (classes) is taken care of. Also, as in the discussion till now if one can procure clean labels for a small set of data, then keeping only clean gold fraction labelled and rest data unlabeled would lead to a semi-supervised learning problem. According to the best of our knowledge, label noise learning problem has not been viewed from lens of unsupervised or semi-supervised learning. We use a theoretically sound technique called CatGAN \cite{springenberg2016iclr} for solving the unsupervised and semi-supervised problems for the binary case. \textcolor{red}{Observations}
	
	\section{Experiments} \label{sec: exper}
	In this section, we empirically verify the performance of our schemes for label noise robust classification. Exact details about synthetic data generation scheme, sampling imbalanced data, dataset sizes and neural network architecture are provided in  \ref{subsec: data_arc_details}.\\
	\textbf{Datasets} We consider 2 synthetic datasets (100 and 300 dimensional), binary class balanced versions for 6 pairs of MNIST \cite{Mnistlecun1998gradient} and 4 pairs of Fashion MNIST \cite{xiao2017fashion} datasets to demonstrate effectiveness of our scheme at very high (near 0.5) noise rates. We also consider 6 binary imbalanced versions of MNIST datasets using different imbalance ratios. In the binary digit pair $(a,b)$, $a$ corresponds to positive class and $b$ corresponds to negative class except for the case when either $a$ or $b$ is 0. In the later case, $0$ corresponds to negative class and the other digit to positive class. To account for randomness in the flips to simulate a given noise rate, we repeat each experiment 5 times, with independent corruptions of the dataset for same noise ($\rho_+, \rho_-$) setting. The binary train and test set are constructed from respective multi-class train and test datasets. The train dataset is then split into $0.1\%$ ({10-12 and 4-6 total clean data points in balanced and imbalanced case respectively}) or $1\%$ ({100-120 and 50-60 total clean data points in balanced and imbalanced case respectively}) gold fraction, $\sim 84\%$ to train WGAN based scheme and $\sim 15\%$ for training final classification model M. In every trial, the last two partitions are induced with class conditional noise. \\
	\textbf{Framework} The generator $G_{dn}$ and $D_{dn}$ are  multilayer perceptron (MLP) with 4 and 3 layers respectively. We trained a 4 layered MLP and {a 3 layered Convolutional Neural Network (CNN) with dropout after second convolution layer as the final classification model M.} The parameters used in \ref{alg: DenoisigGANs} are as follows: $\alpha = 10^{-3}$, $c = 0.01$, $n_c = 5$, $m_b = 64$. The number of iterations $n_{it} = 500 (1000)$ for WGANXtraY(k) (WGANXtraYEntr(k)), if not mentioned otherwise. {To determine the influence of $l$ on the performance of our scheme, we tried l= 2, 5 and 8 and the accuracies were within 2\%  of each other with no single winner. In addition to k=5, we tried k=10 but didn't observe any improvement. Hence, in our experiments, we have used $k=5$ and $l=5$.}	We consider 15 pairs of noise rates $(\rho_+, \rho_-)$ out of which 11 cases have both noise rates more than 0.4 (including some with values $\geq 0.5$) which we treat as very high noise rates.
	% For experimentation, we also consider 4 pairs where noise rates are $\geq 0.5$.
	We use accuracy of the final classification model $M$  on test data as an evaluation metric for balanced datasets. For imbalanced datasets, we use arithmetic mean (AM) of true positive rate (TPR) and true negative rate (TNR) to evaluate the quality of final predictions from $M$.
	For low to moderate noise rates, not satisfactory performance of our scheme (\ref{subsec: ourscheme_low}) further asserts the claim of Lemma 2 that performance of our schemes (with MLP based model M) improve with increase in noise rates.  
	
\begin{figure*}[!h]
		\centering
		\includegraphics[width =1.01\textwidth]{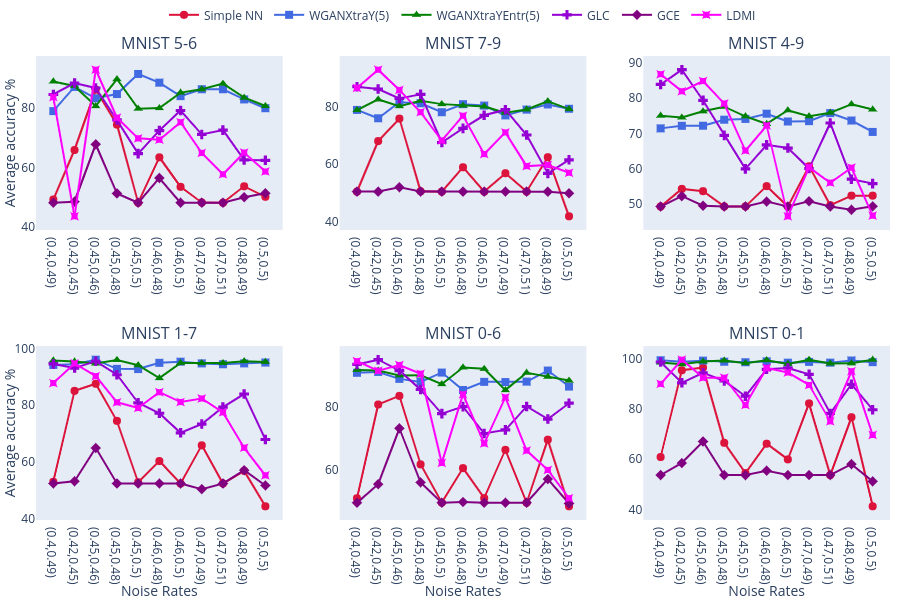}
		\caption{Average accuracy plots for various binary class MNIST datasets across 5 trials when the final classification model M is a MLP. This plot demonstrates that even with only $0.1\%$ clean data points, at high noise rates close to 0.5, our scheme significantly improves over the state of the methods like GLC, GCE and LDMI in most cases.}
		\label{fig: MNIST}
	\end{figure*}

\begin{figure*}[!h]
		\centering
		\includegraphics[width =1.01\textwidth]{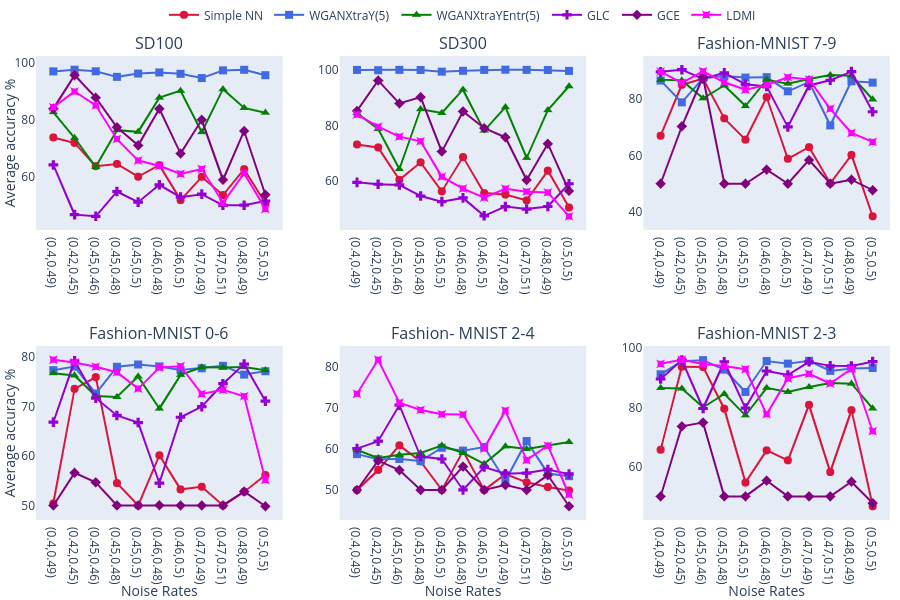}
		\caption{Average accuracy plots for Synthetic datasets SD100 and SD300 and binary class Fashion-MNIST datasets across 5 trials when the final classification model M is a MLP. This plot demonstrates that even with only $0.1\%$ clean data points, at high noise rates close to 0.5, WGANXtraY(5) dominates WGANXtraYEntr(5) and other stat of the art on SD100 and SD300 (Column 1and 2 from left). For Fashion-MNIST 0-6 (Tshirt and Shirt) and Fashion-MNIST 7-9 (Sneakers and Boots), our schemes achieve higher accuracy values than GLC and LDMI in more than half noise rates cases. Fashion-MNIST 2-4 (Pullover and Coat) inherently has low accuracy and adding label noise further decreases it. For Fashion-MNIST 2-3 (Pullover and Dress) our schemes are either better or on par with GLC and LDMI.}
		\label{fig: Syn&FashionMNIST}
	\end{figure*}

\begin{figure*}[!h]
		\centering
		\includegraphics[width = 1.01\textwidth]{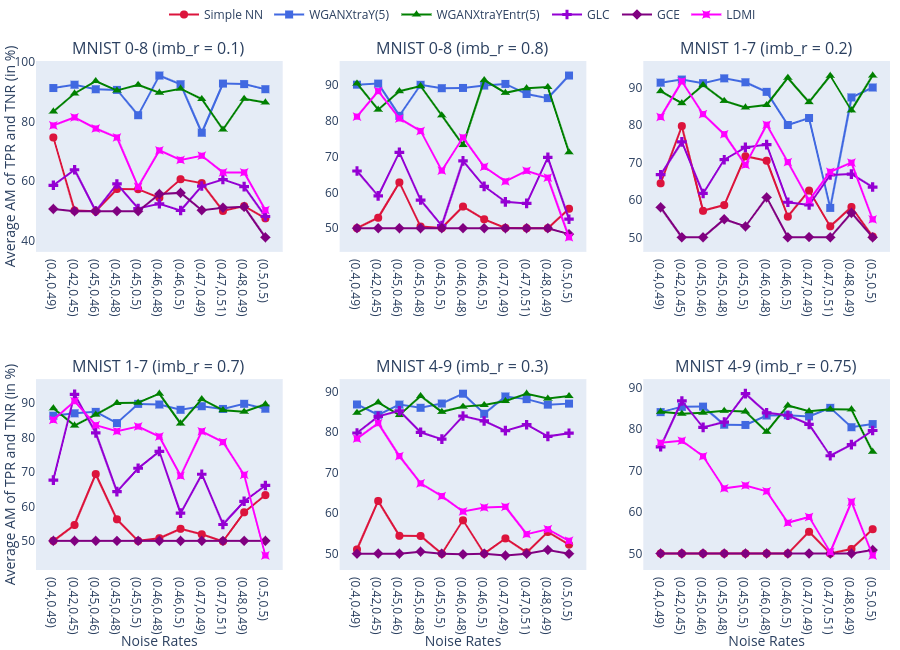}
		\caption{Average AM (Arithmetic Mean of TPR and TNR) value plots for various imbalanced binary class MNIST datasets across 5 trials  when the final classification model M is a MLP. ``imb\_r" denotes the fraction of positive data points. MNIST 0-8 and 1-7 use $0.1\%$ clean data and MNIST 4-9 uses $1\%$ clean data. AM values from SimpleNN on clean imbalanced datasets (from left to right) are $96.3\%, 99.15\%, 98.89\%, 98.30\%, 94.86\%, 96.36\%$. Adverse effects of class conditional noise with imbalanced data on SimpleNN (red line) and GCE (purple line) can be seen by huge drop from aforementioned AM values to $\sim 50\%$ AM value.}
		\label{fig: MNIST_imb}
	\end{figure*}

\begin{figure*}[!ht]
		\centering
		\includegraphics[width =1.01\textwidth]{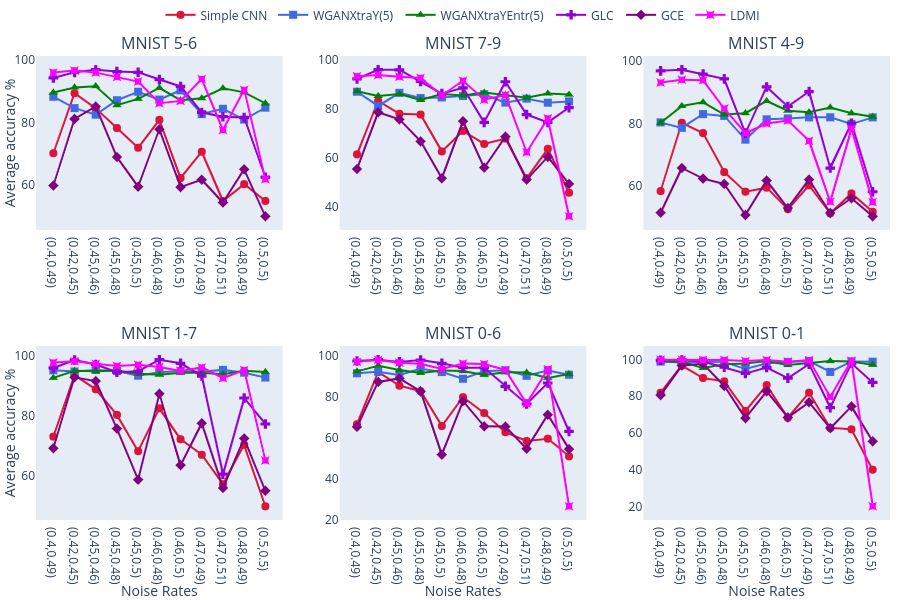}
		\caption{Average accuracy plots for various binary class MNIST datasets across 5 trials when the final classification model M is a CNN. WGAN based schemes, GLC and LDMI are observed to have accuracies within 5\% of each other in most of the cases. However, GCE is no better than using a simple CNN for high noise rates. Also, unlike other schemes WGAN based schemes have very low variation across noise rates.}
		\label{fig: MNIST_CNN}
	\end{figure*}

\begin{figure*}[!ht]
		\centering
		\includegraphics[width =1.01\textwidth]{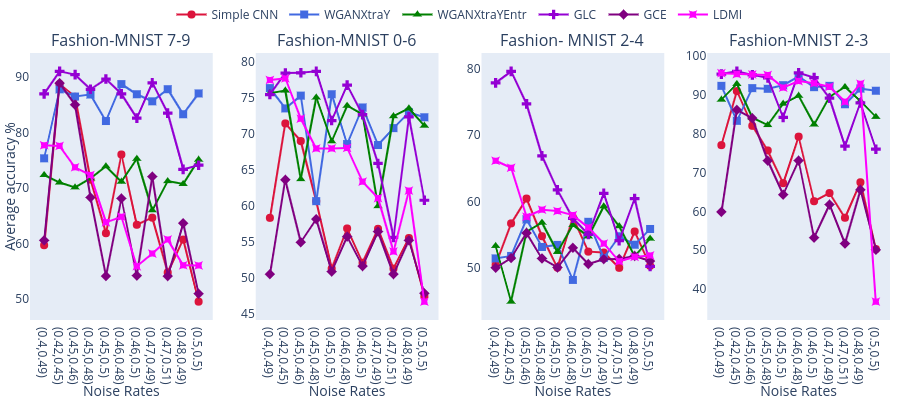}
		\caption{Average accuracy plots for binary class Fashion-MNIST datasets across 5 trials when the final classification model M is a CNN. For Fashion-MNIST 0-6 (Tshirt and Shirt) and Fashion-MNIST 7-9 (Sneakers and Boots), our schemes achieve higher accuracy values than LDMI in more than half noise rates cases and has comparable accuracy to GLC. Fashion-MNIST 2-4 (Pullover and Coat) inherently is a difficult dataset and adding label noise further deteriorates its performance. For Fashion-MNIST 2-3 (Pullover and Dress) our schemes are either better or on par with GLC and LDMI.}
		\label{fig: Syn&FashionMNIST_CNN}
	\end{figure*}

\begin{figure*}[!ht]
		\centering
		\includegraphics[width = 1.01\textwidth]{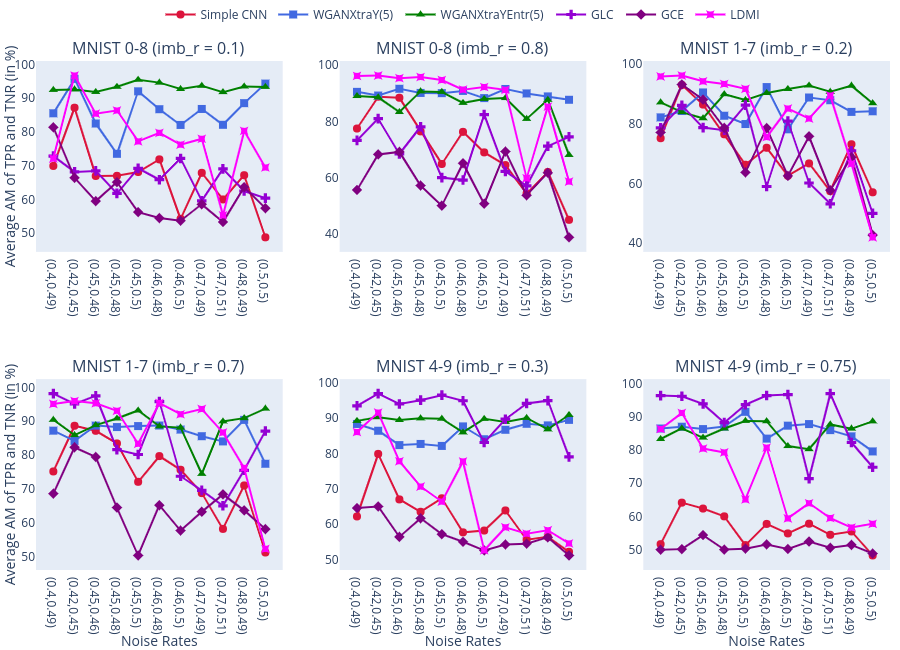}
		\caption{Average AM (Arithmetic Mean of TPR and TNR) value plots for various imbalanced binary class MNIST datasets across 5 trials  when the final classification model M is a CNN. ``imb\_r" denotes the fraction of positive data points. MNIST 0-8 and 1-7 use $0.1\%$ clean data and MNIST 4-9 uses $1\%$ clean data. Our schemes perform better than LDMI and GLC in more than 3 datasets and for more than half of the noise rates. Also, the AM values from our schemes demonstrate more stability across noise rates in comparison to LDMI, GCE and GLC.}
		\label{fig: MNIST_imb_CNN}
	\end{figure*}

	{ We use model $M$ trained on only noisy data (SimpleNN (final MLP classifier), Simple CNN), GLC (with same gold fraction), GCE \cite{zhang2018generalized} (with q = 0.7 in model M) and LDMI \cite{xu2019LDMI}. The code provided by \cite{hendrycks2018using} without any modification in terms of architecture and parameters is used for GLC results. For LDMI, the code provided by the authors is used suitably modified for the model $M$ architecture. As none of the methods used for comparison claim of working in imbalanced settings, it might not be a fair comparison. However, since we do not make any changes to our schemes to account for imbalance, for the sake of completeness, we compare it to GLC, GCE, LDMI and SimpleNN/SimpleCNN.} The results presented in Figure \ref{fig: MNIST}, \ref{fig: Syn&FashionMNIST} and \ref{fig: MNIST_imb} depict the curves of average accuracies and AM when model M is MLP. Since, standard deviation values are very high for almost all noise rates, we present them along with accuracies in detail in  \ref{sec: details_from_exp_main_paper}. In addition, Table \ref{tab: MNIST_5-6_high2} to \ref{tab: MNIST_0-1_high2} and \ref{tab: Fashion_MNIST_7-9_high2} to \ref{tab: Fashion_MNIST_2-3_high2} of  \ref{sec: details_from_exp_main_paper}, have results for 11 more very high noise rates, when $\rho_+, \rho_-$ values are swapped. The results presented in Figure \ref{fig: MNIST_CNN}, \ref{fig: Syn&FashionMNIST_CNN} and \ref{fig: MNIST_imb_CNN} depict the accuracies and AM values for the real datasets when the model M is a CNN. {Based on various binary MNIST versions with the same MLP model M, we observed that dropout and early stopping helps only in case of low uniform label noise (SLN) and not for CCN model which we are interested in. These techniques have been shown to be working well in those noise models which estimate noise rates and hence we do not make a comparison to these schemes.}\\
	\textbf{Observations when model M is MLP} For the case of synthetic dataset SD100 and SD300,  plots in column 1 and 2 of Figure \ref{fig: Syn&FashionMNIST} show that our scheme WGANXtraY(5) has strictly higher accuracies in comparison to existing methods. Also, for SD300, WGANXtraY(5) is not at all affected by very high CCN rates as its accuracies are $\sim 99\%$ always. For binary versions of MNIST balanced datasets in Figure \ref{fig: MNIST}, both our schemes top, especially when both  $\rho_+,\rho_-$ are more than 0.45. For binary balanced Fashion MNIST datasets in Figure \ref{fig: Syn&FashionMNIST}, even though GLC and LDMI show comparable average accuracy for some noise rates the variation across noise trials for them is very high as observed in Table \ref{tab: Fashion_MNIST_7-9_high1} to \ref{tab: Fashion_MNIST_2-3_high2} of  \ref{sec: details_from_exp_main_paper}. Figure \ref{fig: MNIST_imb} shows that for imbalanced MNIST 0-8 and 1-7 datsets, WGANXtraY(5) always outperforms SimpleNN, GCE  and GLC; LDMI is comparable sometime. For MNIST 4-9 imbalanced dataset with imbalance ratio $imb\_r = 0.75$, the performance is comparable to GLC. We would like to point out that the imbalance ratio at two extreme levels is considered which when combined with CCN results in a difficult problem. { We carried out statistical tests for multiple algorithm comparisons using \cite{rodriguez-fdez2015STAC} to test the significance of our results. First, we performed Friedman F test \cite{demvsar2006statistical} individually for every binary dataset to check if there are any significant differences between the performance of various label noise robust schemes. If the the null hypothesis about the equal performance across all learning scheme is rejected we performed pairwise Nemenyi posthoc test. The p values from all the tests are presented in Table \ref{tab: StatTest_MLP}. As can be seen the p value for approximately half of the cases (50/108 in blue colour) is less than 0.1, we conclude that our WGAN based schemes lead to a significant improvement over the existing schemes. Also, for rest of the 58 pair of schemes across various datasets, the performance is comparable. We also observe that there is no significant difference between the performance of WGANXtraY(5) and WGANXtraYEntr(5). Further, based on the p value (in green) we observe that GLC and LDMI have significantly better performance than GCE but have comparable performance among themselves.}   \\
	\textbf{Observations when model M is a CNN} {As can be seen in Figure \ref{fig: MNIST_CNN}, the overall accuracy across all balanced binary CCN corrupted MNIST versions for all schemes has increased. However, the clear improvement of our WGAN based schemes over GLC, LDMI, GCE as observed in Figure \ref{fig: MNIST} is not visible here. Similar kind of phenomenon is observed for Fashion MNIST and imbalanced binary MNIST versions in Figure \ref{fig: Syn&FashionMNIST_CNN} and \ref{fig: MNIST_imb_CNN}. The reason for this phenomenon is as follows: WGAN based schemes use MLP in WGAN architecture and CNN to learn the final classifier whereas the other schemes use CNN throughout the learning process and hence get an undue advantage over WGAN based schemes. This suggests that if one is interested in using a CNN as model M, then one has to identify a suitable CNN architecture compatible representation that can be used in the WGAN based schemes.} \\
	{Finally, unlike other schemes like LDMI (see Figure \ref{fig: MNIST}, \ref{fig: Syn&FashionMNIST} and \ref{fig: MNIST_imb}) that show a  decreasing trend in accuracy as noise rate increases, our schemes show stable accuracy and AM values across noise rates in most of the cases. Also, we would like to point out that GCE is theoretically shown to be SLN noise robust and has been observed to be working empirically for CCN too in the literature. However, in the current experiments, due to high class conditional noise rates, its performance deteriorated.}
	% 	Finally, all the results reported using our schemes do not use any knowledge (no estimation) about the noise rates.
	
% Please add the following required packages to your document preamble:
% \usepackage[table,xcdraw]{xcolor}
% If you use beamer only pass "xcolor=table" option, i.e. \documentclass[xcolor=table]{beamer}
% \usepackage{lscape}
\begin{landscape}
	\begin{table}[]
				\centering
		\bgroup
		\def\arraystretch{1.05}% 
		\tiny{ \setlength{\tabcolsep}{0.2em}
		\begin{tabular}{|c|c|c|c|c|c|c|c|c|c|c|c|c|c|c|c|c|c|c|}
			\hline
			& \multicolumn{6}{c|}{\textbf{MNIST Balanced}} & \multicolumn{2}{c|}{\textbf{Synthetic data}} & \multicolumn{4}{c|}{\textbf{Fashion MNIST}} & \multicolumn{6}{c|}{\textbf{MNIST imbalanced}} \\ \hline
			\textbf{Dataset} & \textbf{5-6 digit} & \textbf{7-9 digit} & \textbf{4-9 digit} & \textbf{1-7 digit} & \textbf{0-6 digit} & \textbf{0-1 digit} & \textbf{SD100} & \textbf{SD300} & \textbf{7-9 pair} & \textbf{0-6 pair} & \textbf{2-4 pair} & \textbf{2-3 pair} & \textbf{0-8 (0.1)} & \textbf{0-8(0.8)} & \textbf{1-7(0.2)} & \textbf{1-7(0.7)} & \textbf{4-9(0.3)} & \textbf{4-9(0.75)} \\ \hline
			\textbf{\begin{tabular}[c]{@{}c@{}}Friedman Test \\ p values\end{tabular}} & 0.000000 & 0.000000 & 0.000000 & 0.000000 & 0.000000 & 0.000000 & 0.000000 & 0.000000 & 0.000000 & 0.000000 & 0.000000 & 0.000000 & 0.000000 & 0.000000 & 0.000000 & 0.000000 & 0.000000 & 0.000000 \\ \hline
			\textbf{Algorithm pairs} & \multicolumn{18}{c|}{\textbf{Nemenyi posthoc test (pairwise) p values}} \\ \hline
			\textbf{\begin{tabular}[c]{@{}c@{}}WGANXtraY(5) \\ vs GLC\end{tabular}} & 1.000000 & 1.000000 & 1.000000 & 1.000000 & 1.000000 & 0.182573 & {\color[HTML]{3166FF} \textbf{0.000000}} & {\color[HTML]{3166FF} \textbf{0.000000}} & 1.000000 & 0.790576 & 1.000000 & 1.000000 & {\color[HTML]{3166FF} \textbf{0.000783}} & {\color[HTML]{3166FF} \textbf{0.078567}} & {\color[HTML]{3166FF} \textbf{0.037922}} & 0.214194 & 1.000000 & 1.000000 \\ \hline
			\textbf{\begin{tabular}[c]{@{}c@{}}WGANXtraYEntr(5)\\  vs GLC\end{tabular}} & 1.000000 & 1.000000 & 1.000000 & \multicolumn{1}{r|}{{\color[HTML]{3166FF} \textbf{0.000001}}} & 1.000000 & 0.603580 & {\color[HTML]{3166FF} \textbf{0.003984}} & {\color[HTML]{3166FF} \textbf{0.001602}} & 1.000000 & 1.000000 & 1.000000 & 0.455527 & {\color[HTML]{3166FF} \textbf{0.000783}} & {\color[HTML]{3166FF} \textbf{0.078567}} & {\color[HTML]{3166FF} \textbf{0.037922}} & 0.214194 & 1.000000 & 1.000000 \\ \hline
			\textbf{\begin{tabular}[c]{@{}c@{}}WGANXtraY(5)\\  vs GCE\end{tabular}} & {\color[HTML]{3166FF} \textbf{0.000101}} & {\color[HTML]{3166FF} \textbf{0.000172}} & {\color[HTML]{3166FF} \textbf{0.000997}} & {\color[HTML]{3166FF} \textbf{0.000025}} & {\color[HTML]{3166FF} \textbf{0.000172}} & {\color[HTML]{3166FF} \textbf{0.000000}} & 0.603580 & 0.603580 & {\color[HTML]{3166FF} \textbf{0.006168}} & {\color[HTML]{3166FF} \textbf{0.000014}} & {\color[HTML]{3166FF} \textbf{0.065782}} & {\color[HTML]{3166FF} \textbf{0.000004}} & {\color[HTML]{3166FF} \textbf{0.000006}} & {\color[HTML]{3166FF} \textbf{0.000004}} & {\color[HTML]{3166FF} \textbf{0.000000}} & {\color[HTML]{3166FF} \textbf{0.000006}} & {\color[HTML]{3166FF} \textbf{0.000001}} & {\color[HTML]{3166FF} \textbf{0.000059}} \\ \hline
			\textbf{\begin{tabular}[c]{@{}c@{}}WGANXtraYEntr(5)\\ vs GCE\end{tabular}} & \multicolumn{1}{r|}{{\color[HTML]{3166FF} \textbf{0.000006}}} & {\color[HTML]{3166FF} \textbf{0.000101}} & {\color[HTML]{3166FF} \textbf{0.000008}} & {\color[HTML]{3166FF} \textbf{0.000001}} & {\color[HTML]{3166FF} \textbf{0.000006}} & {\color[HTML]{3166FF} \textbf{0.000006}} & 1.000000 & 1.000000 & {\color[HTML]{3166FF} \textbf{0.003984}} & {\color[HTML]{3166FF} \textbf{0.000613}} & {\color[HTML]{3166FF} \textbf{0.002542}} & 0.131472 & {\color[HTML]{3166FF} \textbf{0.000006}} & {\color[HTML]{3166FF} \textbf{0.000004}} & {\color[HTML]{3166FF} \textbf{0.000000}} & {\color[HTML]{3166FF} \textbf{0.000006}} & {\color[HTML]{3166FF} \textbf{0.000001}} & {\color[HTML]{3166FF} \textbf{0.000059}} \\ \hline
			\textbf{\begin{tabular}[c]{@{}c@{}}WGANXtraY(5)\\  vs LDMI\end{tabular}} & 1.000000 & 1.000000 & 1.000000 & 0.603580 & 1.000000 & 0.131472 & {\color[HTML]{3166FF} \textbf{0.001602}} & {\color[HTML]{3166FF} \textbf{0.000613}} & 1.000000 & 1.000000 & 0.603580 & 1.000000 & 0.900907 & 1.000000 & 0.691773 & 0.900907 & {\color[HTML]{3166FF} \textbf{0.037922}} & {\color[HTML]{3166FF} \textbf{0.078567}} \\ \hline
			\textbf{\begin{tabular}[c]{@{}c@{}}WGANXtraYEntr(5)\\  vs LDMI\end{tabular}} & {\color[HTML]{3166FF} \textbf{0.339817}} & 1.000000 & 1.000000 & {\color[HTML]{3166FF} \textbf{0.093555}} & 1.000000 & 0.455527 & 1.000000 & 0.603580 & 1.000000 & 1.000000 & 0.603580 & 1.000000 & 0.900907 & 1.000000 & 0.691773 & 0.900907 & {\color[HTML]{3166FF} \textbf{0.037922}} & {\color[HTML]{3166FF} \textbf{0.078567}} \\ \hline
			\textbf{\begin{tabular}[c]{@{}c@{}}WGANXtraY(5) vs \\ WGANXtraYEntr(5)\end{tabular}} & 1.000000 & 1.000000 & 1.000000 & 1.000000 & 1.000000 & 1.000000 & 0.339817 & 0.603580 & 1.000000 & 1.000000 & 1.000000 & 0.182573 & 1.000000 & 1.000000 & 1.000000 & 1.000000 & 1.000000 & 1.000000 \\ \hline
			\textbf{\begin{tabular}[c]{@{}c@{}}Simple NN vs \\ WGANXtraY(5)\end{tabular}} & {\color[HTML]{3166FF} \textbf{0.011614}} & {\color[HTML]{3166FF} \textbf{0.007640}} & {\color[HTML]{3166FF} \textbf{0.093555}} & {\color[HTML]{3166FF} \textbf{0.000372}} & {\color[HTML]{3166FF} \textbf{0.011614}} & {\color[HTML]{3166FF} \textbf{0.000289}} & {\color[HTML]{3166FF} \textbf{0.000045}} & {\color[HTML]{3166FF} \textbf{0.000045}} & {\color[HTML]{3166FF} \textbf{0.093555}} & {\color[HTML]{3166FF} \textbf{0.006168}} & {\color[HTML]{3166FF} \textbf{0.790576}} & {\color[HTML]{3166FF} \textbf{0.002542}} & {\color[HTML]{3166FF} \textbf{0.000783}} & {\color[HTML]{3166FF} \textbf{0.000613}} & {\color[HTML]{3166FF} \textbf{0.002021}} & {\color[HTML]{3166FF} \textbf{0.000289}} & {\color[HTML]{3166FF} \textbf{0.000223}} & {\color[HTML]{3166FF} \textbf{0.000289}} \\ \hline
			\textbf{\begin{tabular}[c]{@{}c@{}}Simple NN vs \\ WGANXtraYEntr(5)\end{tabular}} & {\color[HTML]{3166FF} \textbf{0.001266}} & {\color[HTML]{3166FF} \textbf{0.004965}} & {\color[HTML]{3166FF} \textbf{0.002542}} & {\color[HTML]{3166FF} \textbf{0.000014}} & {\color[HTML]{3166FF} \textbf{0.000783}} & {\color[HTML]{3166FF} \textbf{0.002021}} & 0.250551 & 0.131472 & 0.065782 & {\color[HTML]{3166FF} \textbf{0.093555}} & {\color[HTML]{3166FF} \textbf{0.065782}} & 1.000000 & {\color[HTML]{3166FF} \textbf{0.000783}} & {\color[HTML]{3166FF} \textbf{0.000613}} & {\color[HTML]{3166FF} \textbf{0.002021}} & {\color[HTML]{3166FF} \textbf{0.000289}} & {\color[HTML]{3166FF} \textbf{0.000223}} & {\color[HTML]{3166FF} \textbf{0.000289}} \\ \hline
			\textbf{GLC vs GCE} & {\color[HTML]{32CB00} \textbf{0.004965}} & {\color[HTML]{32CB00} \textbf{0.001266}} & {\color[HTML]{32CB00} \textbf{0.021275}} & {\color[HTML]{32CB00} \textbf{0.021275}} & {\color[HTML]{32CB00} \textbf{0.003187}} & {\color[HTML]{32CB00} \textbf{0.037922}} & {\color[HTML]{32CB00} \textbf{0.001602}} & {\color[HTML]{32CB00} \textbf{0.001602}} & {\color[HTML]{32CB00} \textbf{0.003984}} & {\color[HTML]{32CB00} \textbf{0.045702}} & {\color[HTML]{32CB00} \textbf{0.045702}} & {\color[HTML]{32CB00} \textbf{0.000025}} & 1.000000 & 0.292219 & 0.182573 & 0.131472 & {\color[HTML]{32CB00} \textbf{0.001266}} & {\color[HTML]{32CB00} \textbf{0.000997}} \\ \hline
			\textbf{Simple NN vs GLC} & 0.214194 & {\color[HTML]{32CB00} \textbf{0.037922}} & 0.790576 & 0.131472 & 0.111070 & 1.000000 & 1.000000 & 1.000000 & {\color[HTML]{32CB00} \textbf{0.065782}} & 1.000000 & 1.000000 & {\color[HTML]{32CB00} \textbf{0.009434}} & 1.000000 & 1.000000 & 1.000000 & 1.000000 & {\color[HTML]{32CB00} \textbf{0.078567}} & {\color[HTML]{32CB00} \textbf{0.003984}} \\ \hline
			\textbf{GLC vs LDMI} & 1.000000 & 1.000000 & 1.000000 & 1.000000 & 1.000000 & 1.000000 & 0.603580 & 1.000000 & {\color[HTML]{32CB00} \textbf{0.002542}} & 1.000000 & 0.790576 & 1.000000 & 0.455527 & 1.000000 & 1.000000 & 1.000000 & 1.000000 & 0.455527 \\ \hline
			\textbf{Simple NN vs GCE} & 1.000000 & 1.000000 & 1.000000 & 1.000000 & 1.000000 & 1.000000 & 0.131472 & 0.131472 & 1.000000 & 1.000000 & 1.000000 & 1.000000 & 1.000000 & 1.000000 & 1.000000 & 1.000000 & 1.000000 & 1.000000 \\ \hline
			\textbf{GCE vs LDMI} & {\color[HTML]{32CB00} \textbf{0.078567}} & {\color[HTML]{32CB00} \textbf{0.011614}} & {\color[HTML]{32CB00} \textbf{0.014254}} & {\color[HTML]{32CB00} \textbf{0.093555}} & {\color[HTML]{32CB00} \textbf{0.004965}} & {\color[HTML]{32CB00} \textbf{0.054913}} & 1.000000 & 0.603580 & 1.000000 & {\color[HTML]{32CB00} \textbf{0.000613}} & {\color[HTML]{32CB00} \textbf{0.000014}} & {\color[HTML]{32CB00} \textbf{0.000372}} & {\color[HTML]{32CB00} \textbf{0.021275}} & {\color[HTML]{32CB00} \textbf{0.007640}} & {\color[HTML]{32CB00} \textbf{0.006168}} & {\color[HTML]{32CB00} \textbf{0.021275}} & 0.214194 & 1.000000 \\ \hline
			\textbf{Simple NN vs LDMI} & 1.000000 & {\color[HTML]{32CB00} \textbf{0.214194}} & 0.603580 & 0.455527 & {\color[HTML]{32CB00} \textbf{0.004965}} & 1.000000 & 1.000000 & 1.000000 & {\color[HTML]{32CB00} \textbf{0.045702}} & {\color[HTML]{32CB00} \textbf{0.093555}} & {\color[HTML]{32CB00} \textbf{0.000997}} & {\color[HTML]{32CB00} \textbf{0.065782}} & 0.455527 & 0.214194 & 1.000000 & 0.250551 & 1.000000 & 1.000000 \\ \hline
		\end{tabular}}
	\egroup
		\caption{The table depicts the p value for the Friedman F test (row 3) and Nemenyi posthoc test on the MLP based classifiers' performance. Since all Friedman F test based p values are 0,  there is a significant difference between the performance of the presented schemes.  The p value in {\color[HTML]{3166FF} \textbf{blue }} colour represents the cases where we reject the null hypothesis that the performance of WGAN based schemes is same as the existing scheme (column 1) at 10\% level of significance. Hence, we conclude that WGAN based schemes have either comparable performance or lead to a statistically significant improvement in the accuracy or AM values. {\color[HTML]{32CB00} \textbf{Green}} colour shows the significant values for the comparison among existing schemes. }
		\label{tab: StatTest_MLP}
	\end{table}
\end{landscape}

	\section{Discussion} \label{sec: disc}
	
	We propose WGAN based schemes for CCN robust binary classification using a small set  of clean labels (0.1\% or 1\% of the training data). We exploit the adversarial nature of WGANs for this purpose. But, we need to overcome a couple of challenges to adapt conventional WGANs to be CCN robust. First, we feed clean {\em labelled}  data to the discriminator and corrupted {\em labelled} data to the generator. Next, we include a new representation to address skewed ratio of feature to label dimension. The second scheme, in addition to above, also uses entropy ideas. We theoretically show that high noise rates lead to higher KL divergence between clean and SLN corrupted distribution and in turn, to better performance of our schemes. These aspects along with the generative nature of our scheme makes it improve over existing discriminative schemes like GLC, GCE and LDMI at high noise rates (close to 0.5), using small neural networks and without knowing estimating the noise rates. The generative nature of our schemes impart them with the potential of being CCN robust without any modification, even when the data has class imbalance. To make a well-aware choice of a GAN for generating labelled data, we explored WGAN, PacGAN, VEEGAN and WGAN-GP and based on their performances selected WGAN in our work. {We demonstrate the good performance of our schemes and comparison to GLC, GCE and LDMI on  high dimensional binary class synthetic, MNIST and Fashion MNIST datasets. We perform Friedman F test and Nemenyi posthoc test to  statistically support our claim about the improvement of our schemes over the existing schemes. Finally, in most of our experiments we observed that the performance (accuracy or AM values) of WGAN based schemes is stable across noise rates. This is due to the fact that the labelled data used for training final classification model M is generated by the WGAN while enforcing the fact that its distribution should be as close as possible to the clean distribution. }
	
	Our work is an initial attempt to use GANs for CCN robust classification without knowing/estimating the noise rates. A natural extension is to use it for multi-class datasets where the representation idea has to be accordingly changed in addition to the notion of high noise rates which would be close to 1. { As we are not explicitly modelling the noise, in addition to SLN and CCN, our scheme can possibly handle instance-dependent noise. The use of WGANs with MLP architecture and classification model M with CNN architecture demonstrates `just' comparable performance to existing schemes unlike the significantly better performance when classification model M is MLP. This opens the direction to identify representations which helps us use convolution in WGAN for generating feature-label pair. An interesting avenue is to treat the considered framework as a semi-supervised learning problem. An idea worth exploring based on \cite{shaham2019singan} could be to devise a framework that only uses 1 or 2 clean labelled samples to generate more labelled data and use it for classification. The question here would be identify a suitable label representation that is compatible with the scheme proposed in \cite{shaham2019singan}. In addition, one has to also answer the question of integrating the additional noisy labelled data in the above setup, ignoring which could lead to loss of information. Finally, the thread about the good performance of our schemes could be complete if one could theoretically show that higher divergence between the true and latent distribution in GAN leads to better performance of GAN w.r.t. an appropriate measure. }
	
	% \section*{References}
	
	\bibliography{ref}

\begin{thebibliography}{10}
\expandafter\ifx\csname url\endcsname\relax
  \def\url#1{\texttt{#1}}\fi
\expandafter\ifx\csname urlprefix\endcsname\relax\def\urlprefix{URL }\fi
\expandafter\ifx\csname href\endcsname\relax
  \def\href#1#2{#2} \def\path#1{#1}\fi

\bibitem{zhang2016understanding}
C.~Zhang, S.~Bengio, M.~Hardt, B.~Recht, O.~Vinyals, Understanding deep
  learning requires rethinking generalization, in: International Conference on
  Learning Representations (ICLR), 2017.

\bibitem{sastry2017robustbookchapter}
P.~Sastry, N.~Manwani, Robust learning of classifiers in the presence of label
  noise, in: Pattern Recognition and Big Data, World Scientific, 2017, pp.
  167--197.

\bibitem{hendrycks2018using}
D.~Hendrycks, M.~Mazeika, D.~Wilson, K.~Gimpel, Using trusted data to train
  deep networks on labels corrupted by severe noise, in: Advances in Neural
  Information Processing Systems, 2018, pp. 10456--10465.

\bibitem{ren2018learning}
M.~Ren, W.~Zeng, B.~Yang, R.~Urtasun, Learning to reweight examples for robust
  deep learning, in: International Conference on Machine Learning, 2018, pp.
  4331--4340.

\bibitem{liu2016classification}
T.~Liu, D.~Tao, Classification with noisy labels by importance reweighting,
  IEEE Transactions on Pattern Analysis and Machine Intelligence 38~(3) (2016)
  447--461.

\bibitem{han2018masking}
B.~Han, J.~Yao, G.~Niu, M.~Zhou, I.~Tsang, Y.~Zhang, M.~Sugiyama, Masking: A
  new perspective of noisy supervision, in: Advances in Neural Information
  Processing Systems, 2018, pp. 5836--5846.

\bibitem{ma2018dimensionality}
X.~Ma, Y.~Wang, M.~E. Houle, S.~Zhou, S.~M. Erfani, S.-T. Xia, S.~Wijewickrema,
  J.~Bailey, Dimensionality-driven learning with noisy labels, in:
  International Conference on Machine Learning, 2018, pp. 3361--3370.

\bibitem{goodfellow2014generative}
I.~Goodfellow, J.~Pouget-Abadie, M.~Mirza, B.~Xu, D.~Warde-Farley, S.~Ozair,
  A.~Courville, Y.~Bengio, Generative adversarial nets, in: Advances in Neural
  Information Processing Systems, 2014, pp. 2672--2680.

\bibitem{lin2018pacgan}
Z.~Lin, A.~Khetan, G.~Fanti, S.~Oh, Pac{GAN}: The power of two samples in
  generative adversarial networks, in: Advances in Neural Information
  Processing Systems, 2018, pp. 1498--1507.

\bibitem{gulrajani2017improved}
I.~Gulrajani, F.~Ahmed, M.~Arjovsky, V.~Dumoulin, A.~C. Courville, Improved
  training of {W}asserstein {GAN}s, in: Advances in Neural Information
  Processing Systems, 2017, pp. 5767--5777.

\bibitem{srivastava2017veegan}
A.~Srivastava, L.~Valkov, C.~Russell, M.~U. Gutmann, C.~Sutton, {VEEGAN}:
  Reducing mode collapse in gans using implicit variational learning, in:
  Advances in Neural Information Processing Systems, 2017, pp. 3308--3318.

\bibitem{natarajan2013learning}
N.~Natarajan, I.~S. Dhillon, P.~K. Ravikumar, A.~Tewari, Learning with noisy
  labels, in: Advances in Neural Information Processing Systems, 2013, pp.
  1196--1204.

\bibitem{ghosh2015making}
A.~Ghosh, N.~Manwani, P.~Sastry, Making risk minimization tolerant to label
  noise, Neurocomputing 160 (2015) 93--107.

\bibitem{van2015learning}
B.~Van~Rooyen, A.~Menon, R.~C. Williamson, Learning with symmetric label noise:
  The importance of being unhinged, in: Advances in Neural Information
  Processing Systems, 2015, pp. 10--18.

\bibitem{patrini2016loss}
G.~Patrini, F.~Nielsen, R.~Nock, M.~Carioni, Loss factorization, weakly
  supervised learning and label noise robustness, in: International Conference
  on Machine Learning, 2016, pp. 708--717.

\bibitem{tripathi2019cost}
S.~Tripathi, N.~Hemachandra, Cost sensitive learning in the presence of
  symmetric label noise, in: PAKDD, Springer, 2019, pp. 15--28.

\bibitem{patrini2017making}
G.~Patrini, A.~Rozza, A.~K. Menon, R.~Nock, L.~Qu, Making deep neural networks
  robust to label noise: A loss correction approach, in: 2017 IEEE Conference
  on Computer Vision and Pattern Recognition (CVPR), IEEE, 2017, pp.
  2233--2241.

\bibitem{ghosh2017robust_deep}
A.~Ghosh, H.~Kumar, P.~Sastry, Robust loss functions under label noise for deep
  neural networks., in: AAAI, 2017, pp. 1919--1925.

\bibitem{DBLP:journals/ml/MenonRN18}
A.~K. Menon, B.~van Rooyen, N.~Natarajan,
  \href{https://doi.org/10.1007/s10994-018-5715-3}{Learning from binary labels
  with instance-dependent noise}, Machine Learning 107~(8-10) (2018)
  1561--1595.
\newblock \href {http://dx.doi.org/10.1007/s10994-018-5715-3}
  {\path{doi:10.1007/s10994-018-5715-3}}.
\newline\urlprefix\url{https://doi.org/10.1007/s10994-018-5715-3}

\bibitem{han2018co-teaching}
B.~Han, Q.~Yao, X.~Yu, G.~Niu, M.~Xu, W.~Hu, I.~Tsang, M.~Sugiyama,
  Co-teaching: Robust training of deep neural networks with extremely noisy
  labels, in: Advances in Neural Information Processing Systems, 2018, pp.
  8536--8546.

\bibitem{chen2019understanding}
P.~Chen, B.~B. Liao, G.~Chen, S.~Zhang, Understanding and utilizing deep neural
  networks trained with noisy labels, in: International Conference on Machine
  Learning, 2019, pp. 1062--1070.

\bibitem{amid2019robustBitempered}
E.~Amid, M.~K. Warmuth, R.~Anil, T.~Koren, Robust bi-tempered logistic loss
  based on bregman divergences, in: Advances in Neural Information Processing
  Systems, 2019, pp. 14987--14996.

\bibitem{zhang2018generalized}
Z.~Zhang, M.~Sabuncu, Generalized cross entropy loss for training deep neural
  networks with noisy labels, in: Advances in neural information processing
  systems, 2018, pp. 8778--8788.

\bibitem{xu2019LDMI}
Y.~Xu, P.~Cao, Y.~Kong, Y.~Wang, L\_dmi: A novel information-theoretic loss
  function for training deep nets robust to label noise, in: Advances in Neural
  Information Processing Systems, 2019, pp. 6222--6233.

\bibitem{kim2019nlnl}
Y.~Kim, J.~Yim, J.~Yun, J.~Kim, Nlnl: Negative learning for noisy labels, in:
  Proceedings of the IEEE International Conference on Computer Vision, 2019,
  pp. 101--110.

\bibitem{wang2019symmetric}
Y.~Wang, X.~Ma, Z.~Chen, Y.~Luo, J.~Yi, J.~Bailey, Symmetric cross entropy for
  robust learning with noisy labels, in: Proceedings of the IEEE International
  Conference on Computer Vision, 2019, pp. 322--330.

\bibitem{li2017learningDistillation}
Y.~Li, J.~Yang, Y.~Song, L.~Cao, J.~Luo, L.-J. Li, Learning from noisy labels
  with distillation, in: Proceedings of the IEEE International Conference on
  Computer Vision, 2017, pp. 1910--1918.

\bibitem{vahdat2017toward}
A.~Vahdat, Toward robustness against label noise in training deep
  discriminative neural networks, in: Advances in Neural Information Processing
  Systems, 2017, pp. 5596--5605.

\bibitem{veit2017learning}
A.~Veit, N.~Alldrin, G.~Chechik, I.~Krasin, A.~Gupta, S.~Belongie, Learning
  from noisy large-scale datasets with minimal supervision, in: Proceedings of
  the IEEE Conference on Computer Vision and Pattern Recognition, 2017, pp.
  839--847.

\bibitem{thekumparampil2018robustness}
K.~K. Thekumparampil, A.~Khetan, Z.~Lin, S.~Oh, Robustness of conditional gans
  to noisy labels, in: Advances in Neural Information Processing Systems, 2018,
  pp. 10271--10282.

\bibitem{kaneko2019label}
T.~Kaneko, Y.~Ushiku, T.~Harada, Label-noise robust generative adversarial
  networks, in: Proceedings of the IEEE Conference on Computer Vision and
  Pattern Recognition, 2019, pp. 2467--2476.

\bibitem{arjovsky2017wasserstein}
M.~Arjovsky, S.~Chintala, L.~Bottou, Wasserstein generative adversarial
  networks, in: International Conference on Machine Learning, 2017, pp.
  214--223.

\bibitem{zhao2018classification}
Z.~Zhao, L.~Chu, D.~Tao, J.~Pei, Classification with label noise: {A} {M}arkov
  chain sampling framework, Data Mining and Knowledge Discovery (2018) 1--37.

\bibitem{liu2017approximation}
S.~Liu, O.~Bousquet, K.~Chaudhuri, Approximation and convergence properties of
  generative adversarial learning, in: Advances in Neural Information
  Processing Systems, 2017, pp. 5545--5553.

\bibitem{springenberg2016iclr}
J.~T. Springenberg, \href{https://arxiv.org/abs/1511.06390}{Unsupervised and
  semi-supervised learning with categorical generative adversarial networks},
  in: International Conference on Learning Representations (ICLR), 2016.
\newline\urlprefix\url{https://arxiv.org/abs/1511.06390}

\bibitem{grandvalet2005semi}
Y.~Grandvalet, Y.~Bengio, Semi-supervised learning by entropy minimization, in:
  Advances in neural information processing systems, 2005, pp. 529--536.

\bibitem{Mnistlecun1998gradient}
Y.~LeCun, L.~Bottou, Y.~Bengio, P.~Haffner, et~al., Gradient-based learning
  applied to document recognition, Proceedings of the IEEE 86~(11) (1998)
  2278--2324.

\bibitem{xiao2017fashion}
H.~Xiao, K.~Rasul, R.~Vollgraf, Fashion-{MNIST}: {A} novel image dataset for
  benchmarking machine learning algorithms, arXiv preprint arXiv:1708.07747.

\bibitem{natarajan2017cost}
N.~Natarajan, I.~S. Dhillon, P.~Ravikumar, A.~Tewari,
  \href{http://jmlr.org/papers/v18/15-226.html}{Cost-sensitive learning with
  noisy labels}, Journal of Machine Learning Research 18~(155) (2018) 1--33.
\newline\urlprefix\url{http://jmlr.org/papers/v18/15-226.html}

\bibitem{koziarski2019radial}
M.~Koziarski, B.~Krawczyk, M.~Wo{\'z}niak, Radial-based oversampling for noisy
  imbalanced data classification, Neurocomputing 343 (2019) 19--33.

\bibitem{rodriguez-fdez2015STAC}
I.~Rodr\'{i}guez-Fdez, A.~Canosa, M.~Mucientes, A.~Bugar\'{i}n, {STAC}: a web
  platform for the comparison of algorithms using statistical tests, in:
  Proceedings of the 2015 IEEE International Conference on Fuzzy Systems
  (FUZZ-IEEE), 2015.

\bibitem{demvsar2006statistical}
J.~Dem{\v{s}}ar, Statistical comparisons of classifiers over multiple data
  sets, Journal of Machine learning research 7~(Jan) (2006) 1--30.

\bibitem{shaham2019singan}
T.~R. Shaham, T.~Dekel, T.~Michaeli, Singan: Learning a generative model from a
  single natural image, in: Proceedings of the IEEE International Conference on
  Computer Vision, 2019, pp. 4570--4580.

\bibitem{derExchangeCondn}
Jochen, {Interchange of derivative with an expectation},
  \url{https://math.stackexchange.com/questions/217702/when-can-we-interchange-the-derivative-with-an-expectation}
  (2016).

\bibitem{efron1997improvements}
B.~Efron, R.~Tibshirani, Improvements on cross-validation: the 632+ bootstrap
  method, Journal of the American Statistical Association 92~(438) (1997)
  548--560.

\end{thebibliography}
	
	\appendix
	\section{Proofs}
	\subsection{Proof of Lemma \ref{lem: KL_SLN }} \label{subsec: KL_SLN_proof}
	\begin{proof}
		Let the joint density corresponding to clean ($\mathcal{D}$) and noisy distribution ($\tilde{\mathcal{D}}$) be $p_{c}(\mathbf{x},y)$ and $p_{n}(\mathbf{x},y)$ respectively. Then, we have
		\begin{eqnarray*}
			KL[\mathcal{D}\Vert \tilde{\mathcal{D}}]_{SLN}  = \int\limits_{\mathbf{x}, y} p_{c}(\mathbf{x},y) \log \frac{p_{c}(\mathbf{x},y)}{p_{n}(\mathbf{x},y)} d\mathbf{x}dy \\
			= \int\limits_{\mathbf{x}, y} p_{c}(y|\mathbf{x}) p_{c}(\mathbf{x}) \log \frac{p_{c}(y|\mathbf{x}) p_{c}(\mathbf{x})}{p_{n}(y|\mathbf{x}) p_{n}(\mathbf{x})} d\mathbf{x}dy.
		\end{eqnarray*}
		Since, in $\tilde{\mathcal{D}}$ only label $Y$ is corrupted $p_{n}(\mathbf{x}) = p_{c}(\mathbf{x})$. Also, since $Y$ is binary label, we can write the KL divergence as follows:
		\begin{eqnarray} \nonumber
		KL[\mathcal{D}\Vert \tilde{\mathcal{D}}]_{SLN}  = \int\limits_{\mathbf{x}}\left[ \sum\limits_{y} p_{c}(y|\mathbf{x}) \log \frac{p_{c}(y|\mathbf{x})}{p_{n}(y|\mathbf{x})}\right] p_{x}(\mathbf{x}) d\mathbf{x} \\ \label{eq: KL_withexpSLN}
		= \mathbb{E}_{\mathbf{X}}[ KL(p_{c}(Y | \mathbf{X}) \Vert p_{n}(Y | \mathbf{X})) ].
		\end{eqnarray}
		The KL term inside expectation in Eq. (\ref{eq: KL_withexpSLN}) can be simplified using the fact that $p_{c}(Y = 1|\mathbf{X} = \mathbf{x}) = \eta(\mathbf{x})$ and $p_{n}(Y = 1|\mathbf{X} = \mathbf{x}) = \tilde{\eta}(\mathbf{x})$. 
		\begin{eqnarray} \nonumber
		KL(p_{c}(Y | \mathbf{X} = \mathbf{x}) \Vert p_{n}(Y | \mathbf{X}= \mathbf{x}))
		= \eta(\mathbf{x}) \log \frac{\eta(\mathbf{x})}{\tilde{\eta}(\mathbf{x})}  \\ \label{eq: KL_withEtaExpanSLN}
		+ (1-\eta(\mathbf{x})) \log \frac{1 - \eta(\mathbf{x})}{1 - \tilde{\eta}(\mathbf{x})}.
		\end{eqnarray}
		Also, since for SLN case, $\tilde{\eta}(\mathbf{x}) = (1-2\rho)\eta(\mathbf{x}) + \rho$, we have, 
		\begin{eqnarray} \nonumber
		KL(p_{c}(Y | \mathbf{X} = \mathbf{x}) \Vert p_{n}(Y | \mathbf{X}= \mathbf{x})) \\ \nonumber
		=  -\eta(\mathbf{x}) \log \left(1+ \rho \left(\frac{1-2\eta(\mathbf{x})}{\eta(\mathbf{x})}\right)\right)  \\
		\label{eq: KL_inb/w_condSLN}
		- (1-\eta(\mathbf{x}))\log \left(1- \rho \left(\frac{1-2\eta(\mathbf{x})}{\eta(\mathbf{x})}\right)\right).
		\end{eqnarray}
		Taking derivative of Eq. (\ref{eq: KL_inb/w_condSLN}) w.r.t. noise rate $\rho$, we get
		\begin{equation} \label{eq: der_of_cond_KL_SLN}
		\frac{\rho(1-2\eta(\mathbf{x}))^2}{(\eta(\mathbf{x}) + \rho(1-2\eta(\mathbf{x})))(1-\eta(\mathbf{x}) - \rho(1-2\eta(\mathbf{x})))}.
		\end{equation}
		The derivative in Eq. (\ref{eq: der_of_cond_KL_SLN}) is always positive $\forall ~\mathbf{x}$ implying that the KL divergence between conditional clean and noisy distribution for a given data point $\mathbf{x}$ is an increasing function of $\rho$. Next, to show the monotonicity of $KL[\mathcal{D}\Vert \tilde{\mathcal{D}}]_{SLN}$, using Eq. (\ref{eq: KL_withexpSLN}), we have to verify the following: 
		{
			$$ \frac{d }{d\rho} \mathbb{E}_{\mathbf{X}}[ KL(p_{c}(Y | \mathbf{X}) \Vert p_{n}(Y | \mathbf{X})) ] = \mathbb{E}_{\mathbf{X}}\frac{d}{d\rho}KL(p_{c}(Y | \mathbf{X}) \Vert p_{n}(Y | \mathbf{X})).  $$ }
		The above condition holds if the derivative in Eq. (\ref{eq: der_of_cond_KL_SLN}) can be uniformly upper bounded by an integrable function \cite{derExchangeCondn}. Since, $\rho < 0.5$ lower bounds both the terms in the denominator of Eq. (\ref{eq: der_of_cond_KL_SLN}), we obtain the following:
		\begin{eqnarray}
		\frac{\rho(1-2\eta(\mathbf{x}))^2}{(\eta(\mathbf{x}) + \rho(1-2\eta(\mathbf{x})))(1-\eta(\mathbf{x}) - \rho(1-2\eta(\mathbf{x})))} <  \frac{\rho(1-2\eta(\mathbf{x}))^2}{\rho^2}.
		\end{eqnarray} 
		Now, since, $\eta(\mathbf{x}) \in [0,1]$, $1$ is a trivial upper bound of $(1-2\eta(\mathbf{x}))^2$, 
		% \textcolor{blue}{will dropping the first positive term $1 - \eta (x)$ and then taking the absolute value works? }
		% Further, the RHS of above equation can be upper bounded by separately considering two cases of $\eta(\mathbf{x})$ less than $0.5$ and more than $0.5$ and then combining the bounds. First, let $\eta(\mathbf{x}) <  0.5$ and rewrite $(1-\eta(\mathbf{x}) - \rho(1-2\eta(\mathbf{x}))) = (1-\eta(\mathbf{x}) - 2 \rho(1-2\eta(\mathbf{x})) + \rho(1-2\eta(\mathbf{x})))$. Then, assuming that $\rho \leq 0.5$, which is reasonable, we get
		% $$ \frac{(1-2\eta(\mathbf{x}))}{(1-\eta(\mathbf{x}) - \rho(1-2\eta(\mathbf{x})))} < \frac{1}{\rho}. $$
		% Now, let $\eta(\mathbf{x}) >  0.5$ and rewrite $(1-\eta(\mathbf{x}) - \rho(1-2\eta(\mathbf{x}))) = (1-2\eta(\mathbf{x}) + \eta(\mathbf{x}) - \rho(1-2\eta(\mathbf{x})))$, then we have,
		% $$ \frac{(1-2\eta(\mathbf{x}))}{(1-\eta(\mathbf{x}) - \rho(1-2\eta(\mathbf{x})))} <  1. $$
		% Combining the upper bounds,
		implying that the derivative in Eq. (\ref{eq: der_of_cond_KL_SLN})
		can be upper bounded by $1/\rho$. This implies that the required upper bound on derivative from \cite{derExchangeCondn} is a constant and hence, integrable trivially.  Since, the conditions for interchanging the derivative and expectation are satisfied, the monotonically increasing nature of $KL(p_{c}(Y | \mathbf{X}) \Vert p_{n}(Y | \mathbf{X}))$ implies that $KL[\mathcal{D}\Vert \tilde{\mathcal{D}}]_{SLN}$ is also a monotone increasing function of noise rate $\rho$. 
		% $$ \frac{\rho(1-2\eta(\mathbf{x}))^2}{(\eta(\mathbf{x}) + \rho(1-2\eta(\mathbf{x})))(1-\eta(\mathbf{x}) - \rho(1-2\eta(\mathbf{x})))}$$ 
		
		%\textcolor{blue}{the above should have finite mean (integrable)? if so, perhaps, we can drop some terms to upper bound this so that the resulting upper bound is integrable??}
		%  {If we assume that $\rho < 1$, then using the result at \url{https://math.stackexchange.com/questions/217702/when-can-we-interchange-the-derivative-with-an-expectation}, we know that there exists a function which will upper the derivative of KL w.r.t $\rho$ and hence we can exchange the expectation and derivative.}
		%https://math.stackexchange.com/questions/217702/when-can-we-interchange-the-derivative-with-an-expectation
	\end{proof}
	\subsection{Proof of Lemma \ref{lem: KL_CCN}} \label{subsec: KL_CCN_proof}
	\begin{proof}
		Let the joint density corresponding to clean ($\mathcal{D}$) and noisy distribution ($\tilde{\mathcal{D}}$) be $p_{c}(\mathbf{x},y)$ and $p_{n}(\mathbf{x},y)$ respectively. Then, similar to Eq. (\ref{eq: KL_withexpSLN}) we have,
		\begin{eqnarray*} 
			KL[\mathcal{D}\Vert \tilde{\mathcal{D}}]_{CCN} = \mathbb{E}_{\mathbf{X}}[ KL(p_{c}(Y | \mathbf{X}) \Vert p_{n}(Y | \mathbf{X})) ], \\
			= \mathbb{E}_{\mathbf{X}} \left[\eta(\mathbf{x}) \log \frac{\eta(\mathbf{x})}{\tilde{\eta}(\mathbf{x})}
			+ (1-\eta(\mathbf{x})) \log \frac{1 - \eta(\mathbf{x})}{1 - \tilde{\eta}(\mathbf{x})}\right].
		\end{eqnarray*}
		Since, the in-class probability relation for CCN case is $\tilde{\eta}(\mathbf{x}) = (1-\rho_+ - \rho_-)\eta(\mathbf{x}) + \rho_-$, we have,
		\begin{eqnarray} \nonumber
		KL(p_{c}(Y | \mathbf{X} = \mathbf{x}) \Vert p_{n}(Y | \mathbf{X}= \mathbf{x}))  \\ \nonumber
		= -\eta(\mathbf{x}) \log \left(1-  \rho_+ -\rho_- \left( 1 - \frac{1}{\eta(\mathbf{x})}\right)\right) \\ \label{eq: KL_CCN_cond}
		- (1-\eta(\mathbf{x}))\log \left(1- \rho_- + \rho_+ \left(\frac{\eta(\mathbf{x})}{1 - \eta(\mathbf{x})}\right)\right).
		\end{eqnarray}
		Taking the derivative of Eq. (\ref{eq: KL_CCN_cond}) w.r.t. $\rho_+$, we get
		% 	\begin{scriptsize}
		\begin{eqnarray*}
			\eta(\mathbf{x})\left[\frac{1}{\left(1-  \rho_+ -\rho_- \left( 1 - \frac{1}{\eta(\mathbf{x})}\right)\right)} - \frac{1}{\left(1- \rho_- + \rho_+ \left(\frac{\eta(\mathbf{x})}{1 - \eta(\mathbf{x})}\right)\right)} \right]
		\end{eqnarray*}
		\begin{eqnarray} \label{eq: simp_der_CCN_rho+}
		=\left[ \frac{\eta(\mathbf{x}) (\eta(\mathbf{x})\rho_+ - (1-\eta(\mathbf{x}))\rho_-) }{(1-\eta(\mathbf{x}))\left(1-  \rho_+ -\rho_- \left( 1 - \frac{1}{\eta(\mathbf{x})}\right)\right) \left(1- \rho_- + \rho_+ \left(\frac{\eta(\mathbf{x})}{1 - \eta(\mathbf{x})}\right)\right) } \right].
		\end{eqnarray}
		% 	\end{scriptsize}
		% This derivative is positive if $\rho_+ \geq \frac{\rho_-(1- \eta(\mathbf{x}))}{\eta(\mathbf{x})}, \forall \mathbf{x}$. \textcolor{blue}{not very clear}
		Next, taking derivative w.r.t $\rho_-$, we have,
		% 	\begin{scriptsize}
		\begin{eqnarray*}
			(1-\eta(\mathbf{x}))\left[\frac{-1}{\left(1-  \rho_+ -\rho_- \left( 1 - \frac{1}{\eta(\mathbf{x})}\right)\right)} + \frac{1}{\left(1- \rho_- + \rho_+ \left(\frac{\eta(\mathbf{x})}{1 - \eta(\mathbf{x})}\right)\right)} \right]
		\end{eqnarray*}
		\begin{eqnarray} \label{eq: simp_der_CCN_rho-}
		=\left[ \frac{(1-\eta(\mathbf{x})) ((1-\eta(\mathbf{x}))\rho_- - \eta(\mathbf{x})\rho_+) }{\eta(\mathbf{x})\left(1-  \rho_+ -\rho_- \left( 1 - \frac{1}{\eta(\mathbf{x})}\right)\right) \left(1- \rho_- + \rho_+ \left(\frac{\eta(\mathbf{x})}{1 - \eta(\mathbf{x})}\right)\right) } \right].
		\end{eqnarray}
		% 	\end{scriptsize}
		% This derivative is positive if 
		% $\rho_- \geq \frac{\rho_+ \eta(\mathbf{x})}{(1-\eta(\mathbf{x}))}, \forall \mathbf{x}.$
		Clearly, from Eq. (\ref{eq: simp_der_CCN_rho+}) and (\ref{eq: simp_der_CCN_rho-}) both the derivatives w.r.t. $\rho_+$ and $\rho_-$ cannot be positive simultaneously. Hence, the projection approach of showing monotonicity of a function of two variables fails. Next, for the total derivative approach along the diagonal, we take the inner product of the derivatives with the direction $h = (1,1)$, and obtain the following,
		% 	\begin{scriptsize}
		\begin{eqnarray*}
			\frac{(2\eta(\mathbf{x})-1)(\eta(\mathbf{x})\rho_+ - (1-\eta(\mathbf{x}))\rho_-)}{\eta(\mathbf{x})(1-\eta(\mathbf{x}))\left(1-  \rho_+ -\rho_- \left( 1 - \frac{1}{\eta(\mathbf{x})}\right)\right) \left(1- \rho_- + \rho_+ \left(\frac{\eta(\mathbf{x})}{1 - \eta(\mathbf{x})}\right)\right)}.
		\end{eqnarray*}
		% 	\end{scriptsize}
		This total derivative can be negative for some $\mathbf{x}$ and positive for some other $\mathbf{x}$ and hence, $KL[\mathcal{D}\Vert \tilde{\mathcal{D}}]_{CCN}$ is not an increasing function of noise rates $\rho_+$ and  $\rho_-$.
		% Combining the lower bounds on the noise rates and taking them over all $\mathbf{x}$, we obtain the conditions given in Eq. (\ref{eq: rho_cond_CCN}).
		% {to relate this to the higher noise rate}
	\end{proof}
	
	\section{Additional computational experiments}
	In this section, we provide empirical evidence for the various observations made in the main paper. First, we empirically show the implications of Lemma \ref{lem: KL_SLN }. Next, we show how change in representation of the data lead to improvement in the basic setup. Finally, we demonstrate our claim about some variants of GAN which are not good for generating correctly labelled data, even though in the conventional GAN setup they work well.
	
	\subsection{Dataset sizes and architecture details of different neural networks used}  \label{subsec: data_arc_details}
	In this section, we provide the synthetic data generation scheme, our approach of constructing binary datasets and the architecture of the neural networks used. 
	
	\subsubsection{Synthetic dataset generation}
	For synthetic data generation, we use the method given by \cite{efron1997improvements}. We first generate $6000$ train and $750$ test binary class labels $Y \sim Bern(0.5)$. Next, given feature dimension $n$, we generate two $n$ dimensional mean vectors $\boldsymbol{\mu}_+ = (\mu_{+,1},\cdots, \mu_{+,n}), ~ \boldsymbol{\mu}_- = (\mu_{-,1},\cdots, \mu_{-,n})$ as follows:  for every dimension $j = 1,\cdots,n$, draw a sample from Unif(-2,2) and assign it to $\mu_{+,j}$; next draw a sample from Bern(0.4), if it is 1, $\mu_{-,j} = \mu_{+,j}$ else $\mu_{-,j} = -\mu_{+,j}$. Also, the covariance matrix $\Sigma$ is such that all the diagonal elements (variances) are $8$ and the non-diagonal elements (covariances) are $0$. Now, a n-dimensional feature vector $\mathbf{X}$ for each label $Y$ is drawn from two different Gaussian distributions: $\mathbf{X}|Y =1 \sim N(\boldsymbol{\mu}_+,\Sigma)$ $\&$ $\mathbf{X}|Y =-1 \sim N(\boldsymbol{\mu}_-,\Sigma)$. This process is repeated for both $n=100$ and $n=300$ to obtain the synthetic datasets SD100 and SD300 used in Section \ref{sec: exper}.
	
	\subsubsection{Real datasets} 
	To obtain the binary versions, say class (a,b) of multi-class MNIST and Fashion MNIST datasets, we combine class $a$ and class $b$ examples only to obtain the final binary datasets. In case of imbalanced dataset with imbalance ratio $imb\_r = r$, we randomly sample $100*r$ percent of the data points from class $a$ and $100*(1-r)$ percent of the data points from class $b$. In case when one of the class is $0$, we use the following strategy: if $a (b) = 0$, we sample $100*r\%$ from class $b(a)$ and rest $100*(1-r)\%$ from class $a(b)$. The exact sample sizes for various datasets used in Section \ref{sec: exper} are given in Table \ref{tab: data_size}.
	
	\begin{table}[]
		\centering
		\bgroup
		\def\arraystretch{0.85}% 
		{ \footnotesize  \setlength{\tabcolsep}{0.25em}
			\begin{tabular}{|c|c|c|c|}
				\hline
				\textbf{S. no.} & \textbf{Dataset name} & \textbf{$m_{tr} (p_{tr}, n_{tr})$} & \textbf{$m_{te} (p_{te}, n_{te})$} \\ \hline
				1 & MNIST 5-6 & 10404 (4987, 5417) & 1850 (892, 958) \\ \hline
				2 & MNIST 7-9 & 11169 ( 5715, 5454) & 2037 (1028, 1009) \\ \hline
				3 & MNIST 4-9 & 10761 (5307, 5454) & 1991 (982, 1009) \\ \hline
				4 & MNIST 1-7 & 11894 (6179, 5715) & 2163 (1135, 1028) \\ \hline
				5 & MNIST 0-6 & 10861 (5417, 5444) & 1938 (958, 980) \\ \hline
				6 & MNIST 0-1 & 11623 (6179, 5444) & 2115 (1135, 980) \\ \hline
				7 & \begin{tabular}[c]{@{}c@{}}Fashion MNIST 7-9\\ (Sneakers and Boots)\end{tabular} & 12000 (6000, 6000) & 2000 (1000, 1000) \\ \hline
				8 & \begin{tabular}[c]{@{}c@{}}Fashion MNIST 2-4\\ (Pullover and Coat)\end{tabular} & 12000 (6000, 6000) & 2000 (1000, 1000) \\ \hline
				9 & \begin{tabular}[c]{@{}c@{}}Fashion MNIST 0-6\\ (Tshirt and Shirt)\end{tabular} & 12000 (6000, 6000) & 2000 (1000, 1000) \\ \hline
				10 & \begin{tabular}[c]{@{}c@{}}Fashion MNIST 2-3\\ (Pullover and Dress)\end{tabular} & 12000 (6000, 6000) & 2000 (1000, 1000) \\ \hline
				11 & MNIST 0-8 (imb\_r = 0.1) & 5437 (538, 4899) & 979 (97,882) \\ \hline
				12 & MNIST 0-8 (imb\_r = 0.8) & 5399 (4311, 1088) & 974 (779, 195) \\ \hline
				13 & MNIST 1-7 (imb\_r = 0.2) & 5807 (1235, 4572) & 1049 (222, 822) \\ \hline
				14 & MNIST 1-7 (imb\_r = 0.7) & 6039 (4325, 1714) & 1102 (794, 308) \\ \hline
				15 & MNIST 4-9 (imb\_r = 0.3) & 5409 (1592, 3817) & 1000 (294, 706) \\ \hline
				16 & MNIST 4-9 (imb\_r = 0.75) & 5343 (3980, 1363) & 988 (736, 252) \\ \hline
		\end{tabular}}
		\egroup
		\caption{Dataset details for the binary versions of MNIST and Fashion MNIST datasets.  $m_{tr}$ and $m_{te}$ denote the number of train and test data points. $p_{tr} (n_{tr})$ and  $p_{te} (n_{te})$ denote number of positive (negative) data points in the training and test datasets. Dataset number 1 to 10 are balanced and  11-16 are imbalanced with imbalance ratio  $imb\_r$.}
		\label{tab: data_size}
	\end{table}
	
	\subsubsection{Architecture details} We have used multilayer perceptrons for the generator $G_{dn}$, discriminator $D_{dn}$.The final classification model $M$ can either be an MLP or a CNN. The exact details are provided in Table \ref{tab: archi1}, Table \ref{tab: archi2} and Table \ref{tab: CNN_architecture}. 
% 	A schematic depicting the overall scheme of WGAN based class conditional label noise robust classification is presented in Figure \ref{fig : schematic_WGAN*}.
	
	\begin{table}[!htbp]
		\centering
		\bgroup
		\def\arraystretch{0.85}% 
		{ \footnotesize \setlength{\tabcolsep}{0.25em}
			\begin{tabular}{|c|c|c|ccc}
				\hline
				\multicolumn{3}{|c|}{\textbf{Generator}} & \multicolumn{3}{c|}{\textbf{Discriminator}} \\ \hline
				\textbf{Layer} & \textbf{Neurons} & \textbf{Activation} & \multicolumn{1}{c|}{\textbf{Layer}} & \multicolumn{1}{c|}{\textbf{Neurons}} & \multicolumn{1}{c|}{\textbf{Activation}} \\ \hline
				Input & 785+4 & - & \multicolumn{1}{c|}{Input} & \multicolumn{1}{c|}{785+4} & \multicolumn{1}{c|}{-} \\ \hline
				Fully connected & 64 & Relu & \multicolumn{1}{c|}{Fully connected} & \multicolumn{1}{c|}{128} & \multicolumn{1}{c|}{Relu} \\ \hline
				Fully connected & 128 & Relu & \multicolumn{1}{c|}{Fully connected} & \multicolumn{1}{c|}{128} & \multicolumn{1}{c|}{Relu} \\ \hline
				Fully connected & 128 & Relu & \multicolumn{1}{c|}{Output} & \multicolumn{1}{c|}{1} & \multicolumn{1}{c|}{Linear} \\ \hline
				Output & 785+4 & - &  &  &  \\ \cline{1-3}
		\end{tabular}}
		\egroup
		\caption{The table depicts the architecture of generator and discriminator used in all the computations. For the generator, the last layer has extra 4 dimensions appended once the label has been decided after majority voting.}
		\label{tab: archi1}
	\end{table}
	
	\begin{table}[]
		\centering
		\bgroup
		\def\arraystretch{0.85}% 
		{ \footnotesize \setlength{\tabcolsep}{0.25em}
			\begin{tabular}{|c|c|c|ccc}
				\hline
				\multicolumn{3}{|c|}{\textbf{Classification model M (MLP)}} & \multicolumn{3}{c|}{\textbf{GLC network (MLP)}} \\ \hline
				\textbf{Layer} & \textbf{Neurons} & \textbf{Activation} & \multicolumn{1}{c|}{\textbf{Layer}} & \multicolumn{1}{c|}{\textbf{Neurons}} & \multicolumn{1}{c|}{\textbf{Activation}} \\ \hline
				Input & 784 & - & \multicolumn{1}{c|}{Input} & \multicolumn{1}{c|}{784} & \multicolumn{1}{c|}{-} \\ \hline
				Fully connected & 185 & Relu & \multicolumn{1}{c|}{Fully connected} & \multicolumn{1}{c|}{128} & \multicolumn{1}{c|}{Relu} \\ \hline
				Fully connected & 200 & Relu & \multicolumn{1}{c|}{Fully connected} & \multicolumn{1}{c|}{128} & \multicolumn{1}{c|}{Relu} \\ \hline
				Fully connected & 185 & Relu & \multicolumn{1}{c|}{Output} & \multicolumn{1}{c|}{2} & \multicolumn{1}{c|}{Softmax} \\ \hline
				Output & 2 & Softmax &  &  &  \\ \cline{1-3}
		\end{tabular}}
		\egroup
		\caption{The table provides the architecture of classification network $M$ which is an MLP. It is used by SimpleNN, GCE and LDMI to train on noisy data and our schemes to train on the correctly labelled data points generated by the generator. GLC network is used by the GLC \cite{hendrycks2018using} method for classification.}
		\label{tab: archi2}
	\end{table}
	
\begin{table}[!htbp]
\centering
{ \footnotesize
\begin{tabular}{|c|c|c|}
\hline
\multicolumn{3}{|c|}{\textbf{Classification model M (CNN)}} \\ \hline
\textbf{Layer} & \textbf{Filter/kernel size} & \textbf{Activation} \\ \hline
Input & 1*28*28 & - \\ \hline
Conv2D & \begin{tabular}[c]{@{}c@{}}In channel=1, out channel =16, kernel size = 5, \\ Maxpool (kernel size =2)\end{tabular} & Relu \\ \hline
Conv2D & \begin{tabular}[c]{@{}c@{}}In channel=16, out channel =32, kernel size = 5, \\ Maxpool(kernel size =2), Dropout (p=0.5)\end{tabular} & Relu \\ \hline
Output & 2 & Linear \\ \hline
\end{tabular}}
\caption{The table provides the architecture of classification network M which is a CNN. It is used by Simple CNN, GLC, GCE and LDMI to train on noisy data and our schemes to train on the correctly labelled data points generated by the generator.}
\label{tab: CNN_architecture}
\end{table}
	
	\subsection{Experiments to show that increase in SLN rate $\rho$ leads to improvement in performance of WGANXtraY} \label{subsec: exp_SLN_rho_inc}
	In this section, we present empirical evidence for the implications of Lemma \ref{lem: KL_SLN } in Section \ref{subsec: ourscheme} using the experimental results from 4 binary classification image datasets,viz., MNIST 1-7, MNIST 0-6, MNIST 4-9, and Fashion-MNIST 2-3. As can be seen in Table \ref{tab: SLN_incr_rho_inc_per1} and \ref{tab: SLN_incr_rho_inc_per2}, relative performance of WGANXtraY(5) improves with increase in noise rate $\rho$. The size of clean dataset used is $0.1\%$ of the total training dataset for both GLC and our scheme. Number of outer loop iterations $n_{it} = 1000.$ To account for the randomness while inducing noise, we present averaged accuracy and standard deviation across 5 trials.
	\begin{table}[]
		\centering
		\bgroup
		\def\arraystretch{0.85}% 
		{ \footnotesize \setlength{\tabcolsep}{0.25em}
			\begin{tabular}{|c|c|c|c|c|}
				\hline
				\textbf{$\rho$} & \multicolumn{2}{c|}{\textbf{MNIST 1-7 (0.1\% clean)}} & \multicolumn{2}{c|}{\textbf{MNIST 0-6 (0.1\% clean)}} \\ \hline
				\textbf{} & \textbf{WGANXtraY(5)} & \textbf{GLC} & \textbf{WGANXtraY(5)} & \textbf{GLC} \\ \hline
				\textbf{0.45} & 93.90 $\pm$ 1.88 & \textbf{94.59 $\pm$ 3.32} & 85.20$\pm$ 4.31 & \textbf{89.39 $\pm$ 6.81} \\ \hline
				\textbf{0.46} & \textbf{95.92 $\pm$ 0.30} & 91.63 $\pm$ 6.57 & \textbf{91.86 $\pm$ 1.49} & 84.07$\pm$ 8.79 \\ \hline
				\textbf{0.47} & \textbf{94.88$\pm$ 1.32} & 84.49$\pm$ 8.94 & \textbf{90.33 $\pm$ 1.67} & 83.85 $\pm$ 10.18 \\ \hline
				\textbf{0.48} & \textbf{94.84$\pm$ 1.58} & 82.61 $\pm$ 16.51 & 87.88 $\pm$ 2.89 & \textbf{88.10 $\pm$ 4.56} \\ \hline
				\textbf{0.49} & \textbf{93.58 $\pm$ 1.87} & 76.36 $\pm$ 11.02 & \textbf{89.90 $\pm$ 0.91} & 84.81 $\pm$ 5.05 \\ \hline
		\end{tabular}}
		\egroup
		\caption{Average accuracy and standard deviation over 5 trials on MNIST 1-7 and MNIST 0-6 datasets when the classification model M is an MLP. Table values demonstrates the relative improvement of WGANXtraY(5) over GLC as the SLN rates are increased. This phenomenon is due to increase in adversarial relation (measured in terms of KL divergence between $\mathcal{D}$ and $\tilde{\mathcal{D}}$) as the noise rate increases. Bold-faced values denote highest accuracy values in a row.}
		\label{tab: SLN_incr_rho_inc_per1}
	\end{table}
	
	\begin{table}[]
		\centering
		\bgroup
		\def\arraystretch{0.85}% 
		{ \footnotesize \setlength{\tabcolsep}{0.25em}
			\begin{tabular}{|c|c|c|c|c|}
				\hline
				\textbf{$\rho$} & \multicolumn{2}{c|}{\textbf{MNIST 4-9 (0.1\% clean)}} & \multicolumn{2}{c|}{\textbf{Fas-MNIST 2-3 (0.1\% clean)}} \\ \hline
				\textbf{} & \textbf{WGANXtraY(5)} & \textbf{GLC} & \textbf{WGANXtraY(5)} & \textbf{GLC} \\ \hline
				\textbf{0.45} & 74.64 $\pm$ 5.32 & \textbf{76.66 $\pm$ 7.23} & \textbf{96.03 $\pm$ 0.51} & 95.55 $\pm$ 1.29 \\ \hline
				\textbf{0.46} & 72.07 $\pm$ 4.64 & \textbf{79.38 $\pm$ 5.01} & \textbf{86.69 $\pm$ 17.31} & 84.5 $\pm$ 14.12 \\ \hline
				\textbf{0.47} & \textbf{72.73 $\pm$ 2.76} & 63.21 $\pm$ 11.00 & \textbf{95.97 $\pm$ 0.90} & 88.56 $\pm$ 16.43 \\ \hline
				\textbf{0.48} & \textbf{76.67 $\pm$ 2.25} & 64.74 $\pm$ 11.11 & \textbf{94.16 $\pm$ 2.64} & 90.61 $\pm$ 7.06 \\ \hline
				\textbf{0.49} & \textbf{74.20 $\pm$ 2.42} & 58.29 $\pm$ 9.00 & \textbf{92.75 $\pm$ 2.15} & 85.25 $\pm$ 15.92 \\ \hline
		\end{tabular}}
		\egroup
		\caption{Average accuracy and standard deviation over 5 trials on MNIST 4-9 and Fashion-MNIST 2-3 (Pullover and Dress) datasets when the classification model M is an MLP. Table values demonstrates the relative improvement of WGANXtraY(5) over GLC as the SLN rates are increased. This phenomenon is due to increase in adversarial relation (measured in terms of KL divergence between $\mathcal{D}$ and $\tilde{\mathcal{D}}$) as the noise rate increases. Bold-faced values denote higher accuracy values.}
		\label{tab: SLN_incr_rho_inc_per2}
	\end{table}
	
	\subsection{Experiments to show bad performance of WGANY} \label{subsec: WGANY_bad_perf}
\begin{figure*}
		\centering
		\includegraphics[width = 0.8\textwidth]{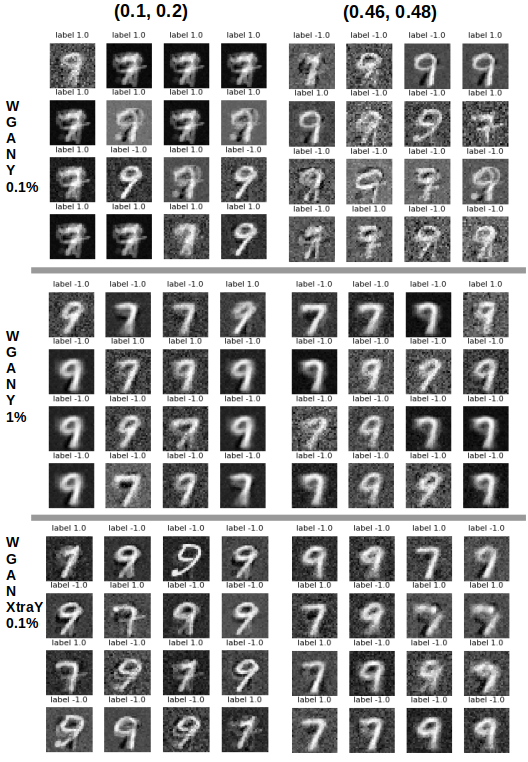}
		\caption{The feature-label pairs generated by WGANY and WGANXtraY(5) when the noise rates were $(0.1,0.2)$ and $(0.46,0.48)$. 7 is positive class and 9 is negative class. First two rows have more inconsistency in the labels assigned to the generated images than last row, demonstrating the improvement due to WGANXtraY(5). Percentage of clean data used is mentioned along with the images.}
		\label{fig: MNIST_7-9_WGNAY_bad}
	\end{figure*}

\begin{figure*}
		\centering
		\includegraphics[width = 0.8\textwidth]{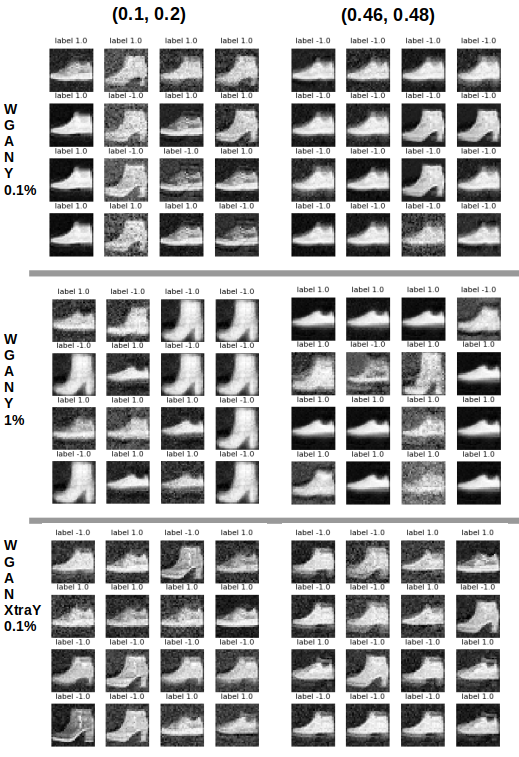}
		\caption{The feature-label pairs generated by WGANY and WGANXtraY(5) when the noise rates were $(0.1,0.2)$ and $(0.46,0.48)$. Sneakers (7) is positive class and Boots (9) is negative class. First two rows have more inconsistency in the labels assigned to the generated images than last row, demonstrating the improvement due to WGANXtraY(5). Percentage of clean data used is mentioned along with the images.}
		\label{fig: Fashion_MNIST_7-9_WGNAY_bad}
	\end{figure*}

	In this section, we present the experimental results showing that both the generated image quality and the performance of final classifier is not good for WGANY (scheme without representation changes). We demonstrate this using 2 binary datasets, viz., MNIST 7-9 and Fashion-MNIST 7-9 (Sneakers and Boots).
	% We consider 2 cases with clean set size as $0.1\%$ and $1\%$ of the total dataset. We compare this to WGANXtraY(5) and observe that even using $0.1\%$ is sufficient to see an improvement over WGANY. 
	As can be seen from Table \ref{tab:Mnist_7-9_WGAN_not_work} and \ref{tab:Fashion_Mnist_7-9_WGAN_not_work}, WGANY has low accuracies in comparison to WGANXtraY(5) across all CCN rates. This situation doesn't improve even when the clean sample size is increased to $1\%$, as can be seen by comparing column 3 and 4 of Table \ref{tab:Mnist_7-9_WGAN_not_work} and \ref{tab:Fashion_Mnist_7-9_WGAN_not_work}. In addition to the accuracy, we show that the correct pairing of feature and label is bad for WGANY but improves by using WGANXtraY(5) as shown in Figure \ref{fig: MNIST_7-9_WGNAY_bad} and Figure \ref{fig: Fashion_MNIST_7-9_WGNAY_bad}.
	
	\begin{table}[]
		\centering
		\bgroup
		\def\arraystretch{0.85}% 
		{ \scriptsize \setlength{\tabcolsep}{0.25em}
			\begin{tabular}{|c|c|c|c|}
				\hline
				\textbf{$\rho_+,\rho_-$} & \multicolumn{2}{c|}{\textbf{WGANY}} & \textbf{WGANXtraY(5)} \\ \hline
				\textbf{clean \%} & \textbf{0.1\%} & \textbf{1\%} & \textbf{0.1\%} \\ \hline
				\textbf{0.1,0.2} & 53.03 $\pm$ 11.67 & 60.03 $\pm$ 7.11 & \textbf{77.20 $\pm$ 1.36} \\ \hline
				\textbf{0.2,0.49} & 54.22$\pm$ 5.52 & 44.82 $\pm$ 14.45 & \textbf{80.88 $\pm$ 1.85} \\ \hline
				\textbf{0.42,0.35} & 53.49$\pm$ 3.98 & 43.93 $\pm$ 11.67 & \textbf{81.90 $\pm$ 2.23} \\ \hline
				\textbf{0.3,0.3} & 58.46 $\pm$ 10.84 & 64.81$\pm$ 17.49 & \textbf{80.56 $\pm$ 5.04} \\ \hline
				\textbf{0.4,0.49} & 54.00 $\pm$ 10.31 & 59.73$\pm$ 7.33 & \textbf{78.8$\pm$ 1.33} \\ \hline
				\textbf{0.42,0.45} & 51.71 $\pm$ 3.81 & 56.14 $\pm$ 17.99 & \textbf{75.9$\pm$ 3.87} \\ \hline
				\textbf{0.45,0.46} & 47.94 $\pm$ 4.13 & 63.05 $\pm$ 11.70 & \textbf{81.49$\pm$ 1.2} \\ \hline
				\textbf{0.45,0.48} & 49.19 $\pm$ 1.19 & 58.85 $\pm$ 17.48 & \textbf{81.1$\pm$ 2.56} \\ \hline
				\textbf{0.45,0.5} & 51.22 $\pm$ 3.12 & 51.12 $\pm$ 7.14 & \textbf{78.0$\pm$ 4.2} \\ \hline
				\textbf{0.46,0.48} & 49.23 $\pm$ 12.08 & 50.08 $\pm$ 19.61 & \textbf{80.79 $\pm$ 2.37} \\ \hline
				\textbf{0.46,0.5} & 54.11 $\pm$ 9.75 & 52.45 $\pm$ 10.29 & \textbf{80.3 $\pm$ 1.88} \\ \hline
				\textbf{0.47,0.49} & 48.89 $\pm$ 2.02 & 51.84 $\pm$ 9.44 & \textbf{76.9$\pm$ 4.54} \\ \hline
				\textbf{0.47,0.51} & 52.00 $\pm$ 2.52 & 53.42 $\pm$ 15.04 & \textbf{78.9$\pm$ 1.70} \\ \hline
				\textbf{0.48,0.49} & 53.18 $\pm$ 6.84 & 50.20 $\pm$ 7.09 & \textbf{80.7 $\pm$ 2.07} \\ \hline
				\textbf{0.5,0.5} & 51.69$\pm$ 3.16 & 51.59$\pm$ 7.07 & \textbf{79.2$\pm$ 1.51} \\ \hline
		\end{tabular}}
		\egroup
		\caption{Averaged accuracy and standard deviation values for MNIST 7-9 dataset for 5 trials when the classification model M is an MLP. The table  demonstrates that WGANY doesn't perform well and WGANXtraY(5) show a significant improvement (last column) over WGANY even when clean data percent is only 0.1\%.  The number of iterations $n_{it}$ is 1000 in all 3 cases. }
		\label{tab:Mnist_7-9_WGAN_not_work}
	\end{table}
	
	\begin{table}[]
		\centering
		\bgroup
		\def\arraystretch{0.85}% 
		{ \scriptsize \setlength{\tabcolsep}{0.25em}
			\begin{tabular}{|c|c|c|c|}
				\hline
				\textbf{$\rho_+,\rho_-$} & \multicolumn{2}{c|}{\textbf{WGANY}} & \textbf{WGANXtraY(5)} \\ \hline
				\textbf{clean \%} & \textbf{0.1\%} & \textbf{1\%} & \textbf{0.1\%} \\ \hline
				\textbf{0.1,0.2} & 49.14 $\pm$ 12.72 & 61.48 $\pm$ 19.92 & \textbf{87.76$\pm$ 1.28} \\ \hline
				\textbf{0.2,0.49} & 35.78 $\pm$ 17.54 & 63.27 $\pm$ 12.65 & \textbf{87.18 $\pm$ 1.80} \\ \hline
				\textbf{0.42,0.35} & 42.36 $\pm$ 21.24 & 59.92$\pm$ 11.73 & \textbf{85.94 $\pm$ 2.08} \\ \hline
				\textbf{0.3,0.3} & 70.82$\pm$ 17.49 & 45.67 $\pm$ 8.39 & \textbf{85.56 $\pm$ 4.88} \\ \hline
				\textbf{0.4,0.49} & 57.57 $\pm$ 15.13 & 61.25 $\pm$ 15.53 & \textbf{86.35 $\pm$ 2.02} \\ \hline
				\textbf{0.42,0.45} & 35.56 $\pm$ 24.41 & 59.06 $\pm$ 5.95 & \textbf{78.66 $\pm$ 14.43} \\ \hline
				\textbf{0.45,0.46} & 40.23 $\pm$ 20.11 & 48.39 $\pm$ 30.03 & \textbf{87.16 $\pm$ 1.03} \\ \hline
				\textbf{0.45,0.48} & 55.61 $\pm$ 7.01 & 48.39 $\pm$ 18.08 & \textbf{87.95 $\pm$ 1.41} \\ \hline
				\textbf{0.45,0.5} & 46.44 $\pm$ 28.41 & 26.97 $\pm$ 20.02 & \textbf{87.38 $\pm$ 1.74} \\ \hline
				\textbf{0.46,0.48} & 48.47 $\pm$ 7.18 & 45.74 $\pm$ 8.59 & \textbf{87.53 $\pm$ 0.69} \\ \hline
				\textbf{0.46,0.5} & 49.07 $\pm$ 22.59 & 51.48 $\pm$ 29.91 & \textbf{82.55 $\pm$ 10.03} \\ \hline
				\textbf{0.47,0.49} & 49.2 $\pm$ 1.59 & 64.57$\pm$ 13.01 & \textbf{85.85 $\pm$ 2.81} \\ \hline
				\textbf{0.47,0.51} & 52.25 $\pm$ 22.08 & 45.01 $\pm$ 23.33 & \textbf{70.55 $\pm$ 35.29} \\ \hline
				\textbf{0.48,0.49} & 45.91 $\pm$ 10.40 & 51.01 $\pm$ 11.79 & \textbf{86.07 $\pm$ 1.32} \\ \hline
				\textbf{0.5,0.5} & 55.1 $\pm$ 10.2 & 26.9 $\pm$ 23.36 & \textbf{85.61 $\pm$ 1.71} \\ \hline
		\end{tabular}}
		\egroup
		\caption{Averaged accuracy and standard deviation values for Fashion-MNIST 7-9 (Sneakers and Boots)  dataset for 5 trials when the classification model M is an MLP. The table  demonstrates that WGANY doesn't perform well and WGANXtraY(5) show a significant improvement (last column) over WGANY even when clean data percent is only 0.1\%. The number of iterations $n_{it}$ is 1000 in all 3 cases. }
		\label{tab:Fashion_Mnist_7-9_WGAN_not_work}
	\end{table}
	
	\subsection{Experiments to show that good GANs for generating samples (feature vector/images) need not be good for generating correctly labelled data} \label{subsec: GAN_whichdint_work}
	In this section, we present empirical evidence to show what choices of GAN (good for generating features (images) only) did not work for us with reasons given in Section \ref{sssec: goodGAN_not}. We verify this on 4 datasets, viz., MNIST 7-9, MNIST 1-7, Fashion-MNIST 7-9 and Fashion-MNIST 2-3 and the results are presented in Table \ref{tab: Mnist7-9_GAN_not_work}, \ref{tab: Mnist1-7_GAN_not_work}, \ref{tab: Fashion_Mnist7-9_GAN_not_work} and \ref{tab: Fashion_Mnist2-3_GAN_not_work}. We present the results for WGANXtraY(5) with Gradient Penalty with parameter $\lambda = 10$ (WGANXtraY-GP), PacGAN adapted to our setup with packing number as 3 (PacGAN(3)), and a variant of WGANXtraY with reconstruction loss (pixel-wise mean squared error between the images) added to the generator objective (WGANXraY-R). The last reconstruction based variant is enforcing the generator $G_{dn}$ that for a corrupted data point $(\mathbf{x},\tilde{y})$, the new data point $(\mathbf{x}',y')$ should have $\mathbf{x}'$ close to $\mathbf{x}$ along with correct label $y'$. After trying different weights for the reconstruction loss term in the generator, we use weight value of 1 as it worked best for us. Also, since, for PacGAN(3) and WGANXtraY-GP using $0.1\%$ gold fraction lead to issue of the minibatch size being more than the size of clean labelled dataset, we use $1\%$ clean data for the computations presented. We continue to take $k=5$ for all WGANXtraY based schemes. We compare the accuracies on final classification model trained on correctly labelled generated data from the schemes aforementioned  and WGANXtraY. The number of iterations $n_{it} = 5000$ for PacGAN(3) and WGANXtraY-GP and $n_{it} = 1000$ for WGANXtraY and WGANXtraY-R. The batch size is taken to be 64 for all schemes except for PacGAN(3) for which batch size is 32.
	
	Clearly, PacGAN(3) adapted for the task of generating correctly labelled data from noisy data doesn't work as seen by the very low accuracy values of the final classification model in all 4 datasets. Even though, WGANXtraY-R is comparable in some cases to WGANXtraY, the variation as quantified by the standard deviation across 5 trials is very high in almost all cases. WGANXtraY-GP performs comparably to WGANXtraY in some cases and better than WGANXtraY in other cases for Fashion MNIST 7-9 and 2-3 datasets. We attribute this behaviour to the use of large number of iterations ($n_{it} = 5000$). Since, we wanted to use minimal number of clean labels and minimal iterations we chose WGANXtraY over WGANXtraY-GP.
	
	\begin{table}[]
		\centering
		\bgroup
		\def\arraystretch{0.85}% 
		\scriptsize{ \setlength{\tabcolsep}{0.25em}
			\begin{tabular}{|c|l|c|c|c|}
				\hline
				\textbf{$\rho_+,\rho_-$} & \textbf{WGANXtraY} & \textbf{WGANXtraY-GP} & \textbf{WGANXtraY-R} & \textbf{PacGAN(3)} \\ \hline
				\textbf{0.4,0.49} & \textbf{90.89 $\pm$ 1.85} & 81.07 $\pm$ 14.96 & 84.00 $\pm$ 2.84 & 49.30 $\pm$ 2.24 \\ \hline
				\textbf{0.42,0.45} & \textbf{89.31 $\pm$ 1.36} & 80.93 $\pm$ 15.39 & 87.43 $\pm$ 2.85 & 50.37$\pm$ 1.27 \\ \hline
				\textbf{0.45,0.46} & \textbf{89.13 $\pm$ 2.40} & 86.88 $\pm$ 4.22 & 85.70$\pm$ 2.40 & 48.72 $\pm$ 4.04 \\ \hline
				\textbf{0.45,0.48} & \textbf{88.84 $\pm$ 3.09} & 79.87$\pm$ 15.19 & 85.43$\pm$ 4.53 & 53.96 $\pm$ 5.11 \\ \hline
				\textbf{0.45,0.5} & \textbf{90.01 $\pm$ 1.99} & 88.64$\pm$ 2.86 & 85.42$\pm$ 4.62 & 52.02$\pm$ 5.17 \\ \hline
				\textbf{0.46,0.48} & \textbf{89.11 $\pm$ 2.50} & 85.10 $\pm$ 4.55 & 83.69 $\pm$ 4.06 & 39.07 $\pm$ 21.75 \\ \hline
				\textbf{0.46,0.5} & \textbf{89.35 $\pm$ 0.81} & 88.76 $\pm$ 2.01 & 86.02 $\pm$ 4.57 & 30.09$\pm$ 24.57 \\ \hline
				\textbf{0.47,0.49} & \textbf{90.56 $\pm$ 0.90} & 88.05 $\pm$ 3.09 & 86.33 $\pm$ 2.46 & 50.21 $\pm$ 0.59 \\ \hline
				\textbf{0.47,0.51} & 89.13 $\pm$ 1.23 & 84.75 $\pm$ 3.81 & \textbf{89.44 $\pm$ 0.87} & 39.15 $\pm$ 19.84 \\ \hline
				\textbf{0.48,0.49} & \textbf{89.53 $\pm$ 0.61} & 86.90 $\pm$ 4.54 & 88.67 $\pm$ 0.57 & 41.65 $\pm$ 21.11 \\ \hline
				\textbf{0.5,0.5} & \textbf{90.01 $\pm$ 1.19} & 85.90 $\pm$ 4.43 & 88.67 $\pm$ 0.57 & 47.41 $\pm$ 4.05 \\ \hline
		\end{tabular}}
		\egroup
		\caption{Averaged accuracy and standard deviation values for MNIST 7-9 dataset across 5 trials when the classification model M is an MLP. The table  implies that WGANXtraY is better than other schemes most of the time when clean data percent available is only 1\%. This demonstrates that good GANs (columns 3,4 and 5) for generating samples need not be good for generating correctly labelled data points from noisy labelled data points. }
		\label{tab: Mnist7-9_GAN_not_work}
	\end{table}
	
	\begin{table}[]
		\centering
		\bgroup
		\def\arraystretch{0.85}% 
		\scriptsize{ \setlength{\tabcolsep}{0.25em}
			\begin{tabular}{|c|l|c|c|c|}
				\hline
				\textbf{$\rho_+,\rho_-$} & \textbf{WGANXtraY} & \textbf{WGANXtraY-GP} & \textbf{WGANXtraY-R} & \textbf{PacGAN(3)} \\ \hline
				\textbf{0.4,0.49} & \textbf{96.06 $\pm$ 1.02} & 93.39$\pm$ 2.71 & 93.52$\pm$ 1.77 & 48.60 $\pm$ 3.55 \\ \hline
				\textbf{0.42,0.45} & 95.85 $\pm$ 1.00 & \textbf{96.23$\pm$ 1.56} & 90.81 $\pm$ 7.39 & 49.50 $\pm$ 2.42 \\ \hline
				\textbf{0.45,0.46} & \textbf{95.58 $\pm$ 2.72} & 91.68$\pm$ 5.82 & 92.21 $\pm$ 4.27 & 51.56 $\pm$ 6.41 \\ \hline
				\textbf{0.45,0.48} & \textbf{95.09 $\pm$ 1.06} & 93.34 $\pm$ 4.14 & 91.00$\pm$ 5.71 & 51.82 $\pm$ 11.04 \\ \hline
				\textbf{0.45,0.5} & \textbf{95.55 $\pm$ 1.21} & 86.13 $\pm$ 17.26 & 91.93 $\pm$ 1.82 & 54.96 $\pm$ 10.16 \\ \hline
				\textbf{0.46,0.48} & \textbf{95.95 $\pm$ 1.25} & 94.40 $\pm$ 4.63 & 90.50 $\pm$ 8.44 & 47.46 $\pm$ 5.21 \\ \hline
				\textbf{0.46,0.5} & \textbf{96.56 $\pm$ 0.45} & 82.92 $\pm$ 17.00 & 93.25$\pm$ 3.15 & 50.28 $\pm$ 3.59 \\ \hline
				\textbf{0.47,0.49} & \textbf{95.10 $\pm$ 0.56} & 93.40 $\pm$ 3.70 & 94.23 $\pm$ 1.82 & 38.65 $\pm$ 20.34 \\ \hline
				\textbf{0.47,0.51} & 95.58$\pm$ 0.35 & \textbf{95.64 $\pm$ 1.53} & 88.68 $\pm$ 9.69 & 49.07 $\pm$ 4.54 \\ \hline
				\textbf{0.48,0.49} & \textbf{96.06$\pm$ 1.69} & 94.85 $\pm$ 3.44 & 91.86 $\pm$ 1.14 & 51.50 $\pm$ 1.98 \\ \hline
				\textbf{0.5,0.5} & \textbf{96.06 $\pm$ 0.41} & 94.93 $\pm$ 3.35 & 95.04 $\pm$ 1.48 & 40.11 $\pm$ 20.19 \\ \hline
		\end{tabular}}
		\egroup
		\caption{Averaged accuracy and standard deviation values for MNIST 1-7 dataset across 5 trials when the classification model M is an MLP. The table  implies that WGANXtraY is better than other schemes most of the time when clean data percent available is only 1\%. This demonstrates that good GANs (columns 3,4 and 5) for generating samples need not be good for generating correctly labelled data points from noisy labelled data points. }
		\label{tab: Mnist1-7_GAN_not_work}
	\end{table}

	\begin{table}[]
		\centering
		\bgroup
		\def\arraystretch{0.85}% 
		\scriptsize{ \setlength{\tabcolsep}{0.25em}
			\begin{tabular}{|c|l|c|c|c|}
				\hline
				\textbf{$\rho_+,\rho_-$} & \textbf{WGANXtraY} & \textbf{WGANXtraY-GP} & \textbf{WGANXtraY-R} & \textbf{PacGAN(3)} \\ \hline
				\textbf{0.4,0.49} & \textbf{89.23$\pm$ 1.29} & 89.16 $\pm$ 1.10 & 87.52 $\pm$ 3.76 & 49.30 $\pm$ 2.24 \\ \hline
				\textbf{0.42,0.45} & \textbf{89.67 $\pm$ 0.75} & 87.45 $\pm$ 3.25 & 89.16 $\pm$ 0.57 & 50.37$\pm$ 1.27 \\ \hline
				\textbf{0.45,0.46} & \textbf{89.36 $\pm$ 0.49} & 86.01 $\pm$ 6.00 & 78.59 $\pm$ 11.83 & 48.72 $\pm$ 4.04 \\ \hline
				\textbf{0.45,0.48} & 89.92 $\pm$ 0.29 & \textbf{90.17 $\pm$ 0.42} & 85.5 $\pm$ 4.04 & 53.96 $\pm$ 5.11 \\ \hline
				\textbf{0.45,0.5} & 88.99 $\pm$ 1.43 & \textbf{89.39 $\pm$ 0.84} & 86.60$\pm$ 2.50 & 52.02$\pm$ 5.17 \\ \hline
				\textbf{0.46,0.48} & 87.31 $\pm$ 2.60 & \textbf{87.92 $\pm$ 2.72} & 79.97 $\pm$ 15.19 & 39.07 $\pm$ 21.75 \\ \hline
				\textbf{0.46,0.5} & 88.94 $\pm$ 0.63 & 88.81 $\pm$ 0.95 & \textbf{89.21 $\pm$ 0.74} & 30.09$\pm$ 24.57 \\ \hline
				\textbf{0.47,0.49} & 87.76 $\pm$ 4.75 & \textbf{89.53 $\pm$ 1.53} & 88.19 $\pm$ 3.189 & 50.21 $\pm$ 0.59 \\ \hline
				\textbf{0.47,0.51} & 89.22 $\pm$ 0.81 & \textbf{89.57 $\pm$ 0.82} & 78.80 $\pm$ 15.21 & 39.15 $\pm$ 19.84 \\ \hline
				\textbf{0.48,0.49} & 86.43 $\pm$ 3.84 & \textbf{89.54 $\pm$ 0.75} & 78.03 $\pm$ 14.61 & 41.65 $\pm$ 21.11 \\ \hline
				\textbf{0.5,0.5} & 86.99$\pm$ 3.69 & \textbf{90.39 $\pm$ 0.93} & 89.24 $\pm$ 0.48 & 47.41 $\pm$ 4.05 \\ \hline
		\end{tabular}}
		\egroup
		\caption{Averaged accuracy and standard deviation values for Fashion MNIST 7-9 (Sneakers and boots) dataset across 5 trials when the classification model M is an MLP. The table demonstrates that WGANXtraY is comparable to WGANXtraY-GP (but with more iterations) most of the time when clean data percent available is only 1\%. This implies that good GANs (columns 3,4 and 5) for generating samples need not be good for generating correctly labelled data points from noisy labelled data points. }
		\label{tab: Fashion_Mnist7-9_GAN_not_work}
	\end{table}

	\begin{table}[]
		\centering
		\bgroup
		\def\arraystretch{0.85}% 
		\scriptsize{ \setlength{\tabcolsep}{0.25em}
			\begin{tabular}{|c|l|c|c|c|}
				\hline
				\textbf{$\rho_+,\rho_-$} & \textbf{WGANXtraY} & \textbf{WGANXtraY-GP} & \textbf{WGANXtraY-R} & \textbf{PacGAN(3)} \\ \hline
				\textbf{0.4,0.49} & 93.22 $\pm$ 3.60 & \textbf{94.91 $\pm$ 2.94} & 90.5 $\pm$ 3.24 & 30.08 $\pm$ 24.56 \\ \hline
				\textbf{0.42,0.45} & 85.01 $\pm$ 17.72 & 92.04 $\pm$ 4.61 & \textbf{92.41 $\pm$ 3.50} & 31.76 $\pm$ 32.26 \\ \hline
				\textbf{0.45,0.46} & 92.53 $\pm$ 3.45 & \textbf{94.98 $\pm$ 1.21} & 90.58 $\pm$ 2.35 & 40.13 $\pm$ 20.06 \\ \hline
				\textbf{0.45,0.48} & 87.49 $\pm$ 14.81 & \textbf{93.57 $\pm$ 3.93} & 93.03 $\pm$ 0.96 & 50.0 $\pm$ 0.0 \\ \hline
				\textbf{0.45,0.5} & \textbf{95.49 $\pm$ 1.08} & 91.56 $\pm$ 4.02 & 91.91$\pm$ 5.63 & 34.37 $\pm$ 29.19 \\ \hline
				\textbf{0.46,0.48} & 95.7 $\pm$ 1.23 & \textbf{96.36 $\pm$ 0.30} & 89.56 $\pm$ 5.75 & 40.0 $\pm$ 20.0 \\ \hline
				\textbf{0.46,0.5} & 81.00$\pm$ 19.00 & \textbf{95.57 $\pm$ 0.81} & 87.55 $\pm$ 7.37 & 49.98 $\pm$ 0.04 \\ \hline
				\textbf{0.47,0.49} & 95.1 $\pm$ 1.47 & \textbf{95.67 $\pm$ 0.79} & 94.49 $\pm$ 1.61 & 50.0 $\pm$ 0.0 \\ \hline
				\textbf{0.47,0.51} & 91.67 $\pm$ 7.29 & \textbf{96.28 $\pm$ 0.64} & 91.52 $\pm$ 7.34 & 42.04 $\pm$ 12.20 \\ \hline
				\textbf{0.48,0.49} & 93.62 $\pm$ 2.65 & \textbf{93.74 $\pm$ 3.17} & 93.08 $\pm$ 3.83 & 40.0 $\pm$ 20.0 \\ \hline
				\textbf{0.5,0.5} & 94.20 $\pm$ 2.46 & 91.64 $\pm$ 9.72 & \textbf{94.22 $\pm$ 2.11} & 40.3 $\pm$ 20.15 \\ \hline
		\end{tabular}}
		\egroup
		\caption{Averaged accuracy and standard deviation values for Fashion MNIST 2-3 (Pullover and Dress) dataset across 5 trials when the classification model M is an MLP. The table demonstrates that WGANXtraY is comparable to WGANXtraY-GP (but with more iterations) most of the time when clean data percent available is only 1\%. This implies that good GANs (columns 3,4 and 5) for generating samples need not be good for generating correctly labelled data points from noisy labelled data points. }
		\label{tab: Fashion_Mnist2-3_GAN_not_work}
	\end{table}
	
	\subsection{Results for low and moderate CCN rates} \label{subsec: ourscheme_low}
	In this section, we demonstrate the performance of WGANXtraY(5) in comparison to GLC at low and moderate noise rates $\rho_+,\rho_-$. We perform the experiments when the gold fraction available is 0.1\% and 1\% for both schemes on 6 binary balanced version of MNIST datasets. For both cases of gold fraction, number of iterations $n_{it} = 1000$. The results are presented in Table \ref{tab: MNIST_low_5-6_7-9}, \ref{tab: MNIST_low_4-9_1-7} and \ref{tab: MNIST_low_0-6_0-1}. For MNIST 0-1 dataset and MNIST 1-7 dataset, our scheme has either comparable or within $5\%$ accuracies of GLC in all cases of noise rates. For other datasets, not satisfactory performance could be due to small KL divergence between $\mathcal{D}$ and $\tilde{\mathcal{D}}$ and inturn low adversarial relation as shown in Lemma \ref{lem: KL_SLN } and \ref{lem: KL_CCN}.
	% \textcolor{red}{write the observations}
	
	\begin{table}[]
		\centering
		\bgroup
		\def\arraystretch{0.85}% 
		\scriptsize{ \setlength{\tabcolsep}{0.25em}
			\begin{tabular}{|c|c|c|c|c|}
				\hline
				\textbf{($\rho_+,\rho_-$)} & \textbf{WGANXtraY} & \textbf{GLC} & \textbf{WGANXtraY} & \textbf{GLC} \\ \hline
				\textbf{Clean \%} & \textbf{0.1 \%} & \textbf{0.1\%} & \textbf{1\%} & \textbf{1\%} \\ \hline
				\textbf{Dataset} & \multicolumn{4}{c|}{\textbf{MNIST 5-6}} \\ \hline
				\textbf{0.01, 0.02} & 88.49$\pm$ 1.93 & \textbf{89.61 $\pm$ 18.91} & 93.87$\pm$1.93 & \textbf{98.99 $\pm$ 0.24} \\ \hline
				\textbf{0.3,0.3} & 82.69 $\pm$ 5.15 & \textbf{97.10 $\pm$ 1.22} & 94.87$\pm$0.35 & \textbf{97.58 $\pm$ 0.30} \\ \hline
				\textbf{0.2,0.49} & 83.68 $\pm$ 7.95 & \textbf{97.29 $\pm$ 0.28} & 95.25$\pm$0.77 & \textbf{96.74 $\pm$ 1.06} \\ \hline
				\textbf{0.42,0.35} & 81.65 $\pm$ 4.22 & \textbf{94.83$\pm$ 0.64} & 95.31$\pm$0.32 & \textbf{95.75 $\pm$ 2.32} \\ \hline
				\textbf{0.1,0.45} & 85.25 $\pm$ 7.33 & \textbf{97.71$\pm$ 1.01} & 93.81$\pm$1.18 & \textbf{98.05$\pm$ 0.29} \\ \hline
				\textbf{Dataset} & \multicolumn{4}{c|}{\textbf{MNIST 7-9}} \\ \hline
				\textbf{0.01, 0.02} & 81.07 $\pm$ 3.12 & \textbf{98.95$\pm$ 0.37} & 90.42$\pm$1.63 & \textbf{99.13 $\pm$ 0.23} \\ \hline
				\textbf{0.3,0.3} & 79.27 $\pm$ 2.14 & \textbf{95.82 $\pm$ 0.78} & 88.25$\pm$2.50 & \textbf{96.63 $\pm$ 1.11} \\ \hline
				\textbf{0.2,0.49} & 79.45 $\pm$ 3.43 & \textbf{95.85 $\pm$ 0.96} & 90.03$\pm$1.38 & \textbf{96.80$\pm$ 0.94} \\ \hline
				\textbf{0.42,0.35} & 82.54 $\pm$ 1.20 & \textbf{95.42 $\pm$ 0.72} & 89.10$\pm$4.99 & \textbf{96.02 $\pm$ 0.84} \\ \hline
				\textbf{0.1,0.45} & 80.13$\pm$ 1.95 & \textbf{97.89 $\pm$ 0.21} & 89.44$\pm$1.44 & \textbf{97.80 $\pm$ 0.25} \\ \hline
		\end{tabular}}
		\egroup
		\caption{Averaged accuracy and standard deviation values for low and moderate noise corrupted MNIST 5-6 and MNIST 7-9 datasets across 5 trials when the classification model M is an MLP.}
		\label{tab: MNIST_low_5-6_7-9}
	\end{table}
	
	\begin{table}[]
		\centering
		\bgroup
		\def\arraystretch{0.85}% 
		\scriptsize{ \setlength{\tabcolsep}{0.25em}
			\begin{tabular}{|c|c|c|c|c|}
				\hline
				\textbf{($\rho_+,\rho_-$)} & \textbf{WGANXtraY} & \textbf{GLC} & \textbf{WGANXtraY} & \textbf{GLC} \\ \hline
				\textbf{Clean \%} & \textbf{0.1 \%} & \textbf{0.1\%} & \textbf{1\%} & \textbf{1\%} \\ \hline
				\textbf{Dataset} & \multicolumn{4}{c|}{\textbf{MNIST 4-9}} \\ \hline
				\textbf{0.01, 0.02} & 73.05 $\pm$ 1.44 & \textbf{98.55 $\pm$ 0.18} & 90.72$\pm$1.59 & \textbf{98.65 $\pm$ 0.28} \\ \hline
				\textbf{0.3,0.3} & 74.33 $\pm$ 0.80 & \textbf{95.59 $\pm$ 1.61} & 90.39$\pm$1.83 & \textbf{96.68 $\pm$ 0.36} \\ \hline
				\textbf{0.2,0.49} & 70.23 $\pm$ 4.75 & \textbf{95.90 $\pm$ 0.46} & 81.17$\pm$15.35 & \textbf{94.87 $\pm$ 1.40} \\ \hline
				\textbf{0.42,0.35} & 77.0 $\pm$ 3.19 & \textbf{93.30 $\pm$ 1.93} & 90.03$\pm$1.05 & \textbf{92.05 $\pm$ 3.31} \\ \hline
				\textbf{0.1,0.45} & 76.45 $\pm$ 3.45 & \textbf{96.63 $\pm$ 0.28} & 91.47$\pm$1.44 & \textbf{97.18 $\pm$ 0.97} \\ \hline
				\textbf{Dataset} & \multicolumn{4}{c|}{\textbf{MNIST 1-7}} \\ \hline
				\textbf{0.01, 0.02} & 95.34 $\pm$ 1.33 & \textbf{99.54$\pm$ 0.08} & 95.62$\pm$1.61 & \textbf{99.60 $\pm$ 0.14} \\ \hline
				\textbf{0.3,0.3} & 95.58 $\pm$ 1.90 & \textbf{98.90 $\pm$ 0.14} & 95.88$\pm$2.44 & \textbf{98.70 $\pm$ 0.60} \\ \hline
				\textbf{0.2,0.49} & 94.49$\pm$ 1.56 & \textbf{98.89$\pm$ 0.26} & 96.64$\pm$0.88 & \textbf{98.86 $\pm$ 0.40} \\ \hline
				\textbf{0.42,0.35} & 96.03 $\pm$ 0.91 & \textbf{97.11 $\pm$ 0.97} & 95.61$\pm$1.13 & \textbf{97.88 $\pm$ 1.29} \\ \hline
				\textbf{0.1,0.45} & 95.32$\pm$ 2.78 & \textbf{98.90 $\pm$ 0.20} & 96.68$\pm$0.93 & \textbf{98.89$\pm$ 0.67} \\ \hline
		\end{tabular}}
		\egroup
		\caption{Averaged accuracy and standard deviation values for low and moderate noise corrupted MNIST 4-9 and MNIST 1-7 datasets across 5 trials when the classification model M is an MLP.}
		\label{tab: MNIST_low_4-9_1-7}
	\end{table}

	\begin{table}[]
		\centering
		\bgroup
		\def\arraystretch{0.85}% 
		\scriptsize{ \setlength{\tabcolsep}{0.25em}
			\begin{tabular}{|c|c|c|c|c|}
				\hline
				\textbf{($\rho_+,\rho_-$)} & \textbf{WGANXtraY} & \textbf{GLC} & \textbf{WGANXtraY} & \textbf{GLC} \\ \hline
				\textbf{Clean \%} & \textbf{0.1 \%} & \textbf{0.1\%} & \textbf{1\%} & \textbf{1\%} \\ \hline
				\textbf{Dataset} & \multicolumn{4}{c|}{\textbf{MNIST 0-6}} \\ \hline
				\textbf{0.01, 0.02} & 88.73$\pm$ 3.77 & \textbf{99.45 $\pm$ 0.14} & 96.17 $\pm$ 0.56 & \textbf{99.46 $\pm$ 0.05} \\ \hline
				\textbf{0.3,0.3} & 88.99$\pm$ 3.37 & \textbf{98.21$\pm$ 0.46} & 86.25 $\pm$ 17.8 & \textbf{98.26 $\pm$ 0.73} \\ \hline
				\textbf{0.2,0.49} & 88.15 $\pm$ 2.43 & \textbf{98.53 $\pm$ 0.29} & 95.34 $\pm$ 0.76 & \textbf{98.15 $\pm$ 0.79} \\ \hline
				\textbf{0.42,0.35} & 91.91 $\pm$ 2.58 & \textbf{95.95 $\pm$ 1.87} & 94.0 $\pm$ 2.66 & \textbf{97.10 $\pm$ 0.84} \\ \hline
				\textbf{0.1,0.45} & 90.68$\pm$ 2.07 & \textbf{98.76 $\pm$ 0.65} & 95.24 $\pm$ 1.04 & \textbf{98.68 $\pm$ 0.28} \\ \hline
				\textbf{Dataset} & \multicolumn{4}{c|}{\textbf{MNIST 0-1}} \\ \hline
				\textbf{0.01, 0.02} & 99.46 $\pm$ 0.28 & \textbf{99.93 $\pm$ 0.03} & 99.38 $\pm$ 0.55 & \textbf{99.88 $\pm$ 0.07} \\ \hline
				\textbf{0.3,0.3} & 97.70 $\pm$ 2.10 & \textbf{99.49$\pm$ 0.67} & 99.13$\pm$ 1.17 & \textbf{99.75$\pm$ 0.16} \\ \hline
				\textbf{0.2,0.49} & \textbf{99.23 $\pm$ 0.78} & 98.47 $\pm$ 2.78 & 99.56 $\pm$ 0.21 & \textbf{99.70 $\pm$ 0.21} \\ \hline
				\textbf{0.42,0.35} & \textbf{99.42$\pm$ 0.34} & 95.20 $\pm$ 8.11 & 99.28 $\pm$ 0.52 & \textbf{99.62 $\pm$ 0.23} \\ \hline
				\textbf{0.1,0.45} & 99.19 $\pm$ 0.55 & \textbf{99.71$\pm$ 0.19} & 98.99 $\pm$ 1.23 & \textbf{99.82 $\pm$ 0.12} \\ \hline
		\end{tabular}}
		\egroup
		\caption{Averaged accuracy and standard deviation values for low and moderate noise corrupted MNIST 0-6 and MNIST 0-1 datasets across 5 trials when the classification model M is an MLP.}
		\label{tab: MNIST_low_0-6_0-1}
	\end{table}

	\section{Details for the plots presented in Section \ref{sec: exper}} \label{sec: details_from_exp_main_paper}
	In this section, we present a detailed comparison of our scheme and GLC on both synthetic and real datasets. We provide the average accuracy and standard deviation values obtained from 5 random trials. All results presented in this section are obtained by using $0.1\%$ clean labelled data except for imbalanced MNIST 4-9 datasets for which clean percentage used is $1\%$. Bold faced values denotes the highest accuracy across the row.
	
	Synthetic data results are reported in Table \ref{tab: Syn_100D} and \ref{tab: Syn_300D}. WGANXtraY(5) outperforms the others methods for both the datasets.
	
	Next, we present the average accuracies along with the standard deviation values for binary balanced class MNIST datasets in Table \ref{tab: MNIST_5-6_high1}, \ref{tab: MNIST_7-9_high1}, \ref{tab: MNIST_4-9_high1}, \ref{tab: MNIST_1-7_high1}, \ref{tab: MNIST_0-6_high1} and  \ref{tab: MNIST_0-1_high1}. In almost all cases for noise rates close to $0.5$, either WGANXtraY(5) or WGANXtraYEntr(5) has higher accuracy value. We also present the results of swapped noise rates in Table \ref{tab: MNIST_5-6_high2}, \ref{tab: MNIST_7-9_high2}, \ref{tab: MNIST_4-9_high2}, \ref{tab: MNIST_1-7_high2}, \ref{tab: MNIST_0-6_high2} and  \ref{tab: MNIST_0-1_high2}. These results also demonstrate better performance of our schemes at high noise rates.

	In Table \ref{tab: Fashion_MNIST_7-9_high1}, \ref{tab: Fashion_MNIST_0-6_high1}, \ref{tab: Fashion_MNIST_2-4_high1} and \ref{tab: Fashion_MNIST_2-3_high1}, we present the average accuracies and the standard deviation values for binary balanced class Fashion MNIST datasets. The results for the same datasets when the noise rates are swapped are provided in Table \ref{tab: Fashion_MNIST_7-9_high2}, \ref{tab: Fashion_MNIST_0-6_high2}, \ref{tab: Fashion_MNIST_2-4_high2} and \ref{tab: Fashion_MNIST_2-3_high2}. One point to note here is that even though GLC has comparable performance in some cases, the standard deviation is very high for almost all cases.
	
	Finally, in Table \ref{tab: MNIST1-7_imb_0.2}, \ref{tab: MNIST1-7_imb_0.7}, \ref{tab: MNIST0-8_imb_0.1}, \ref{tab: MNIST0-8_imb_0.8}, \ref{tab: MNIST4-9_imb_0.3} and \ref{tab: MNIST4-9_imb_0.75}, we present the average AM and standard deviation values for binary imbalanced MNIST datasets for which plots are presented in Figure \ref{fig: MNIST_imb} of Section \ref{sec: exper}. Number of iterations are taken to be $1000$. Except for the case of MNIST 4-9 (imb\_r = 0.75), our schemes with low variations outperform  GLC, GCE and LDMI. For MNIST 4-9 (imb\_r = 0.75) dataset, out of the 3 cases where GLC has higher accuracy, for 2 cases our schemes are within 2 percent of GLC.

	%%%%%%%%%%%%%%%%%%%%%%%%%%%%%%%%%%%%%%%%%%%%%%%%

	\begin{table}[]
		\centering
		\bgroup
		\def\arraystretch{0.85}% 
		\scriptsize{ \setlength{\tabcolsep}{0.25em}
			\begin{tabular}{|c|c|c|c|c|c|c|}
				\hline
				\textbf{$\rho_+,\rho_-$} & \textbf{Simple NN} & \textbf{WGANXtraY} & \textbf{WGANXtraYEntr} & \textbf{GLC}& \textbf{GCE} & \textbf{LDMI} \\ \hline
\textbf{0.4,0.49} & 73.81 $\pm$ 7.47 & \textbf{96.98 $\pm$ 0.35} & 82.69 $\pm$ 17.89 & 64.16 $\pm$ 14.86 & 83.91 $\pm$ 8.31 & 84.42 $\pm$ 12.61 \\ \hline
\textbf{0.42,0.45} & 71.89 $\pm$ 5.61 & \textbf{97.57 $\pm$ 0.70} & 73.65 $\pm$ 20.97 & 46.74$\pm$ 19.86 & 95.65 $\pm$ 3.69 & 89.92 $\pm$ 8.04 \\ \hline
\textbf{0.45,0.46} & 63.75$\pm$ 5.57 & \textbf{97.01 $\pm$ 0.98} & 63.46 $\pm$ 11.37 & 46.16 $\pm$ 12.92 & 87.76 $\pm$ 10.50 & 85.03 $\pm$ 10.41 \\ \hline
\textbf{0.45,0.48} & 64.53 $\pm$ 9.92 & \textbf{95.06 $\pm$ 4.96} & 76.37 $\pm$ 22.41 & 54.85 $\pm$ 8.23 & 77.38 $\pm$ 10.95 & 73.25 $\pm$ 12.71 \\ \hline
\textbf{0.45,0.5} & 60.05 $\pm$ 12.94 & \textbf{96.21 $\pm$ 3.0} & 75.78 $\pm$ 22.16 & 51.12 $\pm$ 13.51 & 71.03 $\pm$ 11.41 & 65.73 $\pm$ 8.06 \\ \hline
\textbf{0.46,0.48} & 64.08 $\pm$ 5.78 & \textbf{96.61 $\pm$ 1.49} & 87.78 $\pm$ 9.64 & 57.2 $\pm$ 9.58 & 83.86 $\pm$ 7.16 & 63.86 $\pm$ 6.44 \\ \hline
\textbf{0.46,0.5} & 51.83 $\pm$ 10.91 & \textbf{96.13 $\pm$ 0.83} & 90.16 $\pm$ 12.12 & 52.88 $\pm$ 9.27 & 68.16 $\pm$ 10.46 & 61.04 $\pm$ 19.39 \\ \hline
\textbf{0.47,0.49} & 60.0 $\pm$ 12.68 & \textbf{94.64 $\pm$ 5.28} & 75.67 $\pm$ 18.66 & 53.92 $\pm$ 5.84 & 80.0 $\pm$ 9.14 & 62.74 $\pm$ 9.72 \\ \hline
\textbf{0.47,0.51} & 53.62 $\pm$ 3.39 & \textbf{97.33 $\pm$ 0.71} & 90.74 $\pm$ 9.58 & 50.02 $\pm$ 5.80 & 58.96 $\pm$ 4.99 & 50.93$\pm$ 3.07 \\ \hline
\textbf{0.48,0.49} & 62.69 $\pm$ 7.48 & \textbf{97.54 $\pm$ 1.32} & 84.16 $\pm$ 15.88 & 50.05 $\pm$ 3.35 & 76.08 $\pm$ 10.80 & 61.27 $\pm$ 11.10 \\ \hline
\textbf{0.5,0.5} & 50.72 $\pm$ 10.71 & \textbf{95.65 $\pm$ 1.87} & 82.50$\pm$ 18.00 & 51.52 $\pm$ 4.47 & 53.68 $\pm$ 7.09 & 48.74 $\pm$ 3.35 \\ \hline
		\end{tabular}}
		\egroup
		\caption{Averaged accuracy and standard deviation values for Synthetic dataset SD100 across 5 trials when the classification model M is an MLP. The table  demonstrates that for very high noise rates, i.e., close to 0.5, WGANXtraY shows a significant improvement (third column) over GLC, GCE and LDMI even when clean data percent available is only 0.1\%. }
		\label{tab: Syn_100D}
	\end{table}

	\begin{table}[]
		\centering
		\bgroup
		\def\arraystretch{0.85}% 
		\scriptsize{ \setlength{\tabcolsep}{0.25em}
			\begin{tabular}{|c|c|c|c|c|c|c|}
				\hline
				\textbf{$\rho_+,\rho_-$} & \textbf{Simple NN} & \textbf{WGANXtraY} & \textbf{WGANXtraYEntr} & \textbf{GLC}& \textbf{GCE} & \textbf{LDMI} \\ \hline
\textbf{0.4,0.49} & 73.11 $\pm$ 6.36 & \textbf{99.92 $\pm$ 0.10} & 84.4 $\pm$ 18.12 & 59.49 $\pm$ 11.56 & 85.17 $\pm$ 3.07 & 83.89 $\pm$ 13.20 \\ \hline
\textbf{0.42,0.45} & 72.05 $\pm$ 8.63 & \textbf{99.81 $\pm$ 0.19} & 78.66 $\pm$ 15.97 & 58.8 $\pm$ 10.52 & 96.10 $\pm$ 4.62 & 79.49 $\pm$ 11.73 \\ \hline
\textbf{0.45,0.46} & 60.4 $\pm$ 9.47 & \textbf{99.94 $\pm$ 0.10} & 64.26 $\pm$ 14.30 & 58.56 $\pm$ 10.22 & 87.86 $\pm$ 9.40 & 75.92 $\pm$ 17.99 \\ \hline
\textbf{0.45,0.48} & 66.69 $\pm$ 11.73 & \textbf{99.89 $\pm$ 0.05} & 85.81 $\pm$ 18.56 & 54.58 $\pm$ 9.94 & 90.13 $\pm$ 3.80 & 74.26 $\pm$ 18.60 \\ \hline
\textbf{0.45,0.5} & 56.26 $\pm$ 12.38 & \textbf{99.30 $\pm$ 0.98} & 84.45 $\pm$ 17.85 & 52.56 $\pm$ 2.32 & 70.58 $\pm$ 9.55 & 61.54 $\pm$ 9.79 \\ \hline
\textbf{0.46,0.48} & 68.58 $\pm$ 7.22 & \textbf{99.60 $\pm$ 0.50} & 92.85 $\pm$ 7.33 & 53.89 $\pm$ 9.28 & 84.98 $\pm$ 3.05 & 57.25 $\pm$ 5.40 \\ \hline
\textbf{0.46,0.5} & 55.62 $\pm$ 8.64 & \textbf{99.92 $\pm$ 0.10} & 78.10 $\pm$ 20.71 & 47.46 $\pm$ 3.55 & 78.98 $\pm$ 9.27 & 54.0 $\pm$ 4.14 \\ \hline
\textbf{0.47,0.49} & 55.09 $\pm$ 9.30 & \textbf{100.0 $\pm$ 0.0} & 86.53 $\pm$ 9.85 & 50.77 $\pm$ 3.34 & 75.70 $\pm$ 9.87 & 57.2 $\pm$ 5.39 \\ \hline
\textbf{0.47,0.51} & 53.01 $\pm$ 8.63 & \textbf{99.94 $\pm$ 0.10} & 68.29 $\pm$ 20.32 & 49.89 $\pm$ 11.21 & 60.32 $\pm$ 9.73 & 56.15 $\pm$ 3.48 \\ \hline
\textbf{0.48,0.49} & 63.70 $\pm$ 10.13 & \textbf{99.89 $\pm$ 0.15} & 85.46 $\pm$ 18.89 & 50.8 $\pm$ 4.46 & 73.35 $\pm$ 10.31 & 55.86 $\pm$ 4.75 \\ \hline
\textbf{0.5,0.5} & 50.45 $\pm$ 12.62 & \textbf{99.57 $\pm$ 0.72} & 94.05 $\pm$ 5.78 & 58.98 $\pm$ 8.71 & 56.4 $\pm$ 13.94 & 47.28 $\pm$ 4.46 \\ \hline
		\end{tabular}}
		\egroup
		\caption{Averaged accuracy and standard deviation values for Synthetic dataset SD300 across 5 trials when the classification model M is an MLP. The table  demonstrates that for very high noise rates, i.e., close to 0.5, WGANXtraY shows a significant improvement (third column) over GLC, GCE and LDMI even when clean data percent available is only 0.1\%.}
		\label{tab: Syn_300D}
	\end{table}

	%%%%%%%%%%%%%%%%%%%%%%%%%%%%%%%%%%%%%%%%%%%%%%%%

	\begin{table}[]
		\centering
		\bgroup
		\def\arraystretch{0.85}% 
		\scriptsize{ \setlength{\tabcolsep}{0.25em}
			\begin{tabular}{|c|c|c|c|c|c|c|}
\hline
\textbf{$\rho_+,\rho_-$} & \textbf{Simple NN} & \textbf{WGANXtraY} & \textbf{WGANXtraYEntr} & \textbf{GLC} & \textbf{GCE} & \textbf{LDMI} \\ \hline
\textbf{0.4,0.49} & 49.28$\pm$ 2.14 & 78.8 $\pm$ 4.8 & \textbf{88.71 $\pm$ 3.23} & 84.35 $\pm$ 10.29 & 48.21 $\pm$ 0.0 & 83.47 $\pm$ 9.92 \\ \hline
\textbf{0.42,0.45} & 65.81 $\pm$ 16.3 & 86.88$\pm$ 5.76 & 87.21 $\pm$ 3.29 & \textbf{88.18 $\pm$ 4.94} & 48.56 $\pm$ 0.69 & 43.67 $\pm$ 34.29 \\ \hline
\textbf{0.45,0.46} & 85.98 $\pm$ 6.75 & 83.08 $\pm$ 3.69 & 80.46 $\pm$ 10.73 & 86.52 $\pm$ 14.70 & 67.72 $\pm$ 16.04 & \textbf{92.63 $\pm$ 1.81} \\ \hline
\textbf{0.45,0.48} & 74.30 $\pm$ 8.89 & 84.5 $\pm$ 9.63 & \textbf{89.47 $\pm$ 3.46} & 75.85 $\pm$ 16.8 & 51.31 $\pm$ 6.20 & 76.61 $\pm$ 10.46 \\ \hline
\textbf{0.45,0.5} & 48.22 $\pm$ 0.02 & \textbf{91.2 $\pm$ 1.81} & 79.62 $\pm$ 6.70 & 64.6 $\pm$ 24.7 & 48.21 $\pm$ 0.0 & 69.70 $\pm$ 14.31 \\ \hline
\textbf{0.46,0.48} & 63.36 $\pm$ 18.2 & \textbf{88.3 $\pm$ 6.06} & 79.81 $\pm$ 4.46 & 72.30 $\pm$ 14.99 & 56.46 $\pm$ 12.60 & 69.25 $\pm$ 13.69 \\ \hline
\textbf{0.46,0.5} & 53.51 $\pm$ 10.6 & 83.8 $\pm$ 7.10 & \textbf{84.98 $\pm$ 7.93} & 79.0 $\pm$ 12.48 & 48.21 $\pm$ 0.0 & 75.09 $\pm$ 10.60 \\ \hline
\textbf{0.47,0.49} & 48.3 $\pm$ 0.22 & \textbf{86.1 $\pm$ 6.24} & 86.09 $\pm$ 2.02 & 71.01 $\pm$ 14.67 & 48.21 $\pm$ 0.0 & 64.84 $\pm$ 8.56 \\ \hline
\textbf{0.47,0.51} & 48.21 $\pm$ 0.0 & 86.1 $\pm$ 2.53 & \textbf{87.97 $\pm$ 3.56} & 72.45 $\pm$ 11.54 & 48.21 $\pm$ 0.0 & 57.72 $\pm$ 9.38 \\ \hline
\textbf{0.48,0.49} & 53.69 $\pm$ 4.66 & 82.7 $\pm$ 8.47 & \textbf{83.28 $\pm$ 4.58} & 62.51 $\pm$ 9.93 & 50.05 $\pm$ 2.53 & 64.95 $\pm$ 10.07 \\ \hline
\textbf{0.5,0.5} & 50.19 $\pm$ 2.69 & 79.77 $\pm$ 2.73 & \textbf{80.50 $\pm$ 5.78} & 62.37$\pm$ 7.30 & 51.37 $\pm$ 1.68 & 58.62 $\pm$ 13.50 \\ \hline
\end{tabular}}
		\egroup
		\caption{Averaged accuracy and standard deviation values for MNIST 5-6 dataset across 5 trials when the classification model M is an MLP. The table demonstrates that for very high noise rates, i.e., close to 0.5, WGAN based schemes have higher accuracies and low variation among the trials than GLC, GCE and LDMI.}
		\label{tab: MNIST_5-6_high1}
	\end{table}

	\begin{table}[]
		\centering
		\bgroup
		\def\arraystretch{0.85}% 
		\scriptsize{ \setlength{\tabcolsep}{0.25em}
			\begin{tabular}{|c|c|c|c|c|c|c|}
\hline
\textbf{$\rho_+,\rho_-$} & \textbf{Simple NN} & \textbf{WGANXtraY} & \textbf{WGANXtraYEntr} & \textbf{GLC} & \textbf{GCE} & \textbf{LDMI} \\ \hline
\textbf{0.4,0.49} & 50.46$\pm$ 0.0 & 78.8$\pm$ 1.33 & 78.76 $\pm$ 3.18 & 86.86 $\pm$ 7.70 & 50.46 $\pm$ 0.0 & \textbf{86.42 $\pm$ 5.62} \\ \hline
\textbf{0.42,0.45} & 67.99 $\pm$ 11.96 & \textbf{75.9$\pm$ 3.87} & 82.43 $\pm$ 1.29 & 86.04 $\pm$ 7.14 & 50.46 $\pm$ 0.0 & \textbf{92.75 $\pm$ 0.57} \\ \hline
\textbf{0.45,0.46} & 75.79 $\pm$ 11.96 & 81.49$\pm$ 1.2 & 80.10 $\pm$ 2.85 & 82.8 $\pm$ 7.5 & 51.98 $\pm$ 2.19 & \textbf{85.73 $\pm$ 3.42} \\ \hline
\textbf{0.45,0.48} & 50.6 $\pm$ 0.451 & 81.1$\pm$ 2.56 & 81.99 $\pm$ 1.58 & \textbf{84.2 $\pm$ 2.78} & 50.46 $\pm$ 0.0 & 78.03 $\pm$ 7.19 \\ \hline
\textbf{0.45,0.5} & 50.46 $\pm$ 0.0 & 78.0$\pm$ 4.2 & \textbf{80.81 $\pm$ 2.36} & 67.5 $\pm$ 17.68 & 50.46 $\pm$ 0.0 & 68.19 $\pm$ 7.0 \\ \hline
\textbf{0.46,0.48} & 58.91 $\pm$ 11.82 & \textbf{80.79 $\pm$ 2.37} & 80.32 $\pm$ 2.44 & 72.4 $\pm$ 8.9 & 50.46 $\pm$ 0.0 & 76.76 $\pm$ 15.33 \\ \hline
\textbf{0.46,0.5} & 50.46 $\pm$ 0.0 & \textbf{80.3 $\pm$ 1.88} & 80.01 $\pm$ 1.09 & 76.94 $\pm$ 2.79 & 50.46 $\pm$ 0.0 & 63.49 $\pm$ 10.09 \\ \hline
\textbf{0.47,0.49} & 56.85$\pm$ 11.02 & 76.9$\pm$ 4.54 & \textbf{77.83 $\pm$ 1.50} & 78.88$\pm$ 4.66 & 50.47 $\pm$ 0.01 & 71.08 $\pm$ 13.62 \\ \hline
\textbf{0.47,0.51} & 50.496 $\pm$ 0.058 & 78.9$\pm$ 1.70 & \textbf{78.96 $\pm$ 3.65} & 70.10 $\pm$ 7.63 & 50.46 $\pm$ 0.0 & 59.31$\pm$ 11.4 \\ \hline
\textbf{0.48,0.49} & 62.4 $\pm$ 7.42 & 80.7 $\pm$ 2.07 & \textbf{81.81$\pm$ 1.19} & 56.77 $\pm$ 14.93 & 50.44 $\pm$ 0.03 & 59.70 $\pm$ 9.3 \\ \hline
\textbf{0.5,0.5} & 41.9 $\pm$ 6.39 & \textbf{79.2$\pm$ 1.51} & 79.09 $\pm$ 4.17 & 61.5 $\pm$ 7.90 & 49.88 $\pm$ 0.78 & 57.00$\pm$ 9.56 \\ \hline
\end{tabular}}
		\egroup
		\caption{Averaged accuracy and standard deviation values for MNIST 7-9 dataset across 5 trials when the classification model M is an MLP. The table demonstrates that for very high noise rates, i.e., close to 0.5, WGAN based schemes have high or comparable accuracies and low variation among the trials than GLC, GCE and LDMI.}
		\label{tab: MNIST_7-9_high1}
	\end{table}
	
	\begin{table}[]
		\centering
		\bgroup
		\def\arraystretch{0.85}% 
		\scriptsize{ \setlength{\tabcolsep}{0.25em}
			\begin{tabular}{|c|c|c|c|c|c|c|}
\hline
\textbf{$\rho_+,\rho_-$} & \textbf{Simple NN} & \textbf{WGANXtraY} & \textbf{WGANXtraYEntr} & \textbf{GLC} & \textbf{GCE} & \textbf{LDMI} \\ \hline
\textbf{0.4,0.49} & 49.32$\pm$ 0.0 & 71.47 $\pm$ 4.22 & 75.07 $\pm$ 4.43 & 83.92$\pm$ 4.60 & 49.32 $\pm$ 0.0 & \textbf{86.82$\pm$ 2.08} \\ \hline
\textbf{0.42,0.45} & 54.32 $\pm$ 8.98 & 72.2 $\pm$ 4.81 & 74.56 $\pm$ 1.74 & \textbf{88.10 $\pm$ 4.16} & 52.25 $\pm$ 5.86 & 81.98$\pm$ 6.48 \\ \hline
\textbf{0.45,0.46} & 53.67 $\pm$ 5.06 & 72.22 $\pm$ 3.4 & 76.30 $\pm$ 1.59 & 79.4 $\pm$ 11.41 & 49.55 $\pm$ 0.46 & \textbf{84.88 $\pm$ 2.73} \\ \hline
\textbf{0.45,0.48} & 49.35 $\pm$ 0.06 & 73.93 $\pm$ 4.26 & 77.62 $\pm$ 3.47 & 69.5 $\pm$ 10.31 & 49.32 $\pm$ 0.0 & \textbf{78.52$\pm$ 7.18} \\ \hline
\textbf{0.45,0.5} & 49.35 $\pm$ 0.06 & 74.15 $\pm$ 2.27 & \textbf{74.89 $\pm$ 1.55} & 60.0 $\pm$ 11.66 & 49.32 $\pm$ 0.0 & 65.18$\pm$ 4.24 \\ \hline
\textbf{0.46,0.48} & 55.1 $\pm$ 10.73 & \textbf{75.63 $\pm$ 1.42} & 72.74 $\pm$ 3.07 & 66.8 $\pm$ 16.03 & 50.72 $\pm$ 2.81 & 72.28$\pm$ 13.43 \\ \hline
\textbf{0.46,0.5} & 49.32$\pm$ 0.0 & 73.42 $\pm$ 1.89 & \textbf{76.60 $\pm$ 2.77} & 65.89 $\pm$ 11.42 & 49.32 $\pm$ 0.0 & 46.55$\pm$ 11.99 \\ \hline
\textbf{0.47,0.49} & 60.73 $\pm$ 13.2 & 73.52 $\pm$ 3.92 & \textbf{74.91 $\pm$ 3.20} & 60.1 $\pm$ 6.88 & 50.81 $\pm$ 2.99 & 60.43$\pm$ 15.92 \\ \hline
\textbf{0.47,0.51} & 49.7$\pm$ 0.94 & 75.8 $\pm$ 2.75 & \textbf{75.90 $\pm$ 2.32} & 73.0 $\pm$ 6.77 & 49.32 $\pm$ 0.0 & 56.07 $\pm$ 6.84 \\ \hline
\textbf{0.48,0.49} & 52.4$\pm$ 2.32 & 73.69 $\pm$ 2.69 & \textbf{78.32$\pm$ 2.08} & 57.06 $\pm$ 7.17 & 48.39 $\pm$ 1.84 & 60.34$\pm$ 3.87 \\ \hline
\textbf{0.5,0.5} & 52.4$\pm$ 4.26 & 70.46 $\pm$ 2.80 & \textbf{76.82$\pm$ 2.90} & 55.82 $\pm$ 7.49 & 49.37 $\pm$ 1.39 & 46.74$\pm$ 8.34 \\ \hline
\end{tabular}}
		\egroup
		\caption{Averaged accuracy and standard deviation values for MNIST 4-9 dataset across 5 trials when the classification model M is an MLP. The table demonstrates that for very high noise rates, i.e., close to 0.5, WGAN based schemes have higher accuracies and low variation among the trials than GLC, GCE and LDMI in most cases.}
		\label{tab: MNIST_4-9_high1}
	\end{table}

	\begin{table}[]
		\centering
		\bgroup
		\def\arraystretch{0.85}% 
		\scriptsize{ \setlength{\tabcolsep}{0.25em}
			\begin{tabular}{|c|c|c|c|c|c|c|}
\hline
\textbf{$\rho_+,\rho_-$} & \textbf{Simple NN} & \textbf{WGANXtraY} & \textbf{WGANXtraYEntr} & \textbf{GLC} & \textbf{GCE} & \textbf{LDMI} \\ \hline
\textbf{0.4,0.49} & 53.08 $\pm$ 1.22 & 94.16 $\pm$ 2.0 & \textbf{95.72 $\pm$ 0.50} & 94.5 $\pm$ 2.22 & 52.47 $\pm$ 0.0 & 87.78 $\pm$ 5.18 \\ \hline
\textbf{0.42,0.45} & 85.04 $\pm$ 7.74 & 94.44 $\pm$ 1.94 & \textbf{95.36 $\pm$ 1.39} & 93.06 $\pm$ 5.66 & 53.25 $\pm$ 1.57 & 94.70 $\pm$ 3.14 \\ \hline
\textbf{0.45,0.46} & 87.54 $\pm$ 11.78 & \textbf{96.0 $\pm$ 1.20} & 94.66 $\pm$ 1.84 & 95.44 $\pm$ 2.6 & 64.95 $\pm$ 15.04 & 90.22 $\pm$ 8.04 \\ \hline
\textbf{0.45,0.48} & 74.49 $\pm$ 17.09 & 92.78 $\pm$ 3.91 & \textbf{95.92 $\pm$ 1.40} & 90.70 $\pm$ 9.84 & 52.47 $\pm$ 0.0 & 81.01 $\pm$ 12.38 \\ \hline
\textbf{0.45,0.5} & 52.9 $\pm$ 0.64 & 92.7 $\pm$ 3.9 & \textbf{94.08 $\pm$ 2.24} & 80.87 $\pm$ 14.9 & 52.47 $\pm$ 0.0 & 79.01 $\pm$ 15.39 \\ \hline
\textbf{0.46,0.48} & 60.36 $\pm$ 15.7 & \textbf{94.9 $\pm$ 0.798} & 89.52 $\pm$ 11.86 & 77.17 $\pm$ 28.45 & 52.47 $\pm$ 0.0 & 84.58 $\pm$ 10.62 \\ \hline
\textbf{0.46,0.5} & 52.475 $\pm$ 0.0 & \textbf{95.3 $\pm$ 2.41} & 94.87 $\pm$ 1.34 & 70.3 $\pm$ 12.04 & 52.47 $\pm$ 0.0 & 81.11 $\pm$ 10.67 \\ \hline
\textbf{0.47,0.49} & 65.89 $\pm$ 15.2 & 94.7 $\pm$ 1.15 & \textbf{94.80 $\pm$ 2.07} & 73.38 $\pm$ 19.57 & 50.49 $\pm$ 2.42 & 82.35 $\pm$ 16.16 \\ \hline
\textbf{0.47,0.51} & 52.47 $\pm$ 0.0 & 94.4 $\pm$ 1.77 & \textbf{94.83 $\pm$ 2.20} & 79.32 $\pm$ 13.71 & 52.47 $\pm$ 0.0 & 77.48 $\pm$ 20.32 \\ \hline
\textbf{0.48,0.49} & 56.9 $\pm$ 10.18 & 94.7 $\pm$ 1.19 & \textbf{95.49 $\pm$ 1.57} & 83.94 $\pm$ 12.04 & 57.13 $\pm$ 9.32 & 65.05 $\pm$ 13.86 \\ \hline
\textbf{0.5,0.5} & 44.45$\pm$ 11.4 & 95.0$\pm$ 0.93 & \textbf{95.22 $\pm$ 2.35} & 67.94 $\pm$ 6.33 & 51.78 $\pm$ 4.20 & 55.34 $\pm$ 15.52 \\ \hline
\end{tabular}}
		\egroup
		\caption{Averaged accuracy and standard deviation values for MNIST 1-7 dataset across 5 trials when the classification model M is an MLP. The table demonstrates that for very high noise rates, i.e., close to 0.5, WGAN based schemes have higher accuracies and low variation among the trials than GLC, GCE and LDMI.}
		\label{tab: MNIST_1-7_high1}
	\end{table}
	
	\begin{table}[]
		\centering
		\bgroup
		\def\arraystretch{0.85}% 
		\scriptsize{ \setlength{\tabcolsep}{0.25em}
			\begin{tabular}{|c|c|c|c|c|c|c|}
\hline
\textbf{$\rho_+,\rho_-$} & \textbf{Simple NN} & \textbf{WGANXtraY} & \textbf{WGANXtraYEntr} & \textbf{GLC} & \textbf{GCE} & \textbf{LDMI} \\ \hline
\textbf{0.4,0.49} & 50.85 $\pm$ 1.53 & 90.78 $\pm$ 1.79 & 91.71 $\pm$ 2.64 & 93.44 $\pm$ 3.09 & 49.43 $\pm$ 0.0 & \textbf{94.47$\pm$ 3.11} \\ \hline
\textbf{0.42,0.45} & 80.69 $\pm$ 15.58 & 91.05 $\pm$ 1.83 & 91.49 $\pm$ 2.91 & \textbf{94.95 $\pm$ 1.73} & 55.32 $\pm$ 11.78 & 91.43 $\pm$ 5.74 \\ \hline
\textbf{0.45,0.46} & 83.43$\pm$ 8.76 & 88.85 $\pm$ 2.47 & 89.95 $\pm$ 2.45 & 91.5 $\pm$ 3.58 & 73.12 $\pm$ 13.73 & \textbf{93.25 $\pm$ 3.00} \\ \hline
\textbf{0.45,0.48} & 61.63 $\pm$ 10.55 & 87.89 $\pm$ 3.08 & 89.98 $\pm$ 2.90 & 85.46$\pm$ 18.07 & 55.89 $\pm$ 12.21 & \textbf{90.48 $\pm$ 6.91} \\ \hline
\textbf{0.45,0.5} & 49.46 $\pm$ 0.06 & \textbf{90.85 $\pm$ 1.57} & 87.17 $\pm$ 5.09 & 77.80 $\pm$ 17.29 & 49.43 $\pm$ 0.0 & 62.09 $\pm$ 14.84 \\ \hline
\textbf{0.46,0.48} & 60.44 $\pm$ 12.34 & 85.30 $\pm$ 4.19 & \textbf{92.49 $\pm$ 1.54} & 80.01 $\pm$ 15.33 & 49.66 $\pm$ 0.47 & 83.90 $\pm$ 9.54 \\ \hline
\textbf{0.46,0.5} & 50.90 $\pm$ 2.49 & 87.9 $\pm$ 2.89 & \textbf{92.04 $\pm$ 0.77} & 71.45 $\pm$ 9.14 & 49.43 $\pm$ 0.0 & 68.29$\pm$ 16.68 \\ \hline
\textbf{0.47,0.49} & 66.25$\pm$ 15.07 & \textbf{87.83 $\pm$ 2.12} & 85.26 $\pm$ 7.76 & 72.6 $\pm$ 18.53 & 49.43 $\pm$ 0.0 & 82.96 $\pm$ 14.51 \\ \hline
\textbf{0.47,0.51} & 49.43 $\pm$ 0.0 & 87.97 $\pm$ 1.08 & \textbf{90.80 $\pm$ 2.19} & 80.06 $\pm$ 10.2 & 49.43 $\pm$ 0.0 & 66.02$\pm$ 18.37 \\ \hline
\textbf{0.48,0.49} & 69.49 $\pm$ 12.48 & \textbf{91.49 $\pm$ 1.32} & 89.52 $\pm$ 3.34 & 76.06 $\pm$ 17.8 & 56.95 $\pm$ 6.02 & 59.87 $\pm$ 16.22 \\ \hline
\textbf{0.5,0.5} & 48.29 $\pm$ 1.98 & 86.4 $\pm$ 1.23 & \textbf{88.31 $\pm$ 3.51} & 81.1 $\pm$ 7.62 & 49.18 $\pm$ 1.70 & 50.88$\pm$ 4.37 \\ \hline
\end{tabular}}
		\egroup
		\caption{Averaged accuracy and standard deviation values for MNIST 0-6 dataset across 5 trials when the classification model M is an MLP. The table demonstrates that for very high noise rates, i.e., close to 0.5, WGAN based schemes have higher accuracies and low variation among the trials than GLC, GCE and LDMI.}
		\label{tab: MNIST_0-6_high1}
	\end{table}

	\begin{table}[]
		\centering
		\bgroup
		\def\arraystretch{0.85}% 
		\scriptsize{ \setlength{\tabcolsep}{0.25em}
			\begin{tabular}{|c|c|c|c|c|c|c|}
\hline
\textbf{$\rho_+,\rho_-$} & \textbf{Simple NN} & \textbf{WGANXtraY} & \textbf{WGANXtraYEntr} & \textbf{GLC} & \textbf{GCE} & \textbf{LDMI} \\ \hline
\textbf{0.4,0.49} & 60.84 $\pm$ 9.05 & \textbf{99.38 $\pm$ 0.39} & 98.52 $\pm$ 1.82 & 98.70 $\pm$ 1.34 & 53.66 $\pm$ 0.0 & 89.98$\pm$ 11.62 \\ \hline
\textbf{0.42,0.45} & 95.4 $\pm$ 2.87 & 98.83 $\pm$ 0.85 & 97.80 $\pm$ 3.16 & 90.4 $\pm$ 16.93 & 58.43 $\pm$ 5.99 & \textbf{99.60$\pm$ 0.55} \\ \hline
\textbf{0.45,0.46} & 96.46 $\pm$ 1.60 & \textbf{99.24 $\pm$ 0.61} & 98.91 $\pm$ 0.74 & 94.4 $\pm$ 8.48 & 67.02 $\pm$ 15.52 & 92.42 $\pm$ 8.94 \\ \hline
\textbf{0.45,0.48} & 66.50 $\pm$ 14.5 & 98.8 $\pm$ 0.81 & \textbf{99.28 $\pm$ 0.29} & 91.13 $\pm$ 9.6 & 53.66 $\pm$ 0.0 & 92.42$\pm$ 7.58 \\ \hline
\textbf{0.45,0.5} & 54.45 $\pm$ 1.09 & \textbf{98.59 $\pm$ 1.14} & 98.22 $\pm$ 2.45 & 85.06 $\pm$ 15.83 & 53.63 $\pm$ 0.0 & 81.56$\pm$ 18.26 \\ \hline
\textbf{0.46,0.48} & 66.14 $\pm$ 13.01 & 98.94 $\pm$ 0.27 & \textbf{99.43 $\pm$ 0.32} & 95.79 $\pm$ 3.89 & 55.40 $\pm$ 3.0 & 96.28$\pm$ 6.35 \\ \hline
\textbf{0.46,0.5} & 59.9 $\pm$ 8.68 & \textbf{98.42 $\pm$ 1.7} & 97.95 $\pm$ 1.44 & 96.3$\pm$ 4.49 & 53.66 $\pm$ 0.0 & 94.51$\pm$ 8.98 \\ \hline
\textbf{0.47,0.49} & 82.2 $\pm$ 12.77 & \textbf{98.73 $\pm$ 1.35} & 99.58$\pm$ 0.15 & 93.76$\pm$ 5.43 & 53.66 $\pm$ 0.0 & 89.50$\pm$ 9.02 \\ \hline
\textbf{0.47,0.51} & 53.66 $\pm$ 7.10 & \textbf{98.43 $\pm$ 1.41} & 98.08$\pm$ 2.29 & 78.2 $\pm$ 15.93 & 53.66 $\pm$ 0.0 & 75.05$\pm$ 15.96 \\ \hline
\textbf{0.48,0.49} & 76.7 $\pm$ 20.01 & \textbf{99.29 $\pm$ 0.43} & 98.15 $\pm$ 2.07 & 89.82 $\pm$ 6.26 & 58.00 $\pm$ 10.56 & 95.00 $\pm$ 4.01 \\ \hline
\textbf{0.5,0.5} & 41.21$\pm$ 11.5 & 98.63 $\pm$ 1.26 & \textbf{99.61 $\pm$ 0.20} & 79.68 $\pm$ 13.28 & 51.14 $\pm$ 6.62 & 69.64 $\pm$ 24.33 \\ \hline
\end{tabular}}
		\egroup
		\caption{Averaged accuracy and standard deviation values for MNIST 0-1 dataset across 5 trials when the classification model M is an MLP. The table demonstrates that for very high noise rates, i.e., close to 0.5, WGAN based schemes have higher accuracies and low variation among the trials than GLC, GCE and LDMI in most cases.}
		\label{tab: MNIST_0-1_high1}
	\end{table}
	
	%%%%%%%%%%%%%%%%%%%%%%%%%%%%%%%%%%%%%%%%%%%%%%%%
	
	\begin{table}[]
		\centering
		\bgroup
		\def\arraystretch{0.85}% 
		\scriptsize{ \setlength{\tabcolsep}{0.25em}
			\begin{tabular}{|c|l|c|c|c|}
				\hline
				\textbf{$\rho_+,\rho_-$} & \textbf{SimpleNN} & \textbf{WGANXtraY} & \textbf{WGANXtraYEntr} & \textbf{GLC} \\ \hline
				\textbf{0.49,0.4} & 51.78 $\pm$ 0.0 & 83.27 $\pm$ 4.70 & 88.51 $\pm$ 2.35 & \textbf{92.30 $\pm$ 3.78} \\ \hline
				\textbf{0.45,0.42} & 63.1$\pm$ 10.34 & 81.41$\pm$ 6.27 & 82.12 $\pm$ 3.46 & \textbf{91.22 $\pm$ 4.87} \\ \hline
				\textbf{0.46,0.45} & 71.2 $\pm$ 12.37 & 82.88 $\pm$ 6.83 & \textbf{85.44$\pm$ 6.57} & 80.86 $\pm$ 13.14 \\ \hline
				\textbf{0.48,0.45} & 57.3 $\pm$ 10.13 & 86.1$\pm$ 3.21 & \textbf{87.84$\pm$ 3.07} & 68.18$\pm$ 25.37 \\ \hline
				\textbf{0.5,0.45} & 51.783 $\pm$ 0.0 & 81.0 $\pm$ 8.43 & \textbf{83.23 $\pm$ 6.71} & 64.42 $\pm$ 16.42 \\ \hline
				\textbf{0.48,0.46} & 56.5 $\pm$ 4.28 & \textbf{84.29 $\pm$ 5.75} & 83.94 $\pm$ 5.96 & 83.3 $\pm$ 11.93 \\ \hline
				\textbf{0.5,0.46} & 51.78$\pm$ 0.0 & 83.97 $\pm$ 5.68 & \textbf{87.89 $\pm$ 3.24} & 71.29 $\pm$ 9.05 \\ \hline
				\textbf{0.49,0.47} & 51.83$\pm$ 0.08 & \textbf{85.16 $\pm$ 4.85} & 82.05 $\pm$ 9.76 & 66.85 $\pm$ 15.13 \\ \hline
				\textbf{0.51,0.47} & 52.2 $\pm$ 1.01 & 85.2 $\pm$ 5.52 & \textbf{89.50$\pm$ 2.38} & 59.7 $\pm$ 6.93 \\ \hline
				\textbf{0.49,0.48} & 57.93$\pm$ 7.64 & 86.69$\pm$ 2.22 & \textbf{87.63 $\pm$ 3.02} & 68.3 $\pm$ 13.33 \\ \hline
		\end{tabular}}
		\egroup
		\caption{Averaged accuracy and standard deviation values for MNIST 5-6 dataset across 5 trials when the classification model M is an MLP. The noise rates have been swapped. The table demonstrates that for very high noise rates, i.e., close to 0.5, WGAN based schemes have higher accuracies and low variation among the trials than GLC.}
		\label{tab: MNIST_5-6_high2}
	\end{table}
	
	\begin{table}[]
		\centering
		\bgroup
		\def\arraystretch{0.85}% 
		\scriptsize{ \setlength{\tabcolsep}{0.25em}
			\begin{tabular}{|c|l|c|c|c|}
				\hline
				\textbf{$\rho_+,\rho_-$} & \textbf{SimpleNN} & \textbf{WGANXtraY} & \textbf{WGANXtraYEntr} & \textbf{GLC} \\ \hline
				\textbf{0.49,0.4} & 49.5 $\pm$ 0.0 & 76.5$\pm$ 2.28 & 78.48 $\pm$ 3.61 & \textbf{84.61 $\pm$ 6.27} \\ \hline
				\textbf{0.45,0.42} & 67.9 $\pm$ 12.86 & 80.58 $\pm$ 2.42 & 78.81 $\pm$ 3.44 & \textbf{92.55 $\pm$ 1.99} \\ \hline
				\textbf{0.46,0.45} & 73.08$\pm$ 9.97 & 82.16 $\pm$ 1.43 & 80.11 $\pm$ 4.14 & \textbf{89.24 $\pm$ 5.11} \\ \hline
				\textbf{0.48,0.45} & 54.8 $\pm$ 10.72 & \textbf{79.86$\pm$ 2.52} & 79.43 $\pm$ 4.25 & 79.14 $\pm$ 8.44 \\ \hline
				\textbf{0.5,0.45} & 49.5 $\pm$ 0.0 & 79.3 $\pm$ 4.44 & \textbf{80.84 $\pm$ 2.54} & 80.2 $\pm$ 7.94 \\ \hline
				\textbf{0.48,0.46} & 54.3 $\pm$ 7.7 & 79.98 $\pm$ 3.12 & 78.03 $\pm$ 2.61 & \textbf{80.22 $\pm$ 8.54} \\ \hline
				\textbf{0.5,0.46} & 49.53 $\pm$ 0.0 & 81.17 $\pm$ 1.84 & \textbf{81.37 $\pm$ 1.60} & 76.62 $\pm$ 9.06 \\ \hline
				\textbf{0.49,0.47} & 56.7 $\pm$ 9.07 & 79.04 $\pm$ 1.23 & \textbf{80.85 $\pm$ 1.77} & 73.19 $\pm$ 9.23 \\ \hline
				\textbf{0.51,0.47} & 49.53 $\pm$ 0.0 & 80.31 $\pm$ 0.87 & \textbf{82.68 $\pm$ 1.32} & 57.77 $\pm$ 10.59 \\ \hline
				\textbf{0.49,0.48} & 52.53 $\pm$ 4.51 & 76.30 $\pm$ 8.98 & \textbf{79.45 $\pm$ 4.14} & 57.89 $\pm$ 6.18 \\ \hline
		\end{tabular}}
		\egroup
		\caption{Averaged accuracy and standard deviation values for MNIST 7-9 dataset across 5 trials when the classification model M is an MLP. The noise rates have been swapped. The table demonstrates that for very high noise rates, i.e., close to 0.5, WGAN based schemes have higher accuracies and low variation among the trials than GLC.}
		\label{tab: MNIST_7-9_high2}
	\end{table}
	
	\begin{table}[]
		\centering
		\bgroup
		\def\arraystretch{0.85}% 
		\scriptsize{ \setlength{\tabcolsep}{0.25em}
			\begin{tabular}{|c|l|c|c|c|}
				\hline
				\textbf{$\rho_+,\rho_-$} & \textbf{SimpleNN} & \textbf{WGANXtraY} & \textbf{WGANXtraYEntr} & \textbf{GLC} \\ \hline
				\textbf{0.49,0.4} & 50.67 $\pm$ 0.0 & 73.32 $\pm$ 3.47 & 72.30 $\pm$ 4.85 & \textbf{86.58 $\pm$ 3.78} \\ \hline
				\textbf{0.45,0.42} & 57.4 $\pm$ 3.57 & 73.2$\pm$ 3.840 & 73.95 $\pm$ 1.41 & \textbf{78.70 $\pm$ 14.49} \\ \hline
				\textbf{0.46,0.45} & 59.47 $\pm$ 10.2 & 74.47 $\pm$ 3.11 & 75.43 $\pm$ 1.99 & \textbf{75.93 $\pm$ 5.38} \\ \hline
				\textbf{0.48,0.45} & 53.63 $\pm$ 5.9 & \textbf{76.30 $\pm$ 1.84} & 72.06$\pm$ 6.07 & 74.86 $\pm$ 13.01 \\ \hline
				\textbf{0.5,0.45} & 50.67$\pm$ 0.0 & 73.52 $\pm$ 2.12 & \textbf{73.81 $\pm$ 4.41} & 59.50$\pm$ 14.12 \\ \hline
				\textbf{0.48,0.46} & 56.39 $\pm$ 7.3 & \textbf{75.72 $\pm$ 0.71} & 73.26 $\pm$ 4.69 & 74.07$\pm$ 3.04 \\ \hline
				\textbf{0.5,0.46} & 50.67$\pm$ 0.0 & \textbf{74.87 $\pm$ 1.95} & 73.68 $\pm$ 2.47 & 63.25 $\pm$ 7.62 \\ \hline
				\textbf{0.49,0.47} & 57.38 $\pm$ 7.19 & 75.34$\pm$ 2.76 & \textbf{77.54 $\pm$ 2.52} & 60.31 $\pm$ 16.9 \\ \hline
				\textbf{0.51,0.47} & 50.67 $\pm$ 0.0 & \textbf{72.55 $\pm$ 3.60} & 70.84 $\pm$ 4.47 & 62.9 $\pm$ 9.09 \\ \hline
				\textbf{0.49,0.48} & 50.7$\pm$ 0.802 & 74.08 $\pm$ 3.77 & \textbf{76.86 $\pm$ 3.18} & 64.82$\pm$ 7.35 \\ \hline
		\end{tabular}}
		\egroup
		\caption{Averaged accuracy and standard deviation values for MNIST 4-9 dataset across 5 trials when the classification model M is an MLP. The noise rates have been swapped. The table demonstrates that for very high noise rates, i.e., close to 0.5, WGAN based schemes have higher accuracies and low variation among the trials than GLC.}
		\label{tab: MNIST_4-9_high2}
	\end{table}
	
	\begin{table}[]
		\centering
		\bgroup
		\def\arraystretch{0.85}% 
		\scriptsize{ \setlength{\tabcolsep}{0.25em}
			\begin{tabular}{|c|l|c|c|c|}
				\hline
				\textbf{$\rho_+,\rho_-$} & \textbf{SimpleNN} & \textbf{WGANXtraY} & \textbf{WGANXtraYEntr} & \textbf{GLC} \\ \hline
				\textbf{0.49,0.4} & 47.52 $\pm$ 0.0 & 94.11 $\pm$ 2.62 & \textbf{96.47 $\pm$ 0.93} & 95.03 $\pm$ 4.32 \\ \hline
				\textbf{0.45,0.42} & 82.94 $\pm$ 16.53 & 96.18 $\pm$ 0.25 & 94.13 $\pm$ 2.15 & \textbf{97.37$\pm$ 0.75} \\ \hline
				\textbf{0.46,0.45} & 91.09 $\pm$ 4.14 & 95.42 $\pm$ 1.21 & 95.70 $\pm$ 0.53 & \textbf{96.93$\pm$ 1.72} \\ \hline
				\textbf{0.48,0.45} & 52.05$\pm$ 7.58 & 94.76 $\pm$ 1.81 & \textbf{95.74 $\pm$ 1.51} & 86.82 $\pm$ 8.57 \\ \hline
				\textbf{0.5,0.45} & 47.52 $\pm$ 0.0 & 95.45 $\pm$ 0.86 & \textbf{95.92 $\pm$ 0.36} & 83.53 $\pm$ 16.28 \\ \hline
				\textbf{0.48,0.46} & 84.89 $\pm$ 18.70 & 95.58$\pm$ 0.92 & \textbf{95.70 $\pm$ 1.25} & 81.68 $\pm$ 6.72 \\ \hline
				\textbf{0.5,0.46} & 56.74 $\pm$ 18.39 & 93.58 $\pm$ 1.76 & \textbf{95.93 $\pm$ 0.56} & 69.8$\pm$ 18.35 \\ \hline
				\textbf{0.49,0.47} & 59.15 $\pm$ 16.95 & 95.17 $\pm$ 0.44 & \textbf{95.77 $\pm$ 0.97} & 87.66$\pm$ 6.74 \\ \hline
				\textbf{0.51,0.47} & 47.52 $\pm$ 0.0 & \textbf{96.19 $\pm$ 0.68} & 95.94 $\pm$ 0.65 & 72.5 $\pm$ 19.72 \\ \hline
				\textbf{0.49,0.48} & 66.77 $\pm$ 15.79 & 96.05$\pm$ 0.3 & \textbf{96.51 $\pm$ 0.75} & 80.6 $\pm$ 11.17 \\ \hline
		\end{tabular}}
		\egroup
		\caption{Averaged accuracy and standard deviation values for MNIST 1-7 dataset across 5 trials when the classification model M is an MLP. The noise rates have been swapped. The table demonstrates that for very high noise rates, i.e., close to 0.5, WGAN based schemes have higher accuracies and low variation among the trials than GLC.}
		\label{tab: MNIST_1-7_high2}
	\end{table}
	
	\begin{table}[]
		\centering
		\bgroup
		\def\arraystretch{0.85}% 
		\scriptsize{ \setlength{\tabcolsep}{0.25em}
			\begin{tabular}{|c|l|c|c|c|}
				\hline
				\textbf{$\rho_+,\rho_-$} & \textbf{SimpleNN} & \textbf{WGANXtraY} & \textbf{WGANXtraYEntr} & \textbf{GLC} \\ \hline
				\textbf{0.49,0.4} & 53.13 $\pm$ 3.47 & 85.41 $\pm$ 9.57 & 85.92 $\pm$ 0.71 & \textbf{86.4 $\pm$ 16.17} \\ \hline
				\textbf{0.45,0.42} & 86.92 $\pm$ 8.73 & 89.83 $\pm$ 2.17 & 91.73 $\pm$ 1.26 & \textbf{91.87 $\pm$ 8.83} \\ \hline
				\textbf{0.46,0.45} & 87.81 $\pm$ 5.72 & 89.64 $\pm$ 3.54 & \textbf{90.06 $\pm$ 3.26} & 78.97 $\pm$ 29.79 \\ \hline
				\textbf{0.48,0.45} & 72.05 $\pm$ 15.0 & \textbf{90.20 $\pm$ 2.23} & 88.95 $\pm$ 5.18 & 89.02 $\pm$ 7.82 \\ \hline
				\textbf{0.5,0.45} & 50.56 $\pm$ 0.0 & 87.34 $\pm$ 1.61 & \textbf{89.01 $\pm$ 1.26} & 87.31 $\pm$ 9.92 \\ \hline
				\textbf{0.48,0.46} & 62.94 $\pm$ 14.05 & 89.09 $\pm$ 3.78 & 86.82 $\pm$ 6.97 & \textbf{92.98 $\pm$ 3.90} \\ \hline
				\textbf{0.5,0.46} & 60.29 $\pm$ 13.79 & \textbf{87.63 $\pm$ 1.45} & 84.48$\pm$ 4.10 & 76.19$\pm$ 14.64 \\ \hline
				\textbf{0.49,0.47} & 60.14 $\pm$ 10.33 & 85.00 $\pm$ 6.33 & \textbf{90.30 $\pm$ 2.18} & 62.04 $\pm$ 18.35 \\ \hline
				\textbf{0.51,0.47} & 50.567 $\pm$ 0.0 & \textbf{87.00 $\pm$ 5.30} & 82.06 $\pm$ 15.81 & 65.6 $\pm$ 18.09 \\ \hline
				\textbf{0.49,0.48} & 57.91 $\pm$ 13.11 & \textbf{91.26 $\pm$ 0.79} & 91.13 $\pm$ 1.82 & 67.8 $\pm$ 10.19 \\ \hline
		\end{tabular}}
		\egroup
		\caption{Averaged accuracy and standard deviation values for MNIST 0-6 dataset across 5 trials when the classification model M is an MLP. The noise rates have been swapped. The table demonstrates that for very high noise rates, i.e., close to 0.5, WGAN based schemes have higher accuracies and low variation among the trials than GLC.}
		\label{tab: MNIST_0-6_high2}
	\end{table}
	
	\begin{table}[]
		\centering
		\bgroup
		\def\arraystretch{0.85}% 
		\scriptsize{ \setlength{\tabcolsep}{0.25em}
			\begin{tabular}{|c|l|c|c|c|}
				\hline
				\textbf{$\rho_+,\rho_-$} & \textbf{SimpleNN} & \textbf{WGANXtraY} & \textbf{WGANXtraYEntr} & \textbf{GLC} \\ \hline
				\textbf{0.49,0.4} & 53.43$\pm$ 14.15 & \textbf{99.47$\pm$ 0.41} & 98.49$\pm$ 1.77 & 96.58 $\pm$ 2.73 \\ \hline
				\textbf{0.45,0.42} & 98.78$\pm$ 1.09 & 98.45 $\pm$ 1.17 & \textbf{99.63$\pm$ 0.22} & 96.52$\pm$ 5.47 \\ \hline
				\textbf{0.46,0.45} & 98.95$\pm$ 0.77 & \textbf{99.3 $\pm$ 0.23} & 96.43 $\pm$ 6.44 & 95.15 $\pm$ 6.12 \\ \hline
				\textbf{0.48,0.45} & 88.31 $\pm$ 20.9 & \textbf{99.28 $\pm$ 0.53} & 99.25 $\pm$ 0.39 & 96.02 $\pm$ 2.72 \\ \hline
				\textbf{0.5,0.45} & 51.06 $\pm$ 7.90 & 97.96 $\pm$ 1.60 & \textbf{99.24$\pm$ 0.68} & 87.13 $\pm$ 14.9 \\ \hline
				\textbf{0.48,0.46} & 67.34 $\pm$ 25.74 & \textbf{99.47 $\pm$ 0.19} & 99.41 $\pm$ 0.26 & 73.50 $\pm$ 14.18 \\ \hline
				\textbf{0.5,0.46} & 50.07 $\pm$ 7.47 & 98.97$\pm$ 0.61 & \textbf{99.13 $\pm$ 0.92} & 95.9 $\pm$ 3.18 \\ \hline
				\textbf{0.49,0.47} & 51.64 $\pm$ 4.29 & 99.15 $\pm$ 0.94 & \textbf{99.59$\pm$ 0.18} & 91.35$\pm$ 13.29 \\ \hline
				\textbf{0.51,0.47} & 46.81 $\pm$ 0.96 & 99.2 $\pm$ 0.78 & \textbf{99.62$\pm$ 0.07} & 83.26 $\pm$ 16.59 \\ \hline
				\textbf{0.49,0.48} & 65.36 $\pm$ 23.45 & 98.02 $\pm$ 2.39 & \textbf{99.16$\pm$ 1.21} & 80.5 $\pm$ 17.81 \\ \hline
		\end{tabular}}
		\egroup
		\caption{Averaged accuracy and standard deviation values for MNIST 0-1 dataset across 5 trials when the classification model M is an MLP. The noise rates have been swapped. The table demonstrates that for very high noise rates, i.e., close to 0.5, WGAN based schemes have higher accuracies and low variation among the trials than GLC.}
		\label{tab: MNIST_0-1_high2}
	\end{table}
	
	%%%%%%%%%%%%%%%%%%%%%%%%%%%%%%%%%%%%%%%%%%%%%%%%%%%%%%

	\begin{table}[]
		\centering
		\bgroup
		\def\arraystretch{0.85}% 
		\scriptsize{ \setlength{\tabcolsep}{0.25em}
			\begin{tabular}{|c|c|c|c|c|c|c|}
\hline
\textbf{$\rho_+,\rho_-$} & \textbf{Simple NN} & \textbf{WGANXtraY} & \textbf{WGANXtraYEntr} & \textbf{GLC} & \textbf{GCE} & \textbf{LDMI} \\ \hline
\textbf{0.4,0.49} & 66.9 $\pm$ 8.86 & 86.35 $\pm$ 2.02 & 86.61 $\pm$ 0.93 & 89.41 $\pm$ 1.03 & 50.0 $\pm$ 0.0 & \textbf{89.47 $\pm$ 2.31} \\ \hline
\textbf{0.42,0.45} & 84.86 $\pm$ 3.15 & 78.66 $\pm$ 14.43 & 86.4 $\pm$ 1.78 & \textbf{90.16 $\pm$ 1.3} & 70.21 $\pm$ 14.74 & 85.52 $\pm$ 6.05 \\ \hline
\textbf{0.45,0.46} & 87.0 $\pm$ 3.26 & 87.16 $\pm$ 1.03 & 80.08 $\pm$ 15.08 & 86.87 $\pm$ 6.28 & 87.65 $\pm$ 2.80 & \textbf{89.61 $\pm$ 1.88} \\ \hline
\textbf{0.45,0.48} & 73.03 $\pm$ 18.83 & 87.95 $\pm$ 1.41 & 84.59 $\pm$ 1.64 & \textbf{89.05 $\pm$ 0.92} & 50.0 $\pm$ 0.0 & 85.75 $\pm$ 3.41 \\ \hline
\textbf{0.45,0.5} & 65.51 $\pm$ 15.09 & \textbf{87.38 $\pm$ 1.74} & 77.35 $\pm$ 14.58 & 85.0 $\pm$ 3.82 & 50.0 $\pm$ 0.0 & 83.13 $\pm$ 7.56 \\ \hline
\textbf{0.46,0.48} & 80.49 $\pm$ 4.82 & \textbf{87.53 $\pm$ 0.69} & 86.63 $\pm$ 0.61 & 84.28 $\pm$ 5.23 & 54.96 $\pm$ 9.87 & 84.8 $\pm$ 5.15 \\ \hline
\textbf{0.46,0.5} & 58.79 $\pm$ 10.39 & 82.55 $\pm$ 10.03 & 85.26 $\pm$ 4.37 & 70.01 $\pm$ 16.46 & 50.0 $\pm$ 0.0 & \textbf{87.49 $\pm$ 2.6} \\ \hline
\textbf{0.47,0.49} & 62.87$\pm$ 14.08 & 85.85 $\pm$ 2.81 & \textbf{86.89 $\pm$ 0.41} & 84.6 $\pm$ 4.85 & 58.31 $\pm$ 13.23 & 86.72 $\pm$ 2.87 \\ \hline
\textbf{0.47,0.51} & 50.17 $\pm$ 0.34 & 70.55 $\pm$ 35.29 & \textbf{88.25 $\pm$ 1.18} & 86.44 $\pm$ 1.15 & 50.0 $\pm$ 0.0 & 76.4 $\pm$ 19.08 \\ \hline
\textbf{0.48,0.49} & 60.14 $\pm$ 13.36 & 86.07 $\pm$ 1.32 & \textbf{88.03 $\pm$ 0.59} & 89.47 $\pm$ 0.40 & 51.42 $\pm$ 2.85 & 67.86 $\pm$ 12.61 \\ \hline
\textbf{0.5,0.5} & 38.52 $\pm$ 9.61 & \textbf{85.61 $\pm$ 1.71} & 79.67 $\pm$ 14.86 & 75.35 $\pm$ 16.16 & 47.71 $\pm$ 4.53 & 64.72 $\pm$ 14.46 \\ \hline
\end{tabular}}
		\egroup
		\caption{Averaged accuracy and standard deviation values for Fashion MNIST 7-9 (Sneakers and Boots) dataset across 5 trials when the classification model M is an MLP. The table demonstrates that for very high noise rates, i.e., close to 0.5, WGAN based schemes have higher accuracies and low variation among the trials than GCE.}
		\label{tab: Fashion_MNIST_7-9_high1}
	\end{table}
	
	\begin{table}[]
		\centering
		\bgroup
		\def\arraystretch{0.85}% 
		\scriptsize{ \setlength{\tabcolsep}{0.25em}
			\begin{tabular}{|c|c|c|c|c|c|c|}
\hline
\textbf{$\rho_+,\rho_-$} & \textbf{Simple NN} & \textbf{WGANXtraY} & \textbf{WGANXtraYEntr} & \textbf{GLC} & \textbf{GCE} & \textbf{LDMI} \\ \hline
\textbf{0.4,0.49} & 50.4 $\pm$ 0.95 & 77.36 $\pm$ 1.11 & 76.75 $\pm$ 2.51 & 66.85 $\pm$ 10.82 & 50.0 $\pm$ 0.0 & \textbf{79.47$\pm$ 1.24} \\ \hline
\textbf{0.42,0.45} & 73.58 $\pm$ 3.83 & 78.07 $\pm$ 0.25 & 76.28 $\pm$ 1.65 & \textbf{79.23 $\pm$ 0.58} & 56.6 $\pm$ 9.99 & 78.91 $\pm$ 1.26 \\ \hline
\textbf{0.45,0.46} & 75.95 $\pm$ 2.36 & 72.63 $\pm$ 11.32 & 72.15 $\pm$ 11.09 & 71.71 $\pm$ 10.92 & 54.67 $\pm$ 7.89 & \textbf{78.05$\pm$ 1.39} \\ \hline
\textbf{0.45,0.48} & 54.52 $\pm$ 8.98 & \textbf{78.04 $\pm$ 0.44} & 71.93 $\pm$ 10.98 & 68.2 $\pm$ 13.37 & 50.0 $\pm$ 0.0 & 76.94 $\pm$ 5.25 \\ \hline
\textbf{0.45,0.5} & 50.0 $\pm$ 0.0 & \textbf{78.50 $\pm$ 0.34} & 76.09 $\pm$ 2.23 & 66.76 $\pm$ 11.03 & 50.0 $\pm$ 0.0 & 73.62 $\pm$ 5.22 \\ \hline
\textbf{0.46,0.48} & 60.14 $\pm$ 4.34 & \textbf{78.11 $\pm$ 0.58} & 69.58 $\pm$ 11.55 & 54.52 $\pm$ 8.96 & 50.04 $\pm$ 0.08 & 77.88 $\pm$ 2.08 \\ \hline
\textbf{0.46,0.5} & 53.25 $\pm$ 5.41 & 77.37 $\pm$ 1.86 & 76.4 $\pm$ 2.73 & 67.83 $\pm$ 13.00 & 50.0 $\pm$ 0.0 & \textbf{78.15 $\pm$ 1.38} \\ \hline
\textbf{0.47,0.49} & 53.78 $\pm$ 5.63 & 77.74 $\pm$ 1.06 & \textbf{77.93 $\pm$ 0.22} & 69.97 $\pm$ 9.37 & 50.0 $\pm$ 0.0 & 72.55 $\pm$ 10.18 \\ \hline
\textbf{0.47,0.51} & 50.0 $\pm$ 0.0 & \textbf{78.23 $\pm$ 0.42} & 77.91 $\pm$ 0.98 & 74.62 $\pm$ 7.72 & 50.0 $\pm$ 0.0 & 73.37 $\pm$ 4.20 \\ \hline
\textbf{0.48,0.49} & 52.77$\pm$ 4.13 & 76.46 $\pm$ 2.08 & 77.97 $\pm$ 0.23 & \textbf{78.59 $\pm$ 0.72} & 52.82 $\pm$ 4.59 & 72.08 $\pm$ 10.52 \\ \hline
\textbf{0.5,0.5} & 56.15 $\pm$ 6.079 & 77.14 $\pm$ 0.83 & \textbf{77.35 $\pm$ 0.66} & 71.10 $\pm$ 7.51 & 49.83 $\pm$ 0.20 & 55.16 $\pm$ 25.36 \\ \hline
\end{tabular}}
		\egroup
		\caption{Averaged accuracy and standard deviation values for Fashion MNIST 0-6 (Tshirt and Shirt) dataset across 5 trials when the classification model M is an MLP. The table demonstrates that for very high noise rates, i.e., close to 0.5, WGAN based schemes have higher accuracies and low variation among the trials than GLC, GCE and LDMI on most cases.}
		\label{tab: Fashion_MNIST_0-6_high1}
	\end{table}
	
	\begin{table}[]
		\centering
		\bgroup
		\def\arraystretch{0.85}% 
		\scriptsize{ \setlength{\tabcolsep}{0.25em}
		\begin{tabular}{|c|c|c|c|c|c|c|}
\hline
\textbf{$\rho_+,\rho_-$} & \textbf{Simple NN} & \textbf{WGANXtraY} & \textbf{WGANXtraYEntr} & \textbf{GLC} & \textbf{GCE} & \textbf{LDMI} \\ \hline
\textbf{0.4,0.49} & 50.0 $\pm$ 0.0 & 58.76$\pm$ 5.07 & 59.77 $\pm$ 5.87 & 60.11 $\pm$ 11.42 & 50.0 $\pm$ 0.0 & \textbf{73.49 $\pm$ 4.54} \\ \hline
\textbf{0.42,0.45} & 54.89 $\pm$ 6.35 & 57.52$\pm$ 2.50 & 57.81 $\pm$ 4.62 & 61.91$\pm$ 13.12 & 57.16 $\pm$ 7.61 & \textbf{81.82 $\pm$ 1.57} \\ \hline
\textbf{0.45,0.46} & 60.9 $\pm$ 7.27 & 57.57 $\pm$ 4.06 & 58.58 $\pm$ 4.20 & 70.72 $\pm$ 6.03 & 54.85 $\pm$ 6.36 & \textbf{71.32 $\pm$ 5.23} \\ \hline
\textbf{0.45,0.48} & 56.98 $\pm$ 8.73 & 57.07 $\pm$ 5.22 & 59.02 $\pm$ 5.14 & 58.17 $\pm$ 10.02 & 50.0 $\pm$ 0.0 & \textbf{69.55 $\pm$ 4.15} \\ \hline
\textbf{0.45,0.5} & 50.0 $\pm$ 0.0 & 60.31 $\pm$ 4.48 & 60.86 $\pm$ 5.19 & 57.56 $\pm$ 8.90 & 50.0 $\pm$ 0.0 & \textbf{68.5 $\pm$ 10.21} \\ \hline
\textbf{0.46,0.48} & 59.54 $\pm$ 7.36 & 59.61 $\pm$ 4.92 & 59.04 $\pm$ 5.94 & 50.0 $\pm$ 0.0 & 55.71 $\pm$ 7.43 & \textbf{68.41 $\pm$ 2.96} \\ \hline
\textbf{0.46,0.5} & 49.98$\pm$ 0.04 & \textbf{60.5 $\pm$ 1.51} & 56.37 $\pm$ 5.63 & 55.6 $\pm$ 6.64 & 50.0 $\pm$ 0.0 & 60.12$\pm$ 7.31 \\ \hline
\textbf{0.47,0.49} & 53.86 $\pm$ 4.61 & 52.25 $\pm$ 9.55 & 60.64 $\pm$ 2.08 & 53.97$\pm$ 5.16 & 51.21 $\pm$ 2.41 & \textbf{69.41 $\pm$ 4.35} \\ \hline
\textbf{0.47,0.51} & 51.87 $\pm$ 3.75 & \textbf{61.95 $\pm$ 0.35} & 60.01 $\pm$ 2.49 & 54.12 $\pm$ 7.67 & 50.0 $\pm$ 0.0 & 57.3 $\pm$ 5.57 \\ \hline
\textbf{0.48,0.49} & 50.72 $\pm$ 3.61 & 54.0 $\pm$ 5.69 & \textbf{60.79 $\pm$ 1.94} & 55.01 $\pm$ 9.41 & 53.61 $\pm$ 5.51 & 60.8 $\pm$ 3.11 \\ \hline
\textbf{0.5,0.5} & 49.87 $\pm$ 0.44 & 53.37 $\pm$ 4.89 & \textbf{61.71 $\pm$ 7.39} & 53.87 $\pm$ 7.48 & 46.01 $\pm$ 4.43 & 48.89 $\pm$ 12.10 \\ \hline
\end{tabular}}
		\egroup
		\caption{Averaged accuracy and standard deviation values for Fashion MNIST 2-4 (Pullover and Coat) dataset across 5 trials when the classification model M is an MLP. The table demonstrates that for very high noise rates, i.e., close to 0.5, WGAN based schemes have higher accuracies and low variation for some noise rates; for others LDMI performs well.}
		\label{tab: Fashion_MNIST_2-4_high1}
	\end{table}
	
	\begin{table}[]
		\centering
		\bgroup
		\def\arraystretch{0.85}% 
		\scriptsize{ \setlength{\tabcolsep}{0.25em}
		\begin{tabular}{|c|c|c|c|c|c|c|}
\hline
\textbf{$\rho_+,\rho_-$} & \textbf{Simple NN} & \textbf{WGANXtraY} & \textbf{WGANXtraYEntr} & \textbf{GLC} & \textbf{GCE} & \textbf{LDMI} \\ \hline
\textbf{0.4,0.49} & 65.74 $\pm$ 13.97 & 91.21 $\pm$ 6.44 & 86.60 $\pm$ 0.93 & 89.61 $\pm$ 11.46 & 50.0 $\pm$ 0.0 & \textbf{94.73 $\pm$ 3.65} \\ \hline
\textbf{0.42,0.45} & 93.75 $\pm$ 3.14 & 95.63 $\pm$ 1.50 & 86.4 $\pm$ 1.78 & 96.00 $\pm$ 0.80 & 73.63 $\pm$ 19.69 & \textbf{96.13 $\pm$ 1.05} \\ \hline
\textbf{0.45,0.46} & 93.65 $\pm$ 1.81 & \textbf{95.97 $\pm$ 0.84} & 80.08 $\pm$ 15.08 & 79.66 $\pm$ 18.77 & 74.91 $\pm$ 19.20 & 94.67 $\pm$ 2.03 \\ \hline
\textbf{0.45,0.48} & 79.6 $\pm$ 16.41 & 92.78 $\pm$ 2.80 & 84.59 $\pm$ 1.64 & \textbf{95.44 $\pm$ 0.92} & 50.0 $\pm$ 0.0 & 93.99 $\pm$ 3.05 \\ \hline
\textbf{0.45,0.5} & 54.67 $\pm$ 5.97 & 85.24 $\pm$ 17.73 & 77.35 $\pm$ 14.58 & 79.80$\pm$ 20.36 & 50.0 $\pm$ 0.0 & \textbf{92.92$\pm$ 6.62} \\ \hline
\textbf{0.46,0.48} & 65.54 $\pm$ 14.56 & \textbf{95.68 $\pm$ 1.07} & 86.63 $\pm$ 0.61 & 92.32 $\pm$ 8.83 & 55.32 $\pm$ 10.12 & 77.75 $\pm$ 12.57 \\ \hline
\textbf{0.46,0.5} & 62.17 $\pm$ 14.41 & \textbf{94.8 $\pm$ 1.89} & 85.26 $\pm$ 4.37 & 90.95 $\pm$ 9.109 & 50.0 $\pm$ 0.0 & 89.81 $\pm$ 6.11 \\ \hline
\textbf{0.47,0.49} & 80.89 $\pm$ 16.05 & \textbf{95.75 $\pm$ 1.07} & 86.89 $\pm$ 0.40 & 95.42 $\pm$ 1.79 & 50.0 $\pm$ 0.0 & 91.37 $\pm$ 2.12 \\ \hline
\textbf{0.47,0.51} & 58.22 $\pm$ 7.47 & \textbf{92.35 $\pm$ 6.14} & 88.25 $\pm$ 1.18 & 94.05 $\pm$ 3.51 & 50.0 $\pm$ 0.0 & 88.1 $\pm$ 4.40 \\ \hline
\textbf{0.48,0.49} & 79.11 $\pm$ 14.85 & \textbf{93.19 $\pm$ 2.84} & 88.03 $\pm$ 0.59 & 94.01 $\pm$ 1.41 & 54.95 $\pm$ 9.9 & 93.06 $\pm$ 2.47 \\ \hline
\textbf{0.5,0.5} & 46.66 $\pm$ 10.15 & \textbf{93.32 $\pm$ 3.20} & 79.67 $\pm$ 14.86 & 95.47 $\pm$ 0.98 & 47.72 $\pm$ 4.61 & 72.0 $\pm$ 27.35 \\ \hline
\end{tabular}}
		\egroup
		\caption{Averaged accuracy and standard deviation values for Fashion MNIST 2-3 (Pullover and Dress) dataset across 5 trials when the classification model M is an MLP. The table demonstrates that for very high noise rates, i.e., close to 0.5, WGAN based schemes have higher accuracies for more than half of the noise rates considered.}
		\label{tab: Fashion_MNIST_2-3_high1}
	\end{table}
	
	%%%%%%%%%%%%%%%%%%%%%%%%%%%%%%%%%%%%%%%%%%%%%%%%%%%%%%%%%%%%%%%%

	\begin{table}[]
		\centering
		\bgroup
		\def\arraystretch{0.85}% 
		\scriptsize{ \setlength{\tabcolsep}{0.25em}
			\begin{tabular}{|c|l|c|c|c|}
				\hline
				\textbf{$\rho_+,\rho_-$} & \textbf{SimpleNN} & \textbf{WGANXtraY} & \textbf{WGANXtraYEntr} & \textbf{GLC} \\ \hline
				\textbf{0.49,0.4} & 58.94 $\pm$ 13.07 & 88.60 $\pm$ 0.88 & 87.92 $\pm$ 0.69 & \textbf{88.66 $\pm$ 3.51} \\ \hline
				\textbf{0.45,0.42} & \textbf{88.04 $\pm$ 3.23} & 87.63 $\pm$ 0.66 & 70.42 $\pm$ 35.22 & 84.27 $\pm$ 7.73 \\ \hline
				\textbf{0.46,0.45} & \textbf{89.15 $\pm$ 2.26} & 87.47 $\pm$ 0.68 & 78.22 $\pm$ 14.30 & 86.58 $\pm$ 3.29 \\ \hline
				\textbf{0.48,0.45} & 72.56 $\pm$ 16.93 & 87.47 $\pm$ 0.86 & 87.65 $\pm$ 1.19 & \textbf{89.72 $\pm$ 0.76} \\ \hline
				\textbf{0.5,0.45} & 50.89 $\pm$ 1.58 & 84.98 $\pm$ 2.76 & \textbf{86.81 $\pm$ 1.67} & 80.22 $\pm$ 15.17 \\ \hline
				\textbf{0.48,0.46} & 70.13 $\pm$ 16.63 & \textbf{87.6 $\pm$ 2.33} & 85.35 $\pm$ 5.52 & 86.56 $\pm$ 3.66 \\ \hline
				\textbf{0.5,0.46} & 64.93 $\pm$ 13.50 & \textbf{88.07 $\pm$ 1.11} & 79.80 $\pm$ 14.93 & 79.62 $\pm$ 14.98 \\ \hline
				\textbf{0.49,0.47} & 78.35 $\pm$ 8.52 & \textbf{88.28 $\pm$ 1.45} & 87.53 $\pm$ 0.96 & 88.12 $\pm$ 2.84 \\ \hline
				\textbf{0.51,0.47} & 50.24 $\pm$ 0.45 & 87.47 $\pm$ 1.16 & 86.28 $\pm$ 2.45 & \textbf{87.53 $\pm$ 2.56} \\ \hline
				\textbf{0.49,0.48} & 70.96 $\pm$ 16.27 & 79.7 $\pm$ 14.89 & \textbf{80.63 $\pm$ 13.24} & 75.72 $\pm$ 14.72 \\ \hline
		\end{tabular}}
		\egroup
		\caption{Averaged accuracy and standard deviation values for Fashion MNIST 7-9 (Sneakers and Boots) dataset across 5 trials when the classification model M is an MLP.  The noise rates have been swapped. The table demonstrates that for very high noise rates, i.e., close to 0.5, WGAN based schemes have higher or comparable average accuracies when compared to GLC.}
		\label{tab: Fashion_MNIST_7-9_high2}
	\end{table}
	
	\begin{table}[]
		\centering
		\bgroup
		\def\arraystretch{0.85}% 
		\scriptsize{ \setlength{\tabcolsep}{0.25em}
			\begin{tabular}{|c|l|c|c|c|}
				\hline
				\textbf{$\rho_+,\rho_-$} & \textbf{SimpleNN} & \textbf{WGANXtraY} & \textbf{WGANXtraYEntr} & \textbf{GLC} \\ \hline
				\textbf{0.49,0.4} & 50.03 $\pm$ 0.05 & 63.52 $\pm$ 8.9 & \textbf{78.47 $\pm$ 0.26} & 72.22 $\pm$ 11.15 \\ \hline
				\textbf{0.45,0.42} & 63.41 $\pm$ 8.98 & 65.44 $\pm$ 15.40 & 72.88 $\pm$ 9.89 & \textbf{79.17 $\pm$ 2.27} \\ \hline
				\textbf{0.46,0.45} & 69.85 $\pm$ 7.82 & 71.95 $\pm$ 10.97 & 76.31 $\pm$ 3.68 & \textbf{78.19 $\pm$ 3.70} \\ \hline
				\textbf{0.48,0.45} & 65.07 $\pm$ 7.36 & 72.86 $\pm$ 10.26 & 61.07 $\pm$ 30.57 & \textbf{77.91 $\pm$ 0.57} \\ \hline
				\textbf{0.5,0.45} & 50.22 $\pm$ 0.44 & 66.41 $\pm$ 13.57 & 66.64 $\pm$ 13.59 & \textbf{73.05 $\pm$ 10.12} \\ \hline
				\textbf{0.48,0.46} & 53.36 $\pm$ 4.45 & 72.17 $\pm$ 11.09 & \textbf{77.17 $\pm$ 1.08} & 68.66 $\pm$ 13.39 \\ \hline
				\textbf{0.5,0.46} & 51.48 $\pm$ 2.98 & \textbf{77.26 $\pm$ 1.14} & 72.83 $\pm$ 11.41 & 58.11 $\pm$ 10.73 \\ \hline
				\textbf{0.49,0.47} & 51.04 $\pm$ 1.48 & 71.8 $\pm$ 10.62 & \textbf{78.25 $\pm$ 0.09} & 66.65 $\pm$ 13.64 \\ \hline
				\textbf{0.51,0.47} & 50.0 $\pm$ 0.0 & 69.47 $\pm$ 9.99 & \textbf{77.03 $\pm$ 1.12} & 70.82$\pm$ 10.92 \\ \hline
				\textbf{0.49,0.48} & 58.39 $\pm$ 10.29 & \textbf{74.85 $\pm$ 4.86} & 59.79 $\pm$ 14.16 & 60.48 $\pm$ 12.12 \\ \hline
		\end{tabular}}
		\egroup
		\caption{Averaged accuracy and standard deviation values for Fashion MNIST 0-6 (Tshirt and Shirt) dataset across 5 trials when the classification model M is an MLP.  The noise rates have been swapped.  The table demonstrates that for very high noise rates, i.e., close to 0.5, WGAN based schemes have higher accuracies than GLC in most cases.}
		\label{tab: Fashion_MNIST_0-6_high2}
	\end{table}
	
	\begin{table}[]
		\centering
		\bgroup
		\def\arraystretch{0.85}% 
		\scriptsize{ \setlength{\tabcolsep}{0.25em}
			\begin{tabular}{|c|l|c|c|c|}
				\hline
				\textbf{$\rho_+,\rho_-$} & \textbf{SimpleNN} & \textbf{WGANXtraY} & \textbf{WGANXtraYEntr} & \textbf{GLC} \\ \hline
				\textbf{0.49,0.4} & 50.0 $\pm$ 0.0 & 57.52 $\pm$ 3.94 & 52.64 $\pm$ 3.62 & \textbf{76.09 $\pm$ 4.23} \\ \hline
				\textbf{0.45,0.42} & 57.01 $\pm$ 5.89 & 55.21 $\pm$ 6.99 & 53.91 $\pm$ 4.44 & \textbf{71.33 $\pm$ 9.24} \\ \hline
				\textbf{0.46,0.45} & 56.96 $\pm$ 9.93 & \textbf{61.4 $\pm$ 3.35} & 60.06 $\pm$ 5.01 & 57.32 $\pm$ 9.50 \\ \hline
				\textbf{0.48,0.45} & 52.73 $\pm$ 2.34 & \textbf{60.73 $\pm$ 2.65} & 59.0 $\pm$ 3.71 & 59.69 $\pm$ 10.82 \\ \hline
				\textbf{0.5,0.45} & 53.89 $\pm$ 7.78 & \textbf{59.70$\pm$ 4.88} & 59.65 $\pm$ 5.20 & 52.74$\pm$ 2.64 \\ \hline
				\textbf{0.48,0.46} & 53.07$\pm$ 4.38 & 59.19 $\pm$ 5.05 & 61.14 $\pm$ 6.87 & \textbf{68.8 $\pm$ 3.59} \\ \hline
				\textbf{0.5,0.46} & 52.26 $\pm$ 4.47 & \textbf{61.37 $\pm$ 1.94} & 52.16 $\pm$ 6.35 & 55.02 $\pm$ 7.00 \\ \hline
				\textbf{0.49,0.47} & 53.07 $\pm$ 4.33 & 56.23 $\pm$ 5.87 & \textbf{60.58 $\pm$ 2.09} & 58.4 $\pm$ 9.12 \\ \hline
				\textbf{0.51,0.47} & 50.01 $\pm$ 0.01 & 54.66$\pm$ 3.69 & \textbf{60.98 $\pm$ 2.15} & 50.63 $\pm$ 1.25 \\ \hline
				\textbf{0.49,0.48} & 55.15 $\pm$ 6.42 & \textbf{61.1 $\pm$ 1.51} & 55.58 $\pm$ 4.86 & 51.89 $\pm$ 3.11 \\ \hline
		\end{tabular}}
		\egroup
		\caption{Averaged accuracy and standard deviation values for Fashion MNIST 2-4 (Pullover and Coat) dataset across 5 trials when the classification model M is an MLP.  The noise rates have been swapped. The table demonstrates that for very high noise rates, i.e., close to 0.5, WGAN based schemes have higher accuracies and low variation among the trials than GLC.}
		\label{tab: Fashion_MNIST_2-4_high2}
	\end{table}
	
	\begin{table}[]
		\centering
		\bgroup
		\def\arraystretch{0.85}% 
		\scriptsize{ \setlength{\tabcolsep}{0.25em}
			\begin{tabular}{|c|l|c|c|c|}
				\hline
				\textbf{$\rho_+,\rho_-$} & \textbf{SimpleNN} & \textbf{WGANXtraY} & \textbf{WGANXtraYEntr} & \textbf{GLC} \\ \hline
				\textbf{0.49,0.4} & 63.33 $\pm$ 13.92 & 91.85 $\pm$ 7.44 & \textbf{93.88 $\pm$ 1.97} & 87.89 $\pm$ 15.82 \\ \hline
				\textbf{0.45,0.42} & 88.25 $\pm$ 9.90 & 94.59 $\pm$ 1.89 & 86.44 $\pm$ 9.31 & \textbf{96.75 $\pm$ 0.11} \\ \hline
				\textbf{0.46,0.45} & 92.82 $\pm$ 3.00 & 92.68 $\pm$ 4.59 & 92.72 $\pm$ 3.49 & \textbf{96.28 $\pm$ 0.75} \\ \hline
				\textbf{0.48,0.45} & 77.72 $\pm$ 15.22 & \textbf{93.19 $\pm$ 3.62} & 76.05 $\pm$ 38.06 & 90.05 $\pm$ 8.21 \\ \hline
				\textbf{0.5,0.45} & 62.89 $\pm$ 11.96 & \textbf{89.86 $\pm$ 6.67} & 73.55 $\pm$ 37.04 & 87.85 $\pm$ 10.80 \\ \hline
				\textbf{0.48,0.46} & 88.56 $\pm$ 7.17 & 95.54 $\pm$ 1.54 & \textbf{96.11 $\pm$ 0.55} & 85.67 $\pm$ 17.85 \\ \hline
				\textbf{0.5,0.46} & 65.85 $\pm$ 15.66 & \textbf{93.32$\pm$ 2.82} & 67.07 $\pm$ 37.81 & 84.42 $\pm$ 16.48 \\ \hline
				\textbf{0.49,0.47} & 68.64 $\pm$ 17.18 & \textbf{90.25 $\pm$ 7.55} & 48.32 $\pm$ 41.68 & 86.97 $\pm$ 18.48 \\ \hline
				\textbf{0.51,0.47} & 56.22 $\pm$ 7.72 & 88.7 $\pm$ 7.64 & 89.01 $\pm$ 9.43 & \textbf{91.44 $\pm$ 6.86} \\ \hline
				\textbf{0.49,0.48} & 65.66 $\pm$ 11.33 & 67.97 $\pm$ 35.06 & \textbf{87.4 $\pm$ 7.64} & 80.58 $\pm$ 16.45 \\ \hline
		\end{tabular}}
		\egroup
		\caption{Averaged accuracy and standard deviation values for Fashion MNIST 2-3 (Pullover and Dress) dataset across 5 trials when the classification model M is an MLP.  The noise rates have been swapped. The table demonstrates that for very high noise rates, i.e., close to 0.5, WGAN based schemes have higher accuracies than GLC in most cases.}
		\label{tab: Fashion_MNIST_2-3_high2}
	\end{table}
	
	%%%%%%%%%%%%%%%%%%%%%%%%%%%%%%%%%%%%%%%%%%%%%%%%%%%%%%%%%%%%%%%%%%%%%%

	\begin{table}[]
		\centering
		\bgroup
		\def\arraystretch{0.85}% 
		\scriptsize{ \setlength{\tabcolsep}{0.25em}
			\begin{tabular}{|c|c|c|c|c|c|c|}
\hline
\textbf{$\rho_+,\rho_-$} & \textbf{Simple NN} & \textbf{WGANXtraY} & \textbf{WGANXtraYEntr} & \textbf{GLC} & \textbf{GCE} & \textbf{LDMI} \\ \hline
\textbf{0.4,0.49} & 64.41 $\pm$ 16.51 & \textbf{91.25 $\pm$ 3.55} & 88.98 $\pm$ 3.92 & 66.7 $\pm$ 20.72 & 58.01 $\pm$ 14.46 & 82.09 $\pm$ 7.15 \\ \hline
\textbf{0.42,0.45} & 79.67 $\pm$ 16.99 & \textbf{92.08 $\pm$ 5.40} & 85.78 $\pm$ 5.27 & 75.46 $\pm$ 17.54 & 50.0 $\pm$ 0.0 & 91.59 $\pm$ 4.2 \\ \hline
\textbf{0.45,0.46} & 57.07 $\pm$ 12.56 & \textbf{91.12 $\pm$ 2.78} & 90.63 $\pm$ 2.34 & 61.7$\pm$ 18.50 & 50.0 $\pm$ 0.0 & 82.86 $\pm$ 8.60 \\ \hline
\textbf{0.45,0.48} & 58.60$\pm$ 14.22 & \textbf{92.4 $\pm$ 5.54} & 86.39$\pm$ 4.68 & 70.70 $\pm$ 21.41 & 54.83 $\pm$ 9.67 & 77.51 $\pm$ 13.95 \\ \hline
\textbf{0.45,0.5} & 71.62 $\pm$ 17.84 & \textbf{91.35$\pm$ 6.88} & 84.63 $\pm$ 4.29 & 73.98 $\pm$ 18.94 & 52.91 $\pm$ 5.82 & 69.33$\pm$ 10.32 \\ \hline
\textbf{0.46,0.48} & 70.43 $\pm$ 16.56 & \textbf{88.80$\pm$ 5.65} & 85.29 $\pm$ 8.76 & 74.73 $\pm$ 14.91 & 60.65 $\pm$ 13.45 & 80.03 $\pm$ 7.55 \\ \hline
\textbf{0.46,0.5} & 55.52 $\pm$ 11.98 & 79.96 $\pm$ 15.4 & \textbf{92.57$\pm$ 1.04} & 59.36 $\pm$ 21.61 & 50.0 $\pm$ 0.0 & 70.06$\pm$ 12.20 \\ \hline
\textbf{0.47,0.49} & 62.47 $\pm$ 13.62 & 81.82 $\pm$ 16.53 & \textbf{86.05 $\pm$ 6.26} & 58.62 $\pm$ 9.00 & 50.07 $\pm$ 0.14 & 59.86 $\pm$ 9.90 \\ \hline
\textbf{0.47,0.51} & 52.94 $\pm$ 5.85 & 57.85 $\pm$ 35.1 & \textbf{93.09 $\pm$ 1.27} & 66.63$\pm$ 11.34 & 50.0 $\pm$ 0.0 & 67.46$\pm$ 15.29 \\ \hline
\textbf{0.48,0.49} & 58.09 $\pm$ 9.95 & \textbf{87.31 $\pm$ 7.30} & 83.85 $\pm$ 7.34 & 66.86 $\pm$ 14.03 & 56.52 $\pm$ 13.04 & 69.92 $\pm$ 17.18 \\ \hline
\textbf{0.5,0.5} & 50.2$\pm$ 11.86 & 89.96 $\pm$ 5.48 & \textbf{93.15 $\pm$ 1.08} & 63.41 $\pm$ 14.72 & 50.0 $\pm$ 0.0 & 54.79 $\pm$ 5.58 \\ \hline
\end{tabular}}
		\egroup
		\caption{Averaged arithmetic mean (AM of TPR and TNR) and standard deviation values for imbalanced MNIST 1-7 dataset (imb\_r = 0.2) across 5 trials when the classification model M is an MLP. Gold fraction used is $0.1\%$. The table demonstrates that for very high noise rates, i.e., close to 0.5, in comparison to GLC, GCE And LDMI, WGAN based schemes have higher accuracies and low variation across the trials.}
		\label{tab: MNIST1-7_imb_0.2}
	\end{table}

	\begin{table}[]
		\centering
		\bgroup
		\def\arraystretch{0.85}% 
		\scriptsize{ \setlength{\tabcolsep}{0.25em}
			\begin{tabular}{|c|c|c|c|c|c|c|}
\hline
\textbf{$\rho_+,\rho_-$} & \textbf{Simple NN} & \textbf{WGANXtraY} & \textbf{WGANXtraYEntr} & \textbf{GLC} & \textbf{GCE} & \textbf{LDMI} \\ \hline
\textbf{0.4,0.49} & 50.0 $\pm$ 0.0 & 86.37 $\pm$ 6.66 & \textbf{88.53 $\pm$ 4.48} & 67.66 $\pm$ 14.65 & 50.0 $\pm$ 0.0 & 85.16 $\pm$ 7.32 \\ \hline
\textbf{0.42,0.45} & 54.61 $\pm$ 8.42 & 87.06 $\pm$ 6.73 & 83.49 $\pm$ 9.19 & 92.55 $\pm$ 4.43 & 50.0 $\pm$ 0.0 & \textbf{90.58 $\pm$ 6.66} \\ \hline
\textbf{0.45,0.46} & 69.41 $\pm$ 16.74 & \textbf{87.48 $\pm$ 5.01} & 86.67 $\pm$ 5.51 & 81.39$\pm$ 16.17 & 50.0 $\pm$ 0.0 & 83.61$\pm$ 12.42 \\ \hline
\textbf{0.45,0.48} & 56.26 $\pm$ 9.74 & 84.22 $\pm$ 7.59 & \textbf{90.06 $\pm$ 7.65} & 64.3 $\pm$ 28.14 & 50.0 $\pm$ 0.0 & 81.85 $\pm$ 12.54 \\ \hline
\textbf{0.45,0.5} & 50.0 $\pm$ 0.0 & 89.8 $\pm$ 3.57 & \textbf{90.14 $\pm$ 3.11} & 71.1 $\pm$ 12.23 & 50.0 $\pm$ 0.0 & 83.23 $\pm$ 13.66 \\ \hline
\textbf{0.46,0.48} & 50.77 $\pm$ 1.55 & 89.62 $\pm$ 2.75 & \textbf{92.77 $\pm$ 1.22} & 75.99 $\pm$ 17.91 & 50.0 $\pm$ 0.0 & 80.32 $\pm$ 14.33 \\ \hline
\textbf{0.46,0.5} & 53.50 $\pm$ 7.01 & \textbf{88.08 $\pm$ 3.34} & 84.08 $\pm$ 6.60 & 58.05$\pm$ 17.69 & 50.0 $\pm$ 0.0 & 68.90 $\pm$ 15.46 \\ \hline
\textbf{0.47,0.49} & 51.91 $\pm$ 2.43 & 89.12$\pm$ 2.62 & \textbf{91.18$\pm$ 3.27} & 69.36 $\pm$ 19.40 & 50.0 $\pm$ 0.0 & 81.84$\pm$ 15.52 \\ \hline
\textbf{0.47,0.51} & 49.86$\pm$ 1.15 & \textbf{88.33 $\pm$ 5.93} & 87.98 $\pm$ 5.11 & 54.76 $\pm$ 5.78 & 50.0 $\pm$ 0.0 & 78.69 $\pm$ 17.61 \\ \hline
\textbf{0.48,0.49} & 58.3 $\pm$ 13.91 & \textbf{89.83$\pm$ 2.13} & 87.53 $\pm$ 10.38 & 61.5$\pm$ 4.66 & 50.0 $\pm$ 0.0 & 69.20 $\pm$ 9.60 \\ \hline
\textbf{0.5,0.5} & 63.3 $\pm$ 11.89 & 88.4 $\pm$ 8.26 & \textbf{89.68 $\pm$ 3.12} & 66.1 $\pm$ 15.28 & 50.0 $\pm$ 0.0 & 45.76 $\pm$ 24.97 \\ \hline
\end{tabular}}
		\egroup
		\caption{Averaged arithmetic mean (AM of TPR and TNR) and standard deviation values for imbalanced MNIST 1-7 dataset (imb\_r = 0.7) across 5 trials when the classification model M is an MLP. Gold fraction used is $0.1\%$. The table demonstrates that for very high noise rates, i.e., close to 0.5, WGAN based schemes have higher accuracies and low variation among the trials than GLC, GCE and LDMI.}
		\label{tab: MNIST1-7_imb_0.7}
	\end{table}

	\begin{table}[]
		\centering
		\bgroup
		\def\arraystretch{0.85}% 
		\scriptsize{ \setlength{\tabcolsep}{0.25em}
			\begin{tabular}{|c|c|c|c|c|c|c|}
\hline
\textbf{$\rho_+,\rho_-$} & \textbf{Simple NN} & \textbf{WGANXtraY} & \textbf{WGANXtraYEntr} & \textbf{GLC} & \textbf{GCE} & \textbf{LDMI} \\ \hline
\textbf{0.4,0.49} & 74.65$\pm$ 12.47 & \textbf{91.19 $\pm$ 7.04} & 83.33 $\pm$ 5.53 & 58.66$\pm$ 13.59 & 50.74 $\pm$ 1.49 & 78.69$\pm$ 11.26 \\ \hline
\textbf{0.42,0.45} & 50.20 $\pm$ 0.41 & \textbf{92.3$\pm$ 6.17} & 89.36 $\pm$ 6.00 & 63.86 $\pm$ 18.63 & 50.0 $\pm$ 0.0 & 81.38 $\pm$ 10.18 \\ \hline
\textbf{0.45,0.46} & 50.0 $\pm$ 0.0 & 90.78 $\pm$ 4.06 & \textbf{93.53 $\pm$ 3.09} & 50.0 $\pm$ 0.0 & 50.0 $\pm$ 0.0 & 77.66 $\pm$ 12.33 \\ \hline
\textbf{0.45,0.48} & 57.42 $\pm$ 10.27 & \textbf{90.57 $\pm$ 6.68} & 90.38 $\pm$ 4.90 & 59.1 $\pm$ 11.29 & 50.0 $\pm$ 0.0 & 74.66 $\pm$ 8.65 \\ \hline
\textbf{0.45,0.5} & 57.36 $\pm$ 13.41 & 82.10 $\pm$ 16.68 & \textbf{92.27$\pm$ 7.29} & 50.9 $\pm$ 1.98 & 50.0 $\pm$ 0.0 & 58.15$\pm$ 7.96 \\ \hline
\textbf{0.46,0.48} & 54.50 $\pm$ 7.76 & \textbf{95.37 $\pm$ 1.90} & 89.62 $\pm$ 6.37 & 52.47 $\pm$ 4.94 & 55.75 $\pm$ 11.51 & 70.33 $\pm$ 8.57 \\ \hline
\textbf{0.46,0.5} & 60.69 $\pm$ 15.26 & \textbf{92.45 $\pm$ 3.93} & 90.98 $\pm$ 4.86 & 50.22$\pm$ 0.45 & 56.12 $\pm$ 11.79 & 67.14 $\pm$ 10.56 \\ \hline
\textbf{0.47,0.49} & 59.38$\pm$ 12.16 & 76.24 $\pm$ 14.65 & \textbf{87.52$\pm$ 10.27} & 58.41 $\pm$ 16.82 & 50.34 $\pm$ 0.46 & 68.55$\pm$ 13.27 \\ \hline
\textbf{0.47,0.51} & 50.08 $\pm$ 0.89 & \textbf{92.68$\pm$ 1.54} & 77.31$\pm$ 15.18 & 60.60 $\pm$ 17.00 & 51.17 $\pm$ 2.35 & 62.97$\pm$ 9.56 \\ \hline
\textbf{0.48,0.49} & 51.69 $\pm$ 1.23 & \textbf{92.5 $\pm$ 4.28} & 87.59 $\pm$ 10.04 & 58.2 $\pm$ 10.47 & 51.44 $\pm$ 2.89 & 62.972 $\pm$ 9.69 \\ \hline
\textbf{0.5,0.5} & 47.6$\pm$ 3.13 & \textbf{90.80 $\pm$ 5.41} & 86.36 $\pm$ 7.65 & 48.22 $\pm$ 3.43 & 41.25 $\pm$ 10.70 & 50.36 $\pm$ 13.93 \\ \hline
\end{tabular}}
		\egroup
		\caption{Averaged arithmetic mean (AM of TPR and TNR) and standard deviation values for imbalanced MNIST 0-8 dataset (imb\_r = 0.1) across 5 trials when the classification model M is an MLP. Gold fraction used is $0.1\%$. The table demonstrates that for very high noise rates, i.e., close to 0.5, WGAN based schemes have higher accuracies  than GLC, GCE and LDMI.}
		\label{tab: MNIST0-8_imb_0.1}
	\end{table}

	\begin{table}[]
		\centering
		\bgroup
		\def\arraystretch{0.85}% 
		\scriptsize{ \setlength{\tabcolsep}{0.25em}
			\begin{tabular}{|c|c|c|c|c|c|c|}
\hline
\textbf{$\rho_+,\rho_-$} & \textbf{Simple NN} & \textbf{WGANXtraY} & \textbf{WGANXtraYEntr} & \textbf{GLC} & \textbf{GCE} & \textbf{LDMI} \\ \hline
\textbf{0.4,0.49} & 50.0 $\pm$ 0.0 & 90.10 $\pm$ 3.44 & \textbf{90.44 $\pm$ 2.95} & 65.98 $\pm$ 16.76 & 50.0 $\pm$ 0.0 & 81.17 $\pm$ 8.19 \\ \hline
\textbf{0.42,0.45} & 52.92 $\pm$ 4.19 & \textbf{90.4$\pm$ 1.91} & 83.12 $\pm$ 16.24 & 59.03 $\pm$ 18.07 & 50.0 $\pm$ 0.0 & 88.29$\pm$ 7.62 \\ \hline
\textbf{0.45,0.46} & 62.79 $\pm$ 17.69 & 81.40 $\pm$ 14.40 & \textbf{88.27$\pm$ 3.00} & 71.2$\pm$ 15.49 & 50.0 $\pm$ 0.0 & 80.66 $\pm$ 5.93 \\ \hline
\textbf{0.45,0.48} & 50.461$\pm$ 0.57 & \textbf{90.06 $\pm$ 2.57} & 89.72 $\pm$ 2.34 & 57.87$\pm$ 15.74 & 50.0 $\pm$ 0.0 & 77.14 $\pm$ 9.8 \\ \hline
\textbf{0.45,0.5} & 50.05 $\pm$ 0.10 & \textbf{89.11 $\pm$ 4.30} & 81.53 $\pm$ 16.01 & 50.73 $\pm$ 8.33 & 50.0 $\pm$ 0.0 & 66.09 $\pm$ 10.61 \\ \hline
\textbf{0.46,0.48} & 56.05$\pm$ 8.02 & \textbf{89.18 $\pm$ 2.84} & 73.23 $\pm$ 36.63 & 68.8$\pm$ 12.44 & 50.0 $\pm$ 0.0 & 75.36 $\pm$ 4.44 \\ \hline
\textbf{0.46,0.5} & 52.49 $\pm$ 4.99 & 89.85$\pm$ 2.34 & \textbf{91.38 $\pm$ 1.16} & 61.7 $\pm$ 16.37 & 50.0 $\pm$ 0.0 & 67.16 - 10.18 \\ \hline
\textbf{0.47,0.49} & 50.08 $\pm$ 0.17 & \textbf{90.31 $\pm$ 1.82} & 87.82 $\pm$ 6.94 & 57.36$\pm$ 9.76 & 50.0 $\pm$ 0.0 & 63.08 $\pm$ 9.33 \\ \hline
\textbf{0.47,0.51} & 50.0 $\pm$ 0.0 & 87.5$\pm$ 5.16 & \textbf{89.09 $\pm$ 4.97} & 56.93 $\pm$ 9.45 & 50.0 $\pm$ 0.0 & 66.056 $\pm$ 9.67 \\ \hline
\textbf{0.48,0.49} & 50.0 $\pm$ 0.0 & 86.34 $\pm$ 4.31 & \textbf{89.45 $\pm$ 2.19} & 69.8 $\pm$ 12.30 & 50.0 $\pm$ 0.0 & 64.12 $\pm$ 8.25 \\ \hline
\textbf{0.5,0.5} & 55.43$\pm$ 14.42 & \textbf{92.65 $\pm$ 0.83} & 71.21 $\pm$ 35.74 & 52.52 $\pm$ 5.07 & 48.34 $\pm$ 2.56 & 47.45 $\pm$ 5.3 \\ \hline
\end{tabular}}
		\egroup
		\caption{Averaged arithmetic mean (AM of TPR and TNR) and standard deviation values for imbalanced MNIST 0-8 dataset (imb\_r = 0.8) across 5 trials when the classification model M is an MLP. Gold fraction used is $0.1\%$. The table demonstrates that for very high noise rates, i.e., close to 0.5, WGAN based schemes have higher accuracies and low variation among the trials than GLC, GCE and LDMI.}
		\label{tab: MNIST0-8_imb_0.8}
	\end{table}

	\begin{table}[]
		\centering
		\bgroup
		\def\arraystretch{0.85}% 
		\scriptsize{ \setlength{\tabcolsep}{0.25em}
			\begin{tabular}{|c|c|c|c|c|c|c|}
\hline
\textbf{$\rho_+,\rho_-$} & \textbf{Simple NN} & \textbf{WGANXtraY} & \textbf{WGANXtraYEntr} & \textbf{GLC} & \textbf{GCE} & \textbf{LDMI} \\ \hline
\textbf{0.4,0.49} & 51.04$\pm$ 1.51 & \textbf{86.85 $\pm$ 2.11} & 84.80$\pm$ 4.43 & 79.8 $\pm$ 7.66 & 50.0 $\pm$ 0.0 & 78.42 $\pm$ 7.06 \\ \hline
\textbf{0.42,0.45} & 63.02 $\pm$ 11.57 & 84.30 $\pm$ 4.55 & \textbf{87.37 $\pm$ 2.0} & 83.91 $\pm$ 8.27 & 50.03 $\pm$ 0.06 & 82.24 $\pm$ 4.51 \\ \hline
\textbf{0.45,0.46} & 54.41$\pm$ 5.46 & \textbf{86.82 $\pm$ 2.81} & 84.24 $\pm$ 4.70 & 85.3 $\pm$ 3.41 & 50.0 $\pm$ 0.0 & 74.10 $\pm$ 10.05 \\ \hline
\textbf{0.45,0.48} & 54.37 $\pm$ 3.96 & 86.03 $\pm$ 3.77 & \textbf{88.94 $\pm$ 2.48} & 79.98 $\pm$ 10.05 & 50.50 $\pm$ 1.01 & 67.37 $\pm$ 4.82 \\ \hline
\textbf{0.45,0.5} & 50.0 $\pm$ 0.0 & \textbf{87.08 $\pm$ 1.61} & 85.03 $\pm$ 2.29 & 78.3 $\pm$ 8.96 & 50.0 $\pm$ 0.0 & 64.22 $\pm$ 6.66 \\ \hline
\textbf{0.46,0.48} & 58.24 $\pm$ 8.39 & \textbf{89.51 $\pm$ 0.82} & 86.32 $\pm$ 2.43 & 84.0 $\pm$ 5.39 & 49.84 $\pm$ 1.27 & 60.42 $\pm$ 7.72 \\ \hline
\textbf{0.46,0.5} & 50.02 $\pm$ 0.05 & 84.56 $\pm$ 3.80 & \textbf{86.72 $\pm$ 1.40} & 82.78$\pm$ 2.66 & 50.0 $\pm$ 0.0 & 61.41 $\pm$ 5.84 \\ \hline
\textbf{0.47,0.49} & 53.77 $\pm$ 4.15 & \textbf{88.82 $\pm$ 2.05} & 87.70 $\pm$ 1.89 & 80.39 $\pm$ 7.60 & 49.56 $\pm$ 0.54 & 61.58 $\pm$ 2.32 \\ \hline
\textbf{0.47,0.51} & 50.32 $\pm$ 0.63 & 88.17 $\pm$ 1.38 & \textbf{89.52 $\pm$ 1.67} & 81.86 $\pm$ 4.45 & 50.0 $\pm$ 0.0 & 54.82 $\pm$ 3.01 \\ \hline
\textbf{0.48,0.49} & 55.36 $\pm$ 10.02 & 86.81 $\pm$ 1.12 & \textbf{88.32$\pm$ 1.26} & 78.98 $\pm$ 9.62 & 50.93 $\pm$ 2.38 & 56.04 $\pm$ 3.67 \\ \hline
\textbf{0.5,0.5} & 52.25 $\pm$ 3.62 & \textbf{87.06 $\pm$ 3.09} & 88.93 $\pm$ 3.20 & 79.76 $\pm$ 5.24 & 49.99 $\pm$ 0.01 & 53.19 $\pm$ 3.51 \\ \hline
\end{tabular}}
		\egroup
		\caption{Averaged arithmetic mean (AM of TPR and TNR) and standard deviation values for imbalanced MNIST 4-9 dataset (imb\_r = 0.3) across 5 trials when the classification model M is an MLP. Gold fraction used is $1\%$. The table demonstrates that for very high noise rates, i.e., close to 0.5, WGAN based schemes have higher accuracies and low variation among the trials than GLC, GCE and LDMI.}
		\label{tab: MNIST4-9_imb_0.3}
	\end{table}
	
	\begin{table}[]
		\centering
		\bgroup
		\def\arraystretch{0.85}% 
		\scriptsize{ \setlength{\tabcolsep}{0.25em}
			\begin{tabular}{|c|c|c|c|c|c|c|}
\hline
\textbf{$\rho_+,\rho_-$} & \textbf{Simple NN} & \textbf{WGANXtraY} & \textbf{WGANXtraYEntr} & \textbf{GLC} & \textbf{GCE} & \textbf{LDMI} \\ \hline
\textbf{0.4,0.49} & 50.0 $\pm$ 0.0 & 84.07 $\pm$ 4.18 & \textbf{84.16 $\pm$ 1.79} & 75.72 $\pm$ 14.79 & 50.0 $\pm$ 0.0 & 76.68 $\pm$ 4.35 \\ \hline
\textbf{0.42,0.45} & 50.0 $\pm$ 0.0 & 85.34$\pm$ 1.75 & 83.68 $\pm$ 3.50 & \textbf{86.77 $\pm$ 3.92} & 50.0 $\pm$ 0.0 & 77.17$\pm$ 6.11 \\ \hline
\textbf{0.45,0.46} & 50.0 $\pm$ 0.0 & \textbf{85.40$\pm$ 6.09} & 83.93$\pm$ 4.46 & 80.39 $\pm$ 13.5 & 50.0 $\pm$ 0.0 & 73.43 $\pm$ 1.28 \\ \hline
\textbf{0.45,0.48} & 50.0 $\pm$ 0.0 & 81.04 $\pm$ 4.51 & \textbf{84.39$\pm$ 3.80} & 81.67 $\pm$ 7.58 & 50.0 $\pm$ 0.0 & 65.70 $\pm$ 6.70 \\ \hline
\textbf{0.45,0.5} & 50.0 $\pm$ 0.0 & 80.98 $\pm$ 3.37 & 84.20 $\pm$ 1.27 & \textbf{88.52$\pm$ 2.64} & 50.0 $\pm$ 0.0 & 66.40$\pm$ 6.79 \\ \hline
\textbf{0.46,0.48} & 50.0 $\pm$ 0.0 & 83.26$\pm$ 2.51 & 79.33$\pm$ 4.64 & \textbf{83.90 $\pm$ 5.88} & 50.0 $\pm$ 0.0 & 65.00 $\pm$ 2.97 \\ \hline
\textbf{0.46,0.5} & 50.0 $\pm$ 0.0 & 83.33 $\pm$ 1.22 & \textbf{85.67 $\pm$ 2.60} & 83.32$\pm$ 3.31 & 50.0 $\pm$ 0.0 & 57.40$\pm$ 5.74 \\ \hline
\textbf{0.47,0.49} & 55.23 $\pm$ 10.47 & 83.03 $\pm$ 5.44 & \textbf{84.28 $\pm$ 2.39} & 81.13$\pm$ 4.07 & 50.0 $\pm$ 0.0 & 58.80 $\pm$ 3.68 \\ \hline
\textbf{0.47,0.51} & 50.0 $\pm$ 0.0 & \textbf{85.1 $\pm$ 2.42} & 84.76$\pm$ 3.77 & 73.56 $\pm$ 9.68 & 50.0 $\pm$ 0.0 & 50.36 $\pm$ 3.60 \\ \hline
\textbf{0.48,0.49} & 51.09$\pm$ 2.18 & 80.43 $\pm$ 4.77 & \textbf{84.73 $\pm$ 3.02} & 76.25 $\pm$ 10.46 & 50.0 $\pm$ 0.0 & 62.47$\pm$ 6.44 \\ \hline
\textbf{0.5,0.5} & 55.86$\pm$ 5.34 & \textbf{81.19 $\pm$ 6.51} & 74.52 $\pm$ 12.84 & 79.65 $\pm$ 7.58 & 50.84 $\pm$ 3.03 & 49.53 $\pm$ 4.82 \\ \hline
\end{tabular}}
		\egroup
		\caption{Averaged arithmetic mean (AM of TPR and TNR) and standard deviation values for imbalanced MNIST 4-9 dataset (imb\_r = 0.75) across 5 trials when the classification model M is an MLP.  Gold fraction used is $1\%$. The table demonstrates that for very high noise rates, i.e., close to 0.5, WGAN based schemes have higher accuracies than GLC, GCE and LDMI in most cases.}
		\label{tab: MNIST4-9_imb_0.75}
	\end{table}
	
\end{document}